\documentclass[letterpaper,12pt]{article}

\usepackage{times}
\usepackage[margin=1in]{geometry}

\usepackage{graphicx}
\DeclareGraphicsExtensions{.pdf,.jpeg,.png}
\usepackage{cite}
\usepackage{amsmath}
\usepackage{amssymb}
\usepackage{amsfonts}
\usepackage{titlesec}
\usepackage{enumerate}
\usepackage[linesnumbered,ruled]{algorithm2e}
\usepackage[justification=centering]{caption}
\titlelabel{\thetitle.\quad}
\titleformat*{\section}{\normalfont\bfseries}
\titleformat*{\subsection}{\normalfont\bfseries}
\titleformat*{\subsubsection}{\normalfont\bfseries}
\titleformat*{\paragraph}{\normalfont\bfseries}
\titleformat*{\subparagraph}{\normalfont\bfseries}

\newtheorem{remark}{Remark}

\usepackage{hyperref}
\hypersetup{hypertex=true,
	colorlinks=true,
	linkcolor=red,
	anchorcolor=blue,
	citecolor=green}
\usepackage[all]{xy}
\usepackage{arydshln}

\begin{document}
\date{}

\title{$SE_2(3)$ based Extended Kalman Filtering and Smoothing Framework for Inertial-Integrated Navigation}

\author{Yarong Luo, 
	yarongluo@whu.edu.cn\\
 	Chi Guo,
	guochi@whu.edu.cn\\
	Shengyong You,
	shengyongyou@whu.edu.cn\\
	Jianlang Hu,
	hujianlang123@whu.edu.cn\\
	Jingnan Liu,
	jnliu@whu.edu.cn\\
 	GNSS Research Center, Wuhan University
}



\maketitle

\thispagestyle{empty}

\noindent
{\bf\normalsize Abstract}\newline
{This paper proposes an $SE_2(3)$ based extended Kalman filtering (EKF) framework for the inertial-integrated state estimation problem. The error representation using the straight difference of two vectors in the inertial navigation system may not be reasonable as it does not take the direction difference into consideration. 
	Therefore, we choose to use the $SE_2(3)$ matrix Lie group to represent the state of the inertial-integrated navigation system which consequently leads to the common frame error representation. 
	With the new velocity and position error definition, we leverage the group affine dynamics with the autonomous error properties and derive the error state differential equation for the inertial-integrated navigation on the north-east-down (NED) navigation frame and the earth-centered earth-fixed (ECEF) frame, respectively, the corresponding EKF, terms as $SE_2(3)$ based EKF has also been derived. It provides a new perspective on the geometric EKF with a more sophisticated formula for the inertial-integrated navigation system. Furthermore, we design two new modified error dynamics on the NED frame and the ECEF frame respectively by introducing new auxiliary vectors. Finally the equivalence of the left-invariant EKF and left $SE_2(3)$ based EKF have been shown in navigation frame and ECEF frame.
} \vspace{2ex}

\noindent
{\bf\normalsize Key Words}\newline
{$SE_2(3)$ matrix Lie group, inertial-integrated navigation,  common frame error representation,  consistent extended Kalman filter, group-affine dynamics, autonomous error}

\section{Introduction}
The state error is commonly defined as the straight difference of the vectors without considering the vector's frame representation~\cite{whittaker2018linearized}. This is unreasonable as the state error may not expressed with respect to the same frame coordinate~\cite{whittaker2017inertial,li2018common}, especially for the inertial-integrated navigation system. 
Li et al. proposed a common frame based unscented quaternion estimator which defines the attitude errors in the body frame expressed by the Uncented Kalman Filter and the IMU errors with respect to the common frame basis~\cite{li2018common}.
Scherzinger et al. proposed a new velocity error transformation for the $\phi$ and $\psi$-angle error models which aimed at removing the specific force term in the transition matrix~\cite{scherzinger1994modified} and resulted the coordinate-frame consistency of velocity error vector.
Wang et al. used the same idea to develop the state transformation extended Kalman filter (ST-EKF) for GPS/SINS tightly couple integration~\cite{wang2018state}, SINS/OD integration~\cite{wang2019consistent}, and initial alignment of strapdown inertial navigation system~\cite{wang2019initialalignment}.
In fact, the ST-EKF can be viewed as the same as the modified $\phi$-model error model proposed by Scherzinger in ~\cite{scherzinger1994modified}. 
Meanwhile, for the left invariant measurement model such as GNSS, ST-EKF used the right invariant error definition which is not consistent with left-innovation update.
Recently, Chang studied the $SE(3)$ based EKF for spacecraft attitude estimation which formulas the attitude and gyroscope bias as elements of SE(3)~\cite{chang2020se}.

The above works motivates us to embed the state and the uncertainties into a specially defined and high dimensional matrix Lie group $SE_2(3)$ so that the state evolves on matrix manifold.
We consider the state error on the $SE_2(3)$ matrix Lie group by defining the error as the product between the true state and the inverse of the estimated state and the product between the estimated state and the inverse of the true state so that not only the velocity error is expressed with respect to the common frame, but also the position error is expressed with respect to the common frame. The significant advantage of this common frame based representation is that the resulting expressions are compact and accurate by means of the Lie group theory.
Although this work is motivated by Geometric EKF~\cite{whittaker2019inertial} for quaternion based inertial navigation system, the whole derivations is completely different from it as we parameterize the attitude by direction cosine matrix (DCM). As the whole state is confined to matrix Lie group, it avoids the over-parameter and normalization constraint of the quaternion representation. Meanwhile, we extended the ST-EKF~\cite{wang2018state} by considering the position error with respect to the common frame. 
Then, we investigate the state error dynamic equations on the local geodetic latitude-longitude-height frame, the local-level north-east-down frame, and the ECEF frame.
Furthermore, we give the invariant measurement model involving lever arm error and show the equivalence between the invariant measurement model\cite{barrau2017the} and the traditional EKF's measurement model.
Finally, we derive the two modified error dynamics on the NED frame and the ECEF frame respectively by introducing new auxiliary velocity vectors.

The contributions of the paper can be summarized as follows:
\begin{enumerate}
	\item  We propose an $SE_2(3)$ based EKF framework for inertial-integrated navigation system which embedded the attitude, velocity, and position into the matrix Lie group $SE_2(3)$.
	\item  We give detailed derivations of the $SE_2(3)$ on the NED frame and the ECEF frame according to the error state defined on the Matrix Lie group $SE_2(3)$, including two left-invariant error forms and two right-invariant error forms which is termed as left $SE_2(3)$ based EKF and right $SE_2(3)$ based EKF. It is amazing that all the right $SE_2(3)$ based EKF algorithms remove the specific force term in the transition matrix.
	\item  We design two modified error dynamics on the NED frame and the ECEF frame by introducing new auxiliary velocity vectors.
    \item  We show the equivalence of the left-invariant EKF and the left $SE_2(3)$ based EKF when lever arm are considered in navigation frame and ECEF frame.
\end{enumerate}

This remainder of this paper is organized as follows. Preliminaries are presented in Section 2. 
In section 3 the $SE_2(3)$ based EKF framework is introduced. In section 4 the $SE_2(3)$ based EKF for NED navigation is derived. In section 5 the $SE_2(3)$ based EKF for another NED navigation is derived. Modified error state dynamic equations on NED frame is given in Section 6. Section 7 formulates the $SE_2(3)$ based EKF for ECEF navigation. Section 8 formulates the $SE_2(3)$ based EKF for another ECEF navigation.  Modified error state dynamic equations on ECEF frame is given in Section 9. The equivalence of the invariant EKF algorithms and the left $SE_2(3)$ based EKF algorithm are shown in Section 10. 
The $SE_2(3)$ based smoothing algorithm is introduced in Section 11. Conclusion and future work are given in Section 12.
\section{Preliminaries}
The kinematics of the vehicles are described by the velocity, position and the direction, which are expressed on the manifold space and identified by different frames. The velocity and position can be represented by the vectors and the attitude in the 3-dimensional vector space can be represented by the direction cosine matrix (DCM). 
This three quantities can be reformulated as an element of the $SE_2(3)$ matrix Lie group. 
Meanwhile, the vector $v_{ab}^c$ describes the vector points from point a to point b and expressed in the c frame. The direction cosine matrix $C_d^f$ represents the rotation from the d frame to the f frame. Therefore, we summarize the commonly used frames in the inertial navigation and give detailed navigation equations in both the NED frame and the ECEF frame.
\subsection{The $SE_2(3)$ Matrix Lie Group}
The $SE_2(3)$ matrix Lie group is also called the group of direct spatial isometries~\cite{barrau2017the} and it represents the space of matrices that apply a rigid body rotation and 2 translations to points in $\mathbb{R}^3$. Moreover, the group $SE_2(3)$ has the structure of the semidirect product of SO(3) group by $\mathbb{R}^3\times\mathbb{R}^3$ and can be expressed as $SE_2(3)=SO(3)\ltimes \underbrace{\mathbb{R}^3\times\mathbb{R}^3}_2$~\cite{luo2020geometry}. The relationship between the Lie algebra and the associated vector is described by a linear isomorphism $\Lambda$: $\mathbb{R}^9 \rightarrow \mathfrak{se}_2(3)$, i.e. 
\begin{equation}\label{vector_algebra}
\Lambda(\xi)=\begin{bmatrix}
\phi\times & \vartheta &\zeta\\ 0_{1\times3}&0&0\\0_{1\times3}&0&0
\end{bmatrix}\in \mathfrak{se}_2(3), \forall \xi=\begin{bmatrix}
\phi \\ \vartheta \\ \zeta
\end{bmatrix}\in \mathbb{R}^9,\phi,\vartheta,\zeta\in\mathbb{R}^3
\end{equation}

The exponential mapping from the Lie algebra to the corresponding Lie group is given as
\begin{equation}\label{algebra_group}
T=\exp_G(\Lambda(\xi))=\sum_{n=0}^{\infty}\frac{1}{n!}\left(\Lambda(\xi)\right)^n=\exp_G\left(\begin{bmatrix}
\phi\times & \vartheta &\zeta\\ 0_{1\times3}&0&0\\0_{1\times3}&0&0
\end{bmatrix} \right)=\begin{bmatrix}
\exp_G(\phi\times) & J\vartheta &J\zeta\\ 0_{1\times3}&1&0\\0_{1\times3}&0&1
\end{bmatrix} 
\end{equation}
where $\phi\times$ denotes the skew-symmetric matrix generated from a 3D vector $\phi\in\mathbb{R}^3$; $\exp_G$ denotes the matrix exponential mapping; J is the left Jacobian matrix of  the 3D orthogonal rotation matrices group $SO(3)$ which is given by.
\begin{equation}\label{left_Jacobian}
J=J_l(\phi)=\sum_{n=0}^{\infty}\frac{1}{(n+1)!}(\phi_{\wedge})^n=I_3+\frac{1-\cos\theta}{\theta^2}\phi_{\wedge}+\frac{\theta-\sin\theta}{\theta^3}\phi_{\wedge}^2,\theta=||\phi||
\end{equation}

The closed form expression for $T$ from the exponential map can also be obtained as
\begin{equation}\label{Exponential_map_se_k_3}
\begin{aligned}
T=\sum_{n=0}^{\infty}\frac{1}{n!}\left(\Lambda(\xi)\right)^n
=I_{5\times 5}+\Lambda(\xi)+\frac{1-\cos\theta}{\theta^2}\Lambda(\xi)^2+\frac{\theta-\sin\theta}{\theta^3}\Lambda(\xi)^3
\end{aligned}
\end{equation}

$SE_2(3)$ is common used as the extended poses (orientation, velocity, position) for 3-dimensional inertial navigation. 
\subsection{Uncertainty and Concentrated Gaussian Distribution on Matrix Lie Group $SE_2(3)$} 
The uncertainties on matrix Lie group $SE_2(3)$ can be represented by left multiplication and right multiplication
\begin{equation}\label{uncertainty}
\begin{aligned}
\text{left multiplication}:{T}_l&=\hat{T}\exp_G(\Lambda{(\varepsilon_l)})=\hat{T}\exp_G(\Lambda{(\varepsilon_l)})\hat{T}^{-1}\hat{T}=\exp_G(\Lambda{(Ad_{\hat{T}}(\varepsilon_l))})\hat{T}\\
\text{right multiplication}:{T}_r&=\exp_G(\Lambda{(\varepsilon_r)})\hat{T}
\end{aligned}
\end{equation}

Therefore, the probability distributions for the random variables $T\in SE_2(3)$ can be defined as left-invariant concentrated Gaussian distribution on $SE_2(3)$ and right-invariant concentrated Gaussian distribution on $SE_2(3)$:
\begin{equation}\label{concentrated}
\begin{aligned}
\text{left-invariant}:T\sim \mathcal{N}_L(\hat{T},P),{T}_l&=\hat{T}\exp_G(\Lambda{(\varepsilon_l)}),\varepsilon_l\sim \mathcal{N}(0,P)\\
\text{right-invariant}:T\sim  \mathcal{N}_R(\hat{T},P),{T}_r&=\exp_G(\Lambda{(\varepsilon_r)})\hat{T},\varepsilon_r\sim   \mathcal{N}(0,P)
\end{aligned}
\end{equation}
where $\mathcal{N}(\cdot,\cdot)$ is the classical Gaussian distribution in Euclidean space and $P\in\mathbb{R}^{3(K+1)\times 3(K+1)}$ is a covariance matrix. The invariant property can be verified by $\exp_G(\Lambda{(\varepsilon_r)})=({T}_r \Gamma) (\hat{T}\Gamma)^{-1}={T}_r \hat{T}^{-1}$ and $\exp_G(\Lambda{(\varepsilon_l)})=(\Gamma\hat{T})^{-1}(\Gamma{T}_l ) =\hat{T}^{-1}{T}_l$. The noise-free quantity $\hat{T}$ is viewed as the mean, and the dispersion arises through left multiplication or right multiplication with the matrix exponential of a zero mean Gaussian random variable.
\subsection{Reference Frames}
The commonly used reference frames~\cite{shin2005estimation} in inertial-integrated navigation system are summarized in this section. 

Earth-Centered-Inertial (ECI) Frames (i-frame) is an ideal frame of reference in which ideal accelerometers and gyroscopes fixed to the i-frame have zero outputs and it has its origin at the center of the Earth and axes that are non-rotating with respect to the fixed stars with its z-axis parallel to the spin axis of the Earth, x-axis pointing towards the mean vernal equinox, and y-axis completing a right-handed orthogonal frame.

Earth-Centered-Earth-Fixed (ECEF) Frames (e-frame) has its origin at the center of mass of the Earth and axes that are fixed with respect to the Earth. Its x-axis points towards the mean meridian of Greenwich, z-axis is parallel to the mean spin axis of the Earth, and y-axis completes a right-handed orthogonal frame.

Navigation Frames (n-frame) is a local geodetic frame which has its origin coinciding with that of the sensor frame, with its x-axis pointing towards geodetic north, z-axis orthogonal to the reference ellipsoid pointing down, and y-axis completing a right-handed
orthogonal frame, i.e. the north-east-down (NED) system. The local geodetic coordinate system can be represented by north coordinate X, east coordinate Y and height Z (XYZ, units:m, m, m), or by latitude $\varphi$, longitude $\lambda$ and height $h$ (LLH, unit: rad, rad, m), and longitude and latitude can be converted one-to-one to XY.

Body Frames (b-frame) is an orthogonal axis set which is fixed onto the vehicle and rotate with it, therefore, it is aligned with the roll, pitch and heading axes of a vehicle, i.e. forward-transversal-down.
\subsection{The gravitational vectors in different frames}
The gravitational vector in ECI frame is given as
\begin{equation}\label{gravitational_vector_i}
g_{ib}^i=G_{ib}^i-(\omega_{ie}^i\times)^2r_{eb}^i
\end{equation}
where $g_{ib}^i$ is the gravity vector expressed in ECI frame and it is sometimes referred to as "plumb-bob gravity"~\cite{savage2000strapdown}; $G_{ib}^i$ is the gravitational vector expressed in the ECI frame.

According to equation(\ref{gravitational_vector_i}), the gravitational vector in ECEF frame is given as
\begin{equation}\label{gravitational_vector_e}
g_{ib}^e=C_i^eg_{ib}^i=C_i^eG_{ib}^i-C_i^e(\omega_{ie}^i\times)C_e^iC_i^e(\omega_{ie}^i\times)C_e^iC_i^er_{eb}^i=G_{ib}^e-(\omega_{ie}^e\times)^2r_{eb}^e
\end{equation}
where $g_{ib}^e$ is the gravity vector expressed in ECEF frame; $G_{ib}^e$ is the gravitational vector expressed in the ECEF frame.The perturbation on the gravity $\delta g_{ib}^n$ can be written as~\cite{groves2013principles}
\begin{equation}\label{perturbation_g_ib_e}
\delta g_{ib}^e\triangleq \tilde{g}_{ib}^e-g_{ib}^e\approx -\frac{\mu}{||r_{ib}^e||^3}\delta r_{ib}^e
\end{equation}
where $\mu$ is defined in Chapter2 of~\cite{groves2013principles}.

Similarly, we can get the gravitational vector in navigation frame according to equation(\ref{gravitational_vector_i})
\begin{equation}\label{gravitational_vector_n}
g_{ib}^n=C_i^ng_{ib}^i=C_i^nG_{ib}^i-C_i^n(\omega_{ie}^i\times)C_n^iC_i^n(\omega_{ie}^i\times)C_n^iC_i^nr_{eb}^i=G_{ib}^n-(\omega_{ie}^n\times)^2r_{eb}^n
\end{equation}
where $g_{ib}^n$ is the gravity vector expressed in navigation frame; $G_{ib}^n$ is the gravitational vector expressed in the navigation frame. The perturbation on the gravity $\delta g_{ib}^n$ can be written with a simplified inverse gravity modal as~\cite{shin2005estimation}
\begin{equation}\label{perturbation_gravity}
\delta g_{ib}^n\triangleq\tilde{g}_{ib}^n-g_{ib}^n \approx\begin{bmatrix}
0\\0\\ \frac{2g_{ib}^n}{\sqrt{R_MR_N}+h}\delta r_D
\end{bmatrix}\triangleq L(\delta r_{eb}^n)
\end{equation}
where $\sqrt{R_MR_N}$ is the Gaussian mean Earth radius of curvature; $\delta r_D$ is perturbation of the error position vector $\delta r_{eb}^n$ in the down direction of NED frame.
\subsection{NED Navigation Equations when position is represented in terms of LLH}
The attitude in the NED frame can be represented by the DCM $C_b^n$, 
The differential equation of $C_b^n$ and $C_n^b$ are given by
\begin{equation}\label{C_b_n_d_e}
\dot{C}_b^n=C_b^n(\omega_{ib}^b\times)-(\omega_{in}^n\times)C_b^n
\end{equation}
\begin{equation}\label{C_n_b_d_e}
\dot{C}_n^b=C_n^b(\omega_{in}^n\times)-(\omega_{ib}^b\times)C_n^b
\end{equation}
where $\omega_{ib}^b$ is the angular rate vector of the body frame relative to the inertial frame expressed in the body frame; $\omega_{in}^n$ is the angular rate vector of the navigation frame relative to the inertial frame expressed in the navigation frame.

The differential equation of the velocity vector in the NED local-level navigation frame is given by
\begin{equation}\label{v_eb_n_d_e}
\dot{v}_{eb}^n=C_b^nf_{ib}^b-\left[ (2\omega_{ie}^n+\omega_{en}^n)\times\right]v_{eb}^n+g_{ib}^n
\end{equation}
where $\omega_{ie}^n$ is the earth rotation vector expressed in the navigation frame; $f_{ib}^b$ is the specific force vector in navigation frame; $\omega_{en}^n=\omega_{in}^n-\omega_{ie}^n$ is the angular rate vector of the navigation frame relative to the earth frame expressed in the navigation frame which is also call the transport rate; and $g_{ib}^n$ is the gravity vector.

The differential equation of the velocity error with all parameters expressed in the navigation frame can be written as follows:
\begin{equation}\label{delta_velocity}
\delta \dot{v}_{eb}^n=(C_b^nf_{ib}^b)\times\phi^n+C_b^n\delta f_{ib}^b-(2\omega_{ie}^n+\omega_{en}^n)\times \delta v_{eb}^n-(2\delta\omega_{ie}^n+\delta\omega_{en}^n)\times v_{eb}^n+\delta g_{ib}^n
\end{equation}
where $\delta v_{eb}^n$ and $\phi^n$ are the velocity error vector and attitude error vector, respectively; $\delta\omega_{ie}^n$ and $\delta\omega_{en}^n$ are the angular rate errors corresponding to $\omega_{ie}^n$ and $\omega_{en}^n$, respectively; $\delta g^n$ is the normal gravity error in the local navigation frame.

$\omega_{ie}^n$ and $\omega_{en}^n$ can be given as follows
\begin{equation}\label{omega_ie_n}
\begin{aligned}
&\omega_{ie}^n=\begin{bmatrix}
\omega_{ie}\cos\varphi\\0\\-\omega_{ie}\sin\varphi
\end{bmatrix},\omega_{en}^n=\begin{bmatrix}
\dot{\lambda}\cos\varphi \\-\dot{\varphi} \\ -\dot{\lambda}\sin\varphi
\end{bmatrix}=\begin{bmatrix}
\frac{v_E}{R_N+h}\\ \frac{-v_N}{R_M+h}\\ \frac{-v_E\tan\varphi}{R_N+h}
\end{bmatrix}\\
&\omega_{in}^n=\omega_{ie}^n+\omega_{en}^n=\begin{bmatrix}
\omega_{ie}\cos\varphi+\frac{v_E}{R_N+h}\\ \frac{-v_N}{R_M+h}\\ -\omega_{ie}\sin\varphi -\frac{v_E\tan\varphi}{R_N+h}
\end{bmatrix},2\omega_{ie}^n+\omega_{en}^n=\begin{bmatrix}
2\omega_{ie}\cos\varphi+\frac{v_E}{R_N+h}\\ \frac{-v_N}{R_M+h}\\ -2\omega_{ie}\sin\varphi -\frac{v_E\tan\varphi}{R_N+h}
\end{bmatrix}
\end{aligned}
\end{equation}
where $\omega_{ie}=0.000072921151467rad/s$ is the magnitude of the earth's rotation angular rate; $v_N$ and $v_E$ are velocities in the north and east direction, respectively; $h$ is ellipsoidal height; $R_M$ and $R_N$ are radii of curvature in the meridian and prime vertical; $\dot{\varphi}=\frac{v_N}{R_M+h}$ and $\dot{\lambda}=\frac{v_E}{(R_N+h)\cos\varphi}$ are used in the derivation.

When the position vector is expressed in terms of the geodetic latitude $\varphi$, longitude $\lambda$, and height $h$,  the differential equation of the position vector is given by
\begin{equation}\label{r_eb_l_l_b_h}
\dot{r}_{eb}^l=\begin{bmatrix}
\dot{\varphi} \\ \dot{\lambda} \\ \dot{h}
\end{bmatrix}=\begin{bmatrix}
\frac{1}{R_M+h} & 0&0\\
0&\frac{1}{(R_N+h)\cos\varphi} &0\\
0&0&-1
\end{bmatrix}\begin{bmatrix}
v_N\\v_E\\v_D
\end{bmatrix}=N_{rv}v_{eb}^n
\end{equation}
The corresponding position error differential equation can be written as~\cite{shin2001accuracy}
\begin{equation}\label{position_error_differential_quation_lbh}
\begin{aligned}
\delta \dot{\varphi}&=-\frac{v_N}{(R_M+h)^2}\delta h+\frac{1}{R_M+h}\delta v_N\\
\delta \dot{\lambda}&=\frac{v_E\tan\varphi}{(R_N+h)\cos\varphi}\delta\varphi-\frac{v_E}{(R_N+h)^2\cos\varphi}\delta h+\frac{1}{(R_N+h)\cos\varphi}\delta v_E\\
\delta \dot{h}&=-\delta v_D
\end{aligned}
\end{equation}
where $\delta \varphi$, $\delta\lambda$, and $\delta h$ are the latitude error, the longitude error, and the height error respectively.

Therefore, the matrix form of the position error differential equation in terms of the geodetic latitude, longitude and elevation is
\begin{equation}\label{position_error_differential_equation_matrix_lbh}
\begin{aligned}
\begin{bmatrix}
\delta \dot{\varphi}\\
\delta \dot{\lambda}\\
\delta \dot{h}
\end{bmatrix}&=\begin{bmatrix}
0&0& -\frac{v_N}{(R_M+h)^2}\\
\frac{v_E\tan\varphi}{(R_N+h)\cos\varphi}&0&-\frac{v_E}{(R_N+h)^2\cos\varphi}\\
0&0&0
\end{bmatrix}\begin{bmatrix}
\delta \varphi\\
\delta \lambda\\
\delta h
\end{bmatrix}+\begin{bmatrix}
\frac{1}{R_M+h} & 0&0\\
0& \frac{1}{(R_N+h)\cos\varphi} & 0\\
0&0&-1
\end{bmatrix}\begin{bmatrix}
\delta v_N\\
\delta v_E\\
\delta v_D
\end{bmatrix}\\
&=\delta\dot{r}_{eb}^l=N_{rr}\delta r_{eb}^l+N_{rv}\delta v_{eb}^n
\end{aligned}
\end{equation}

Perturbations on $\omega_{ie}^n$, $\omega_{en}^n$, and $\omega_{in}^n$ can be given as follows
\begin{equation}\label{perturbation_omega_ie_n_LLH}
\delta\omega_{ie}^n=\begin{bmatrix}
\frac{-\omega_{ie}\sin\varphi \delta r_N}{R_M+h} \\0 \\ \frac{-\omega_{ie}\cos\varphi\delta r_N}{R_M+h}
\end{bmatrix}
=\begin{bmatrix}
-\omega_{ie}\sin\varphi \delta \varphi \\0 \\ -\omega_{ie}\cos\varphi\delta \varphi
\end{bmatrix}=\begin{bmatrix}
-\omega_{ie}\sin\varphi  &0&0 \\0&0&0 \\ -\omega_{ie}\cos\varphi
&0&0\end{bmatrix}\begin{bmatrix}
\delta\varphi \\ \delta\lambda \\ \delta h
\end{bmatrix}=N_1\delta r_{eb}^l
\end{equation}
where $\delta\varphi=\frac{\delta r_N}{R_M+h}$ is used.
\begin{equation}\label{perturbation_omega_en_n_LLH}
\begin{aligned}
&\delta\omega_{en}^n=\begin{bmatrix}
\frac{v_E\delta r_D}{(R_N+h)^2}+\frac{\delta v_E}{R_N+h}\\
-\frac{v_N\delta r_D}{(R_M+h)^2}-\frac{\delta v_N}{R_M+h}\\
-\frac{v_E\delta r_N}{(R_N+h)(R_M+h)\cos^2\varphi}-\frac{v_E\tan\varphi\delta r_D}{(R_N+h)^2}-\frac{\tan\varphi \delta v_E}{R_N+h}
\end{bmatrix}\\
=&\begin{bmatrix}
0 & 0 & -\frac{v_E}{(R_N+h)^2}\\
0& 0& \frac{v_N}{(R_M+h)^2}\\
-\frac{v_E}{(R_N+h)\cos^2\varphi} & 0 & \frac{v_E\tan\varphi}{(R_N+h)^2}
\end{bmatrix}\delta r_{eb}^l+\begin{bmatrix}
0& \frac{1}{R_N+h} & 0\\
-\frac{1}{R_M+h} & 0& 0\\
0& -\frac{\tan\varphi}{R_N+h}&0
\end{bmatrix}\delta v_{eb}^n
=N_3\delta r_{eb}^l+N_2\delta v_{eb}^n
\end{aligned}
\end{equation}
where $\delta h=-\delta r_D$ is used.
\begin{equation}\label{perturbation_omega_in_n_LLH}
\delta\omega_{in}^n=\delta\omega_{ie}^n+\delta\omega_{en}^n\\
=N_1\delta r_{eb}^l+N_3\delta r_{eb}^l+N_2\delta v_{eb}^n=(N_1+N_3)\delta r_{eb}^l+N_2\delta v_{eb}^n
\end{equation}

\subsection{NED Navigation Equations when position is represented in terms of XYZ}
The position error state expressed in radians is usually very small, which will cause numerical instability in Kalman filtering calculation. Therefore, it is usually to represent the position error vector in terms of the XYZ coordinate system, that is 
\begin{equation}\label{position_error_NED_frame}
\delta r_{eb}^n=\begin{bmatrix}
\delta r_N \\ \delta r_E \\ \delta r_D
\end{bmatrix}=\begin{bmatrix}
(R_M+h)\delta \varphi \\ (R_N+h)\cos\varphi \delta \lambda \\ -\delta h
\end{bmatrix}
\end{equation}

The differential equation of the position vector in the XYZ local-level navigation frame is given by~\cite{shin2005estimation}
\begin{equation}\label{position_error_NED_frame_differential_equation}
\begin{aligned}
\delta \dot{r}_N&=(\dot{R}_M+\dot{h})\delta \varphi+(R_M+h)\delta \dot{\varphi}\\
\delta \dot{r}_E&=(R_N+h)\cos\varphi\delta\dot{\lambda}-(R_N+h)\sin\varphi\delta\lambda\dot{\varphi}+(\dot{R}_N+\dot{h})\cos\varphi\delta\lambda \\
\delta \dot{r}_D&=-\delta \dot{h}
\end{aligned}
\end{equation}

Substituting position differential equation (\ref{r_eb_l_l_b_h}) and position error differential equation (\ref{position_error_differential_quation_lbh}) into above equation, we can get the matrix form of the the position error differential equation in terms of the NED coordinate system
\begin{equation}\label{NED_position_matrix}
\begin{aligned}
\begin{bmatrix}
\delta \dot{r}_N\\
\delta \dot{r}_E\\
\delta \dot{r}_D
\end{bmatrix}&=\begin{bmatrix}
-\frac{v_D}{(R_M+h)}&0& \frac{v_N}{(R_M+h)}\\
\frac{v_E\tan\varphi}{(R_N+h)}&-\frac{\tan\varphi v_N}{(R_M+h)}-\frac{v_D}{(R_N+h)}&\frac{v_E}{(R_N+h)}\\
0&0&0
\end{bmatrix}\begin{bmatrix}
\delta r_N\\
\delta r_E\\
\delta r_D
\end{bmatrix}+\begin{bmatrix}
\delta v_N\\
\delta v_E\\
\delta v_D
\end{bmatrix}\\
&=\delta\dot{r}_{eb}^n=-\omega_{en}^n\delta r_{eb}^n+\delta\theta\times v_{eb}^n+\delta v_{eb}^n=F_{rr}\delta r_{eb}^n+\delta v_{eb}^n
\end{aligned}
\end{equation}
where $\delta\theta$ is the difference between the computer frame and the true navigation frame~\cite{savage2000strapdown} and can be calculated by
\begin{equation}\label{delta_theta_c_n}
\delta \theta=\begin{bmatrix}
\frac{\delta r_E}{R_N+h}\\
\frac{-\delta r_N}{R_M+h}\\
\frac{-\delta r_E\tan\varphi}{R_N+h}
\end{bmatrix}
\end{equation}

The position vector differential equation in terms of the NED coordinate system can be calculated as
\begin{equation}\label{r_eb_n_d_e}
\dot{r}_{eb}^n=\frac{d}{dt}(C_e^n r_{eb}^e)=\frac{d}{dt}(C_e^n )r_{eb}^e+C_e^n\dot{r}_{eb}^e=C_e^n(\omega_{ne}^e\times)r_{eb}^e+C_e^nv_{eb}^e=-\omega_{en}^n\times r_{eb}^n+v_{eb}^n
\end{equation}

Perturbations on $\omega_{ie}^n$, $\omega_{en}^n$, and $\omega_{in}^n$ can be given as follows
\begin{equation}\label{perturbation_omega_ie_n}
\delta\omega_{ie}^n=\begin{bmatrix}
\frac{-\omega_{ie}\sin\varphi \delta r_N}{R_M+h} \\0 \\ \frac{-\omega_{ie}\cos\varphi\delta r_N}{R_M+h}
\end{bmatrix}=M_1\delta r_{eb}^n
\end{equation}
\begin{equation}\label{perturbation_omega_en_n}
\delta\omega_{en}^n=\begin{bmatrix}
\frac{v_E\delta r_D}{(R_N+h)^2}+\frac{\delta v_E}{R_N+h}\\
-\frac{v_N\delta r_D}{(R_M+h)^2}-\frac{\delta v_N}{R_M+h}\\
-\frac{v_E\delta r_N}{(R_N+h)(R_M+h)\cos^2\varphi}-\frac{v_E\tan\varphi\delta r_D}{(R_N+h)^2}-\frac{\tan\varphi \delta v_E}{R_N+h}
\end{bmatrix}=M_3\delta r_{eb}^n+M_2\delta v_{eb}^n
\end{equation}
\begin{equation}\label{perturbation_omega_in_n}
\delta\omega_{in}^n=\delta\omega_{ie}^n+\delta\omega_{en}^n=M_1\delta r_{eb}^n+M_3\delta r_{eb}^n+M_2\delta v_{eb}^n=(M_1+M_3)\delta r_{eb}^n+M_2\delta v_{eb}^n
\end{equation}

\subsection{Another NED Navigation Equations when position is represented in terms of XYZ}
When the attitude, velocity, and position are represented as $C_b^n$, $v_{ib}^n$, and $r_{ib}^n$, their differential equations are also considered. The differential equation for the attitude $C_b^n$ has been given in equation (\ref{C_b_n_d_e}).
As the velocity has the relationship $v_{ib}^n=C_i^nv_{ib}^i$, the differential equation of the velocity $v_{ib}^n$ can be calculated as
\begin{equation}\label{v_ib_n_new_navigation_frame}
\dot{v}_{ib}^n=\frac{d}{dt}(C_i^n v_{ib}^i)=\dot{C}v_{ib}^i+C_i^n\dot{v}_{ib}^i=(-\omega_{in}^n\times)C_i^n v_{ib}^i+C_i^n(C_b^if_{ib}^b+G_{ib}^i)
=-\omega_{in}^n\times v_{ib}^n+C_b^n f_{ib}^b+G_{ib}^n
\end{equation}

The differential equation of the position vector $r_{ib}^n$ can be given similarly as
\begin{equation}\label{r_ib_n_new_navigation_frame}
\dot{r}_{ib}^n=\frac{d}{dt}(C_i^n r_{ib}^i)=\dot{C}_i^n r_{ib}^i+ C_i^n \dot{r}_{ib}^i=-\omega_{in}^n\times r_{ib}^n+v_{ib}^n
\end{equation} 

\subsection{ECEF Navigation Equations when position is represented in terms of XYZ}
The differential equation of the attitude matrix in the ECEF frame can be represented as
\begin{equation}\label{C_b_e_d_e}
\dot{C}_b^e=C_b^e(\omega_{ib}^b\times)-(\omega_{ie}^e\times)C_b^e
\end{equation}
\begin{equation}\label{C_e_b_d_e}
\dot{C}_e^b=C_e^b(\omega_{ie}^e\times)-(\omega_{ib}^b\times)C_e^b
\end{equation}

The differential equation of the velocity vector in the ECEF frame is given as
\begin{equation}\label{v_eb_e_d_e}
\dot{v}_{eb}^e=C_b^ef_{ib}^b-2\omega_{ie}^e\times v_{eb}^e+g_{ib}^e
\end{equation}

The differential equation of the position vector in the ECEF frame is given as
\begin{equation}\label{r_eb_e_d_e}
\dot{r}_{eb}^e=v_{eb}^e
\end{equation}
\subsection{Another ECEF Navigation Equations with position represented as XYZ}
As the ECEF frame has the same origin as the ECI frame, so $r_{ie}^i=0$ and $r_{ib}^i=r_{ie}^i+r_{eb}^i=r_{eb}^i=C_e^ir_{eb}^e$. Meanwhile, we also get $r_{ib}^e=r_{eb}^e=C_i^er_{eb}^i$.
The differential equation of the attitude $C_b^e$ has been given in equation(\ref{C_b_e_d_e}).
As the velocity has the relationship $v_{ib}^e=C_i^ev_{ib}^i$, so the differential equation of the velocity $v_{ib}^e$ can be calculated as
\begin{equation}\label{ECEF_ground_nonvelocity}
\begin{aligned}
\dot{v}_{ib}^e&=\frac{d}{dt}(C_i^ev_{ib}^i)=\dot{C}_i^ev_{ib}^i+C_i^e\dot{v}_{ib}^i=(-\omega_{ie}^e\times){C}_i^ev_{ib}^i+C_i^e\left(C_b^if_{ib}^b+G_{ib}^i\right)\\
&=(-\omega_{ie}^e\times)v_{ib}^e+C_i^eC_b^if_{ib}^b+C_i^eG_{ib}^i=(-\omega_{ie}^e\times)v_{ib}^e+C_b^ef_{ib}^b+G_{ib}^e
\end{aligned}
\end{equation}
where $G_{ib}^e$ is the gravity acceleration expressed in the ECEF frame.

The differential equation of the position ${r}_{ib}^e$ is given as
\begin{equation}\label{ECEF_ground_nonpostion}
v_{eb}^e=\dot{r}_{eb}^e=\dot{r}_{ib}^e=(-\omega_{ie}^e\times)C_i^er_{ib}^i+C_i^e\dot{r}_{ib}^i=(-\omega_{ie}^e\times)r_{ib}^e+v_{ib}^e
\end{equation}

According to the differential equation of position (\ref{ECEF_ground_nonpostion}) we can know that $v_{ib}^e=v_{eb}^e+(\omega_{ie}^e\times)r_{ib}^e$, so the differential equation of velocity $v_{ib}^e$ can also be deduced as follows:
\begin{equation}\label{ECEF_non_ground_gravity_velocity1}
\dot{v}_{ib}^e=\dot{v}_{eb}^e+(\omega_{ie}^e\times)\dot{r}_{ib}^e
\end{equation}

Substituting equation(\ref{gravitational_vector_e}) into equation(\ref{ECEF_non_ground_gravity_velocity1}) and we can get
\begin{equation}\label{ECEF_non_ground_gravity_velocity2}
\begin{aligned}
\dot{v}_{ib}^e&=C_b^ef_{ib}^b+g_{ib}^e-2(\omega_{ie}^e\times)v_{eb}^e+(\omega_{ie}^e\times)\dot{r}_{eb}^e=C_b^ef_{ib}^b+g_{ib}^e-(\omega_{ie}^e\times)v_{eb}^e\\
&=C_b^ef_{ib}^b+g_{ib}^e-(\omega_{ie}^e\times)((-\omega_{ie}^e\times)r_{ib}^e+v_{ib}^e)\\
&=C_b^ef_{ib}^b+g_{ib}^e+(\omega_{ie}^e\times)^2r_{ib}^e-(\omega_{ie}^e)\times v_{ib}^e=C_b^ef_{ib}^b+G_{ib}^e-(\omega_{ie}^e)\times v_{ib}^e
\end{aligned}
\end{equation}
This result is the same as the equation(\ref{ECEF_ground_nonvelocity}).

In the end, we get different differential equations of the attitude, velocity and the position in the ECEF frame.

\subsection{Sensor Error Modeling}
If the biases, scale factors, and non-orthogonalities of the accelerometers and gyroscopes are considered, then the uncertainty of the sensors can be expressed as~\cite{shin2005estimation}
\begin{equation}\label{uncertainty_sensors}
\begin{aligned}
\delta f_{ib}^b&=b_a+diag(f_{ib}^b)s_a+\Gamma_a\gamma_a\\
\delta \omega_{ib}^b&=b_g+diag(\omega_{ib}^b)s_g+\Gamma_g\gamma_g
\end{aligned}
\end{equation} 
where $b_a$ and $b_g$ are residual biases of the accelerometers and gyroscopes, respectively; $s_a$ and $s_g$ are the scale factors of the accelerometers and gyroscopes, respectively; $\gamma_a$ and $\gamma_g$ are the non-orthogonalities of the accelerometer triad and gyroscope triad, respectively.
$diag(a)$ represents the diagonal matrix form of a 3-dimensional vector $a$. $\Gamma_a$ and $\Gamma_g$ can be found in~\cite{shin2005estimation}.
The random constant, the random walk and the first-order Gauss-Markov models are typically used in modeling the inertial sensor errors~\cite{shin2005estimation}.

The sensor errors of the accelerometers and gyroscopes for consumer-grade inertial measurement unit (IMU) are modeled as one-order Gauss-Markov model:
\begin{equation}\label{accelerometers_bias}
\delta f_{ib}^b=b_a+w_a, \dot{b}_a=-\frac{1}{\tau_a}b_a+w_{b_a}
\end{equation}
\begin{equation}\label{gyroscopes_bias}
\delta \omega_{ib}^b=b_g+w_g, \dot{b}_g=-\frac{1}{\tau_g}b_g+w_{b_g}
\end{equation}
where $w_a$ and $w_g$ are the Gaussian white noises of the accelerometers and gyroscopes, respectively; $w_{b_a}$  and  $w_{b_g}$  are the Gaussian white noises of the accelerometer biases and gyroscope biases, respectively; $\tau_a$ and $\tau_g$ are the correlation times of accelerometer biases and gyroscope biases, respectively.

Of course, the sensor errors of accelerometers and gyroscopes can also be modeled as random constant process for intermediate-grade IMU and the navigation-grade IMU:
\begin{equation}\label{accelerometers_bias_rw}
\delta f_{ib}^b=b_a+w_a, \dot{b}_a=0
\end{equation}
\begin{equation}\label{gyroscopes_bias_rw}
\delta \omega_{ib}^b=b_g+w_g, \dot{b}_g=0
\end{equation}
\section{$SE_2(3)$ based EKF framework for Inertial-integrated Navigation}
As the error can be defined by the multiplication of the element and its inverse on matrix manifold. The error state can be defined in one of four ways:
$\eta=\tilde{\mathcal{X}}\mathcal{X}^{-1}$, $\eta={\mathcal{X}}\tilde{\mathcal{X}}^{-1}$, $\eta=\tilde{\mathcal{X}}^{-1}\mathcal{X}$, and $\eta={\mathcal{X}}^{-1}\tilde{\mathcal{X}}$. The first two error states are left invariant, the last two error states are right invariant. While the first and fourth error state definitions are similar to the error definition in Euclidean space, that is the estimated value minus the true value, and the second and third error state definitions are similar to the error definition in Euclidean space, that is the true value minus the estimated value. 
Different error state definition will lead to different error state dynamical equations. Therefore, we give a $SE_2(3)$ based EKF framework first, then the specific error differential equations is derived according to the different error state definition and different frames. 
There are four kinds of dynamic equations in NED frame and ECEF frame and four error state definitions on the matrix Lie group, and the $SE_2(3)$ based EKF will be adopted to any combination of the frames and error state definitions.

We first define an element and its inverse of the matrix Lie group $SE_2(3)$ as
\begin{equation}\label{element_inverse}
\mathcal{X}=\begin{bmatrix}
C_a^b & v_{ca}^b & r_{ca}^b\\
0_{1\times 3} &1&0\\
0_{1\times 3} & 0&1
\end{bmatrix},\mathcal{X}^{-1}=\begin{bmatrix}
C_b^a & -C_b^av_{ca}^b & -C_b^ar_{ca}^b\\
0_{1\times 3} &1&0\\
0_{1\times 3} & 0&1
\end{bmatrix}
\end{equation}
where $C_a^b$ represents the attitude matrix; $v_{ca}^b$ represents the velocity vectors expressed in $b$ frame, it can be the ECI frame, the ECEF frame or navigation frame in the navigation problem; $r_{ca}^b$ represents the position vectors expressed in $b$ frame. 
Different applications require different frames to represent the attitude, velocity, and position. Consequently, different error state dynamic equations can be derived. Furthermore, we can design different velocity transformation to obtain the transition matrix with nice property, such as eliminating the specific force term in state transition matrix by defining new velocity errors~\cite{scherzinger1994modified}.

As we can see, each of these quantities in the matrix Lie group $SE_2(3)$ has its own physical explanation. This is reasonable and nature to describe the physical quantities in the world in a compact mathematical formula and leverage its property by the mathematical tools.
Therefore, we give the error state in the matrix Lie group $SE_2(3)$ as the multiplication of one element and its inverse, denoted as
\begin{equation}\label{erro_state_definition}
\eta=f(\mathcal{X},\tilde{\mathcal{X}}) 
\end{equation}
where $f(\cdot,\cdot)$ is a mapping defined as $SE_2(3)\times SE_2(3)\rightarrow SE_2(3)$; $\eta$ is also an element of the matrix Lie group $SE_2(3)$ according to the closure property of the group.

As the operation of the matrix Lie group $SE_2(3)$ is the matrix multiplication. $f(\cdot,\cdot)$ can be one of the four error state definitions declared previously. Moreover, no matter whichever error state is defined, both the new velocity and new position terms in the error state $\eta$ take the attitude difference into account and lead to common frame representation. Barrau~\cite{barrau2015non} studied the autonomy of the error state in his dissertation. We will also leverage the group-affine property of the error state dynamic equations. Moreover, this motivates us to design new modified error state dynamic equations on different frames.

On the other side, the Lie algebra can be converted to the matrix Lie group by the matrix exponential mapping.  We can define a vector in Euclidean space and map it to the error state in the matrix Lie group by a linear isomorphism and an exponential mapping, that is
\begin{equation}\label{lie_algebra_lie_group}
\eta=\begin{bmatrix}
\exp_G(\phi\times) & J\rho_{v} & J\rho_{r}\\
0_{1\times 3} &1&0\\
0_{1\times 3} & 0&1
\end{bmatrix}=\exp_G\left(\begin{bmatrix}
\phi\times & \rho_{v} & \rho_{r}\\
0_{1\times 3} &0&0\\
0_{1\times 3} & 0&0
\end{bmatrix} \right)=\exp_G\left(\Lambda\begin{bmatrix}
\phi \\ \rho_{v} \\ \rho_{r}
\end{bmatrix} \right)=\exp_G(\Lambda(\rho))
\end{equation}
where $J$ is the left Jacobian of the $\exp_G(\phi\times)$, $\phi$ is the attitude error state vector defined in the Euclidean space; $\rho_{v}$ is the velocity error state vector defined in the Euclidean space; $\rho_{r}$ is the position error state vector defined in the Euclidean space; $\rho=\begin{bmatrix}
\phi^T& \rho_{v}^T& \rho_{r}^T
\end{bmatrix}^T$ is a 9-dimensional state error vector defined on the Euclidean space that corresponding to the error state $\eta$ which is defined on the matrix Lie group. $J$ can be approximated as $J\approx I_{3\times 3}$ if $||\phi||$ is small enough.

It is obvious the state error can be converted to the Euclidean space by explicit analytical expression by the Lie group and Lie algebra theory.
Then, we can derive the differential equations for the new attitude error state $\phi$, the new velocity error state $J\rho_{v}$, and the new position error state $J\rho_{r}$. If $J\approx I_{3\times 3}$, then $J\rho_{v}\approx \rho_{v}$ and $J\rho_{r}\approx \rho_{r}$ and we get the differential equations for the error state on the Euclidean space.

Once the new error state dynamic equations for the attitude, velocity, and position are obtained, we construct the measurement matrix according to the new error states corresponding to different measurement sensors.
As an example, we only consider the observation model with global navigation satellite system (GNSS) position estimation results as the observation values, but different measurement sensors such as odometry and GNSS velocity observation, can be formulated similarly.
\section{$SE_2(3)$ based EKF for NED Navigation}
\subsection{$SE_2(3)$ based EKF for NED Navigation with navigation Frame Attitude Error}
The velocity vector $v_{eb}^n$, position vector $r_{eb}^n$, and attitude matrix $C_b^n$ can formula as the element of the $SE_2(3)$ matrix Lie group, that is
\begin{equation}\label{matrix_Lie_group_eb_n}
\mathcal{X}=\begin{bmatrix}
C_b^n & v_{eb}^n & r_{eb}^n\\
0_{1\times3} & 1 & 0\\
0_{1\times 3} & 0& 1
\end{bmatrix}\in SE_2(3)
\end{equation}

The inverse of the element can be written as follows
\begin{equation}\label{inverse_matrix_Lie_group_eb_b}
\mathcal{X}^{-1}=\begin{bmatrix}
C_n^b & -C_n^bv_{eb}^n & -C_n^br_{eb}^n\\
0_{1\times 3} &1 &0\\
0_{1\times 3} &0&1
\end{bmatrix}=\begin{bmatrix}
C_n^b & -v_{eb}^b & -r_{eb}^b\\
0_{1\times 3} &1 &0\\
0_{1\times 3} &0&1
\end{bmatrix}\in SE_2(3)
\end{equation}

Therefore, the differential equation of the $\mathcal{X}$ can be calculated as
\begin{equation}\label{differential}
\begin{aligned}
&\frac{d}{dt}\mathcal{X}=f_{u_t}(\mathcal{X})=\frac{d}{dt}\begin{bmatrix}
C_b^n & v_{eb}^n & r_{eb}^n\\
0_{1\times3} & 1 & 0\\
0_{1\times 3} & 0& 1
\end{bmatrix}
=\begin{bmatrix}
\dot{C}_b^n & \dot{v}_{eb}^n & \dot{r}_{eb}^n\\
0_{1\times3} & 0 & 0\\
0_{1\times 3} & 0& 0
\end{bmatrix}\\
=&\begin{bmatrix}
C_b^n(\omega_{ib}^b\times)-(\omega_{in}^n\times)C_b^n & C_b^nf_{ib}^b-\left[ (2\omega_{ie}^n+\omega_{en}^n)\times\right]v_{eb}^n+g_{ib}^n & -\omega_{en}^n\times r_{eb}^n+v_{eb}^n\\
0_{1\times3} & 0 & 0\\
0_{1\times 3} & 0& 0
\end{bmatrix}\\
\triangleq& \mathcal{X}W_1+W_2\mathcal{X}
\end{aligned}
\end{equation}
where $u_t$ is a sequence of inputs; $W_1$ and $W_2$ are denoted as
\begin{equation}\label{W_1_W_2}
W_1=\begin{bmatrix}
\omega_{ib}^b\times & f_{ib}^b & 0\\
0_{1\times3} & 0 & 0\\
0_{1\times 3} & 0& 0
\end{bmatrix},W_2=\begin{bmatrix}
-\omega_{in}^n\times & g_{ib}^n-\omega_{ie}^n\times v_{eb}^n & v_{eb}^n+\omega_{ie}^n\times r_{eb}^n\\
0_{1\times3} & 0 & 0\\
0_{1\times 3} & 0& 0
\end{bmatrix}
\end{equation}

It is easy to verify that the dynamical equation $f_{u_t}(\mathcal{X})$ is group-affine and the group-affine system owns the log-linear property of the corresponding error propagation~\cite{barrau2017the}:
\begin{equation}\label{proof_invariance}
\begin{aligned}
&f_{u_t}(\mathcal{X}_A)\mathcal{X}_B+\mathcal{X}_Af_{u_t}(\mathcal{X}_B)-\mathcal{X}_Af_{u_t}(I_d)\mathcal{X}_B\\
=&(\mathcal{X}_AW_1+W_2\mathcal{X}_A)\mathcal{X}_B+\mathcal{X}_A(\mathcal{X}_BW_1+W_2\mathcal{X}_B)-\mathcal{X}_A(W_1+W_2)\mathcal{X}_B\\
=&\mathcal{X}_A\mathcal{X}_BW_1+W_2\mathcal{X}_A\mathcal{X}_B\triangleq f_{u_t}(\mathcal{X}_A\mathcal{X}_B)
\end{aligned}
\end{equation}

Similar to the state error defined in the Euclidean space, i.e. the difference of the truth minus the estimate, we can define the state error on the matrix Lie group as the group operation that the true state multiplies the inverse of the estimated state, i.e. $\eta=\mathcal{X}\tilde{\mathcal{X}}^{-1}$, this is where all the errors are defined in the navigation frame. It is obvious that the error defined on the matrix Lie group is right invariant by the right action of the matrix Lie group and it is verified by  $\eta=(\mathcal{X}R){(\tilde{\mathcal{X}}R)}^{-1}=\mathcal{X}\tilde{\mathcal{X}}^{-1},\forall R\in SE_2(3)$. Therefor we can define the state error on the matrix Lie group by attitude error component, the velocity error component and the position error component as follows
\begin{equation}\label{state_error_component_attitude_velocity_position}
\eta\triangleq\mathcal{X}\tilde{\mathcal{X}}^{-1}=\begin{bmatrix}
C_b^n\tilde{C}_n^b & v_{eb}^n-C_b^n\tilde{C}_n^b\tilde{v}_{eb}^n & r_{eb}^n-C_b^n\tilde{C}_n^b\tilde{r}_{eb}^n\\
0_{1\times 3} &1 &0\\
0_{1\times 3} &0&1
\end{bmatrix}\triangleq\begin{bmatrix}
\eta^a & \eta^v & \eta^r\\
0_{1\times 3} &1 &0\\
0_{1\times 3} &0&1
\end{bmatrix}\in SE_2(3)
\end{equation}
where $\eta^a$ is the attitude error expressed in the navigation frame; $\eta^v$ is the velocity error expressed in the navigation frame; $\eta^r$ is the position error expressed in the navigation frame. 

According to the matrix exponential mapping from the Lie algebra to the matrix Lie group, the state error can be converted back to the corresponding Lie algebra as follows
\begin{equation}\label{Lie_group_to_Lie_algebra}
\eta\triangleq\begin{bmatrix}
\exp_G(\phi^n\times) & J\rho_v^n & J\rho_r^n\\
0_{1\times 3} &1 &0\\
0_{1\times 3} &0&1
\end{bmatrix}=\exp_G\left( \begin{bmatrix}
\phi^n\times & \rho_v^n & \rho_r^n\\
0_{1\times 3} &0 &0\\
0_{1\times 3} &0&0
\end{bmatrix}\right)=\exp_G\left(\Lambda \begin{bmatrix}
\phi^n \\ \rho_v^n \\ \rho_r^n
\end{bmatrix}\right)=\exp_G\left(\Lambda (\rho) \right)
\end{equation}
where $\phi^n$ is the attitude error expressed in the navigation frame; $\Lambda(\cdot)$ represents a linear isomorphism between the vector space $\mathbb{R}^9$ and the Lie algebra $\mathfrak{se}_2(3)$; $\exp_G$ represents the matrix exponential mapping from the Lie algebra to the Lie group; $\rho=\begin{pmatrix}
(\phi^n)^T & (\rho_v^n)^T &(\rho_r^n)^T
\end{pmatrix}^T$ represents the Lie algebra corresponding to the state error $\eta$; $\exp_G(\phi^n\times)$ is the Rodriguez formula of the rotation vector; $J$ is the left Jacobian matrix of the Rodriguez formula and can be calculated by 
\begin{equation}\label{Rodriguez_n}
\exp_G(\phi^n\times)=\cos\phi I_{3\times3}+\frac{1-\cos\phi}{\phi^2}\phi^n{\phi^n}^T+\frac{\sin\phi}{\phi}(\phi^n\times),\phi=||\phi^n||
\end{equation}
\begin{equation}\label{left_Jacobian_n}
J=\frac{\sin\phi}{\phi}I_{3\times 3}+\frac{1}{\phi^2}(1-\frac{\sin\phi}{\phi})\phi^n{\phi^n}^T+\frac{1-\cos\phi}{\phi^2}(\phi^n\times),\phi=||\phi^n||
\end{equation}

Comparing equation (\ref{state_error_component_attitude_velocity_position}) and equation (\ref{Lie_group_to_Lie_algebra}), we can get 
\begin{equation}\label{error_new_definition_attitude}
\eta^a=C_b^n\tilde{C}_n^b=\exp_G(\phi^n\times)\approx I_{3\times3}+\phi^n\times, \text{if $||\phi^n||$ is small}
\end{equation}
\begin{equation}\label{error_new_definition_velocity}
\eta^v=J\rho_v^n=v_{eb}^n-C_b^n\tilde{C}_n^b\tilde{v}_{eb}^n\approx v_{eb}^n-(I_{3\times3}+\phi^n\times)\tilde{v}_{eb}^n=-\delta v_{eb}^n -\phi^n\times \tilde{v}_{eb}^n=-\delta v_{eb}^n + \tilde{v}_{eb}^n\times \phi^n
\end{equation}
\begin{equation}\label{error_new_definition_position}
\eta^r= J\rho_r^n=r_{eb}^n-C_b^n\tilde{C}_n^b\tilde{r}_{eb}^n\approx r_{eb}^n-(I_{3\times3}+\phi^n\times)\tilde{r}_{eb}^n=-\delta r_{eb}^n -\phi^n\times \tilde{r}_{eb}^n=-\delta r_{eb}^n + \tilde{r}_{eb}^n\times \phi^n
\end{equation}

It is obvious from the above equations we can find the new state error definition is different from the traditional state error definition. The new state error takes into account the magnitude and direction difference of the state vectors in both the true navigation frame and the calculated navigation frame. This is a new perspective of the common frame error definition by a more sophisticated formula that using the difference of two elements form the $SE_2(3)$ matrix Lie group. This is more nature for the navigation system modeling as the state truly evolves on the $SE_2(3)$ matrix Lie group and reasonable for the error definition for all inertial-integrated navigation as orientation variations may lead inconsistent error~\cite{li2018common}.
Meanwhile, it is also worth noting that this new error definition extends the ST-EKF~\cite{wang2018state} by redefining the position error.

Now, we consider the differential equations for the attitude error, velocity error, and position error which can form  a element of the $SE_2(3)$ matrix Lie group. On the one hand, by taking differential of attitude error $\eta^a$ with respect to time, we can get
\begin{equation}\label{attitude_n_d_right}
\begin{aligned}
&\frac{d}{dt}\eta^a=\frac{d}{dt}C_b^n\tilde{C}_n^b=\dot{C}_b^n\tilde{C}_n^b+C_b^n\dot{\tilde{C}}_n^b\\
=&\left( C_b^n(\omega_{ib}^b\times)-(\omega_{in}^n\times)C_b^n\right)\tilde{C}_n^b+C_b^n\left(\tilde{C}_n^b(\tilde{\omega}_{in}^n\times)-(\tilde{\omega}_{ib}^b\times)\tilde{C}_n^b \right)\\
=&C_b^n(\omega_{ib}^b\times)\tilde{C}_n^b-(\omega_{in}^n\times)C_b^n\tilde{C}_n^b+C_b^n\tilde{C}_n^b(\tilde{\omega}_{in}^n\times)-C_b^n(\tilde{\omega}_{ib}^b\times)\tilde{C}_n^b \\
\approx&-(\omega_{in}^n\times)( I_{3\times3}+\phi^n\times)+( I_{3\times3}+\phi^n\times)((\omega_{in}^n+\delta\omega_{in}^n)\times)-C_b^n\left(\tilde{\omega}_{ib}^b-\omega_{ib}^b)\times\right)\tilde{C}_n^b\\
\approx&\delta\omega_{in}^n\times+(\phi^n\times\omega_{in}^n)\times+(\phi^n\times)(\delta\omega_{in}^n\times)-\delta\omega_{ib}^n\times(I+\phi^n\times)\\
\approx &\delta\omega_{in}^n\times+(\phi^n\times\omega_{in}^n)\times-\delta\omega_{ib}^n\times
\end{aligned}
\end{equation}
where the  2-order small quantities $(\phi^n\times)(\delta\omega_{in}^n\times)$ and $(\delta\omega_{ib}^n\times)(\phi^n\times)$ are neglected at the last step; $\delta\omega_{in}^n$ is defined as $\delta\omega_{in}^n\triangleq \tilde{\omega}_{in}^n-\omega_{in}^n$; $\delta\omega_{ib}^b$ is defined as $\delta\omega_{ib}^b\triangleq \tilde{\omega}_{ib}^b-\omega_{ib}^b$.

On the other hand, 
\begin{equation}\label{attitude_n_d_approx}
\frac{d}{dt}\eta^a\approx \frac{d}{dt}(I_{3\times3}+\phi^n\times)=\dot{\phi}^n\times
\end{equation}

Therefore, the state error differential equation for the attitude error can be written as follows
\begin{equation}\label{attitude_n_d}
\dot{\phi}^n=\delta\omega_{in}^n+(\phi^n\times\omega_{in}^n)-\delta\omega_{ib}^n=-\omega_{in}^n\times\phi^n+\delta\omega_{in}^n-C_b^n\delta\omega_{ib}^b
\end{equation}

By taking differential of velocity error $\eta^v$ with respect to time and substituting equation (\ref{v_eb_n_d_e}) into it, we can get
\begin{equation}\label{Velocity_n_d}
\begin{aligned}
&\frac{d}{dt}\eta^v=\frac{d}{dt}\left(v_{eb}^n-C_b^n\tilde{C}_n^b\tilde{v}_{eb}^n\right)=\dot{v}_{eb}^n-\frac{d}{dt}(C_b^n\tilde{C}_n^b)\tilde{v}_{eb}^n-C_b^n\tilde{C}_n^b\dot{\tilde{v}}_{eb}^n\\
=&C_b^nf_{ib}^b-\left[ (2\omega_{ie}^n+\omega_{en}^n)\times\right]v_{eb}^n+g_{ib}^n-C_b^n\tilde{C}_n^b\left[\tilde{C}_b^n\tilde{f}_{ib}^b-\left[ (2\tilde{\omega}_{ie}^n+\tilde{\omega}_{en}^n)\times\right]\tilde{v}_{eb}^n+\tilde{g}_{ib}^n\right]\\
&-\left[C_b^n(\omega_{ib}^b\times)\tilde{C}_n^b-(\omega_{in}^n\times)C_b^n\tilde{C}_n^b+C_b^n\tilde{C}_n^b(\tilde{\omega}_{in}^n\times)-C_b^n(\tilde{\omega}_{ib}^b\times)\tilde{C}_n^b \right]\tilde{v}_{eb}^n\\
=&-C_b^n(\tilde{f}_{ib}^b-f_{ib}^b)-\left[ (2\omega_{ie}^n+\omega_{en}^n)\times\right](C_b^n\tilde{C}_n^b\tilde{v}_{eb}^n+\eta^v)+C_b^n\tilde{C}_n^b\left[ (2\tilde{\omega}_{ie}^n+\tilde{\omega}_{en}^n)\times\right]\tilde{v}_{eb}^n\\
&+(\omega_{in}^n\times)C_b^n\tilde{C}_n^b\tilde{v}_{eb}^n-C_b^n\tilde{C}_n^b(\tilde{\omega}_{in}^n\times)\tilde{v}_{eb}^n
-C_b^n(\omega_{ib}^b\times)\tilde{C}_n^b\tilde{v}_{eb}^n+C_b^n(\tilde{\omega}_{ib}^b\times)\tilde{C}_n^b\tilde{v}_{eb}^n\\&+(g_{ib}^n-C_b^n\tilde{C}_n^b\tilde{g}_{ib}^n)\\
\approx &-C_b^n\delta f_{ib}^b-(\omega_{ie}^n\times)C_b^n\tilde{C}_n^b\tilde{v}_{eb}^n-\left[ (2\omega_{ie}^n+\omega_{en}^n)\times\right]\eta^v+C_b^n\tilde{C}_n^b(\tilde{\omega}_{ie}^n\times)\tilde{v}_{eb}^n\\
&-C_b^n(\omega_{ib}^b-\tilde{\omega}_{ib}^b)C_n^bC_b^n\tilde{C}_n^b\tilde{v}_{eb}^n+(g_{ib}^n-\tilde{g}_{ib}^n)-\phi^n\times \tilde{g}_{ib}^n\\
\approx & -C_b^n\delta f_{ib}^b-\left[ (2\omega_{ie}^n+\omega_{en}^n)\times\right]\eta^v+\delta\omega_{ie}^n\times\tilde{v}_{eb}^n-(\omega_{ie}^n\times )(\phi^n\times) \tilde{v}_{eb}^n+(\phi^n\times)(\omega_{ie}^n\times) \tilde{v}_{eb}^n\\
& +\phi^n\times\delta\omega_{ie}^n\times\tilde{v}_{eb}^n
+(C_b^n\delta\omega_{ib}^b)\times({v}_{eb}^n-\eta^v)+(C_b^n\delta\omega_{ib}^b)\times\phi^n\times\tilde{v}_{eb}^n-\delta g_{ib}^n-\phi^n\times \tilde{g}_{ib}^n\\
= & -C_b^n\delta f_{ib}^b-\left[ (2\omega_{ie}^n+\omega_{en}^n)\times\right]\eta^v+\delta\omega_{ie}^n\times\tilde{v}_{eb}^n-(\omega_{ie}^n\times \phi^n)\times \tilde{v}_{eb}^n +\phi^n\times\delta\omega_{ie}^n\times\tilde{v}_{eb}^n\\
&+(C_b^n\delta\omega_{ib}^b)\times {v}_{eb}^n-(C_b^n\delta\omega_{ib}^b)\times\eta^v-\delta g_{ib}^n-\phi^n\times \tilde{g}_{ib}^n\\
\approx &-C_b^n\delta f_{ib}^b-\left[ (2\omega_{ie}^n+\omega_{en}^n)\times\right]\eta^v-\tilde{v}_{eb}^n\times\delta\omega_{ie}^n+(\tilde{v}_{eb}^n\times)(\omega_{ie}^n\times)\phi^n  \\
&-{v}_{eb}^n\times(C_b^n\delta\omega_{ib}^b)-\delta g^n+\tilde{g}_{ib}^n\times \phi^n
\end{aligned}
\end{equation}
where 2-order small quantities $(C_b^n\delta\omega_{ib}^b)\times\eta^v$ and $\phi^n\times\delta\omega_{ie}^n\times\tilde{v}_{eb}^n$ are neglected at the last step; $\delta f_{ib}^b$ is defined as $\delta f_{ib}^b \triangleq \tilde{f}_{ib}^b- f_{ib}^b$; $\delta\omega_{ie}^n$ is defined as $\delta\omega_{ie}^n\triangleq \tilde{\omega}_{ie}^n-\omega_{ie}^n$; $\delta g_{ib}^n$ is defined as $\delta g_{ib}^n\triangleq \tilde{g}_{ib}^n-g_{ib}^n$ and it can be neglected as the change of $g_{ib}^n$ is quite small for carrier's local navigation.
\begin{remark}
	The perturbation on the gravity can be taken into consideration by equation(\ref{perturbation_g_ib_e}) with ECEF frame and equation(\ref{perturbation_gravity}) with NED frame.
\end{remark}

As we can see that there is no specific force term $f_{ib}^b$ in the relationship between the attitude error term and the velocity error term. The result and merits have been shown in ST-EKF~\cite{wang2018state}.

By taking differential of position error $\eta^r$ with respect to time and substituting equation (\ref{r_eb_n_d_e}) into it, we can get
\begin{equation}\label{Position_n_d}
\begin{aligned}
&\frac{d}{dt}\eta^r= \frac{d}{dt}(r_{eb}^n-C_b^n\tilde{C}_n^b\tilde{r}_{eb}^n)=\dot{r}_{eb}^n-\frac{d}{dt}(C_b^n\tilde{C}_n^b)\tilde{r}_{eb}^n-C_b^n\tilde{C}_n^b\dot{\tilde{r}}_{eb}^n\\
=&(-\omega_{en}^n\times r_{eb}^n+v_{eb}^n)-C_b^n\tilde{C}_n^b(-\tilde{\omega}_{en}^n\times \tilde{r}_{eb}^n+\tilde{v}_{eb}^n)\\
&-\left[C_b^n(\omega_{ib}^b\times)\tilde{C}_n^b-(\omega_{in}^n\times)C_b^n\tilde{C}_n^b+C_b^n\tilde{C}_n^b(\tilde{\omega}_{in}^n\times)-C_b^n(\tilde{\omega}_{ib}^b\times)\tilde{C}_n^b \right]\tilde{r}_{eb}^n\\
=&-(\omega_{en}^n\times)(C_b^n\tilde{C}_n^b\tilde{r}_{eb}^n+\eta^r)+(v_{eb}^n-C_b^n\tilde{C}_n^b\tilde{v}_{eb}^n)+C_b^n\tilde{C}_n^b(\tilde{\omega}_{en}^n\times) \tilde{r}_{eb}^n\\
&+(\omega_{in}^n\times)C_b^n\tilde{C}_n^b\tilde{r}_{eb}^n-C_b^n\tilde{C}_n^b(\tilde{\omega}_{in}^n\times)\tilde{r}_{eb}^n
-C_b^n(\omega_{ib}^b\times)\tilde{C}_n^b\tilde{r}_{eb}^n+C_b^n(\tilde{\omega}_{ib}^b\times)\tilde{C}_n^b\tilde{r}_{eb}^n\\
=&-\omega_{en}^n\times\eta^r+\eta^v+(\omega_{ie}^n\times)C_b^n\tilde{C}_n^b\tilde{r}_{eb}^n-C_b^n\tilde{C}_n^b(\tilde{\omega}_{ie}^n\times)\tilde{r}_{eb}^n+C_b^n(\delta\omega_{ib}^b\times)\tilde{C}_n^b\tilde{r}_{eb}^n\\
\approx &-\omega_{en}^n\times\eta^r+\eta^v+(\omega_{ie}^n\times)(\phi^n\times)\tilde{r}_{eb}^n-(\phi^n\times)(\omega_{ie}^n\times)\tilde{r}_{eb}^n-\delta\omega_{ie}^n\times \tilde{r}_{eb}^n\\
&-\phi^n\times\delta\omega_{ie}^n\times\tilde{r}_{eb}^n+\delta\omega_{ib}^n\times {r}_{eb}^n-\delta\omega_{ib}^n\times\eta^r\\
= &-\omega_{en}^n\times\eta^r+\eta^v+(\omega_{ie}^n\times\phi^n)\times\tilde{r}_{eb}^n-\delta\omega_{ie}^n\times \tilde{r}_{eb}^n-\phi^n\times\delta\omega_{ie}^n\times\tilde{r}_{eb}^n\\
&+\delta\omega_{ib}^n\times {r}_{eb}^n-\delta\omega_{ib}^n\times\eta^r\\
\approx&-\omega_{en}^n\times\eta^r+\eta^v-(\tilde{r}_{eb}^n\times)(\omega_{ie}^n\times)\phi^n+\tilde{r}_{eb}^n\times \delta\omega_{ie}^n -{r}_{eb}^n\times(C_b^n\delta\omega_{ib}^b)
\end{aligned}
\end{equation}
where 2-order small quantities $(C_b^n\delta\omega_{ib}^b)\times\eta^r$ and $\phi^n\times\delta\omega_{ie}^n\times\tilde{r}_{eb}^n$ are neglected at the last step.

With the new definition of the attitude error, velocity error, and position error, we substitute equation (\ref{error_new_definition_velocity}) and equation (\ref{error_new_definition_position}) into equation (\ref{perturbation_omega_en_n}) and equation (\ref{perturbation_omega_in_n}):
\begin{equation}\label{perturbation_omega_ie_n_new}
\delta\omega_{ie}^n=M_1\delta r_{eb}^n=-M_1(\eta^r-\tilde{r}_{eb}^n\times \phi^n)
\end{equation}
\begin{equation}\label{perturbation_omega_in_n_new}
\delta\omega_{in}^n=(M_1+M_3)\delta r_{eb}^n+M_2\delta v_{eb}^n=-(M_1+M_3)(\eta^r-\tilde{r}_{eb}^n\times \phi^n)-M_2(\eta^v-\tilde{v}_{eb}^n\times \phi^n)
\end{equation}

Consequently, the state error dynamical equations with respect to the navigation frame can be written as follows
\begin{equation}\label{attitude_n_d_new}
\begin{aligned}
&\dot{\phi}^n=-\omega_{in}^n\times\phi^n-(M_1+M_3)(\eta^r-\tilde{r}_{eb}^n\times \phi^n)-M_2(\eta^v-\tilde{v}_{eb}^n\times \phi^n)-C_b^n(b_g+w_g)\\
=&-(M_1+M_3)\eta^r-M_2\eta^v-\left((\omega_{in}^n\times)-M_2(\tilde{v}_{eb}^n\times)-(M_1+M_3)(\tilde{r}_{eb}^n\times)\right)\phi^n-C_b^n(b_g+w_g)
\end{aligned}
\end{equation}
 \begin{equation}\label{Velocity_n_d_new}
 \begin{aligned}
 \frac{d}{dt}\eta^v=
  &-C_b^n(b_a+w_a)-\left[ (2\omega_{ie}^n+\omega_{en}^n)\times\right]\eta^v+(\tilde{v}_{eb}^n\times)M_1(\eta^r-\tilde{r}_{eb}^n\times \phi^n)\\
  &+(\tilde{v}_{eb}^n\times)(\omega_{ie}^n\times)\phi^n  
 -\tilde{v}_{eb}^n\times(C_b^n(b_g+w_g))+\tilde{g}_{ib}^n\times \phi^n\\
 =&(\tilde{v}_{eb}^n\times)M_1\eta^r-\left[ (2\omega_{ie}^n+\omega_{en}^n)\times\right]\eta^v-\tilde{v}_{eb}^n\times(C_b^n(b_g+w_g))\\
 &+\left(-(\tilde{v}_{eb}^n\times)M_1(\tilde{r}_{eb}^n\times) +(\tilde{v}_{eb}^n\times)(\omega_{ie}^n\times)+(\tilde{g}_{ib}^n\times) \right)\phi^n-C_b^n(b_a+w_a)
 \end{aligned}
 \end{equation}
\begin{equation}\label{Position_n_d_new}
\begin{aligned}
\frac{d}{dt}\eta^r=&-\omega_{en}^n\times\eta^r+\eta^v-(\tilde{r}_{eb}^n\times)(\omega_{ie}^n\times)\phi^n\\
&-\tilde{r}_{eb}^n\times M_1(\eta^r-\tilde{r}_{eb}^n\times \phi^n) -\tilde{r}_{eb}^n\times(C_b^n(b_g+w_g))\\
=&-(\tilde{r}_{eb}^n\times M_1+(\omega_{en}^n\times))\eta^r+\eta^v-((\tilde{r}_{eb}^n\times)(\omega_{ie}^n\times)-(\tilde{r}_{eb}^n\times) M_1(\tilde{r}_{eb}^n\times))\phi^n\\
&-\tilde{r}_{eb}^n\times(C_b^n(b_g+w_g))
\end{aligned}
\end{equation}

If the position is represented in terms of LLH, $\delta\omega_{ie}^n$ and $\delta\omega_{in}^n$ can be calculated as
\begin{equation}\label{perturbation_omega_ie_n_new_LLH}
\delta\omega_{ie}^n=N_1\delta r_{eb}^l=N_1(\eta^r-\tilde{r}_{eb}^l\times \phi^n)
\end{equation}
\begin{equation}\label{perturbation_omega_in_n_new_LLH}
\delta\omega_{in}^n=(N_1+N_3)\delta r_{eb}^l+N_2\delta v_{eb}^n=(N_1+N_3)(\eta^r-\tilde{r}_{eb}^l\times \phi^n)+N_2(\eta^v-\tilde{v}_{eb}^n\times \phi^n)
\end{equation}

Then, the new differential equations of attitude error, velocity error and position error can be calculated as
\begin{equation}\label{attitude_n_d_new_LLH}
\begin{aligned}
&\dot{\phi}^n=-\omega_{in}^n\times\phi^n+(N_1+N_3)(\eta^r-\tilde{r}_{eb}^l\times \phi^n)+N_2(\eta^v-\tilde{v}_{eb}^n\times \phi^n)-C_b^n(b_g+w_g)\\
=&(N_1+N_3)\eta^r+N_2\eta^v-\left((\omega_{in}^n\times)+N_2(\tilde{v}_{eb}^n\times)+(N_1+N_3)(\tilde{r}_{eb}^l\times)\right)\phi^n-C_b^n(b_g+w_g)
\end{aligned}
\end{equation}
\begin{equation}\label{Velocity_n_d_new_LLH}
\begin{aligned}
\frac{d}{dt}\eta^v=
&-C_b^n(b_a+w_a)-\left[ (2\omega_{ie}^n+\omega_{en}^n)\times\right]\eta^v-(\tilde{v}_{eb}^n\times)N_1(\eta^r-\tilde{r}_{eb}^l\times \phi^n)\\
&+(\tilde{v}_{eb}^n\times)(\omega_{ie}^n\times)\phi^n  
-\tilde{v}_{eb}^n\times(C_b^n(b_g+w_g))+\tilde{g}_{ib}^n\times \phi^n\\
=&-(\tilde{v}_{eb}^n\times)N_1\eta^r-\left[ (2\omega_{ie}^n+\omega_{en}^n)\times\right]\eta^v-\tilde{v}_{eb}^n\times(C_b^n(b_g+w_g))\\
&+\left((\tilde{v}_{eb}^n\times)N_1(\tilde{r}_{eb}^l\times) +(\tilde{v}_{eb}^n\times)(\omega_{ie}^n\times)+(\tilde{g}_{ib}^n\times) \right)\phi^n-C_b^n(b_a+w_a)
\end{aligned}
\end{equation}
\begin{equation}\label{Position_n_d_new_LLH}
\begin{aligned}
&\frac{d}{dt}\eta^r= \frac{d}{dt}(r_{eb}^l-C_b^n\tilde{C}_n^b\tilde{r}_{eb}^l)\approx \frac{d}{dt}(r_{eb}^l-\tilde{r}_{eb}^l-\phi^n\times\tilde{r}_{eb}^l)=\frac{d}{dt}(\delta r_{eb}^l+\tilde{r}_{eb}^l\times \phi^n)\\
=&(N_{rr}\delta r_{eb}^l+N_{rv}\delta v_{eb}^n)+((\tilde{N}_{rv}\tilde{v}_{eb}^n)\times)\phi^n+(\tilde{r}_{eb}^l\times)(-\omega_{in}^n\times\phi^n+\delta\omega_{in}^n-C_b^n\delta\omega_{ib}^b)\\
=&\left(N_{rr}(\eta^r-\tilde{r}_{eb}^l\times \phi^n)+N_{rv}(\eta^v-\tilde{v}_{eb}^n\times \phi^n)\right)+\left((\tilde{N}_{rv}\tilde{v}_{eb}^n)\times\right)\phi^n\\
&+(\tilde{r}_{eb}^l\times)(N_1+N_3)\eta^r+(\tilde{r}_{eb}^l\times)N_2\eta^v\\
&-(\tilde{r}_{eb}^l\times)\left((\omega_{in}^n\times)+N_2(\tilde{v}_{eb}^n\times)+(N_1+N_3)(\tilde{r}_{eb}^l\times)\right)\phi^n-(\tilde{r}_{eb}^l\times)C_b^n(b_g+w_g)\\
=&\left(N_{rr}+(\tilde{r}_{eb}^l\times)(N_1+N_3)\right)\eta^r+\left(N_{rv}+(\tilde{r}_{eb}^l\times)N_2\right)\eta^v\\
&-\left(N_{rr}(\tilde{r}_{eb}^l\times)+N_{rv}(\tilde{v}_{eb}^n\times)-\left((\tilde{N}_{rv}\tilde{v}_{eb}^n)\times\right)+(\tilde{r}_{eb}^l\times)(\omega_{in}^n\times)+(\tilde{r}_{eb}^l\times)N_2(\tilde{v}_{eb}^n\times)\right.\\
&\left. +(\tilde{r}_{eb}^l\times)(N_1+N_3)(\tilde{r}_{eb}^l\times)\right)\phi^n-(\tilde{r}_{eb}^l\times)C_b^n(b_g+w_g)
\end{aligned}
\end{equation}

\subsection{$SE_2(3)$ based EKF for NED Navigation with estimated navigation frame Attitude Error}
If we define the state error as the group operator that the estimate multiplies the truth, then   $\varepsilon=(\tilde{\mathcal{X}}R){({\mathcal{X}}R)}^{-1}=\tilde{\mathcal{X}}{\mathcal{X}}^{-1},\forall R\in SE_2(3)$.  
This is where all the errors are defined in the estimated navigation frame.

\begin{equation}\label{state_error_component_attitude_velocity_position_estimated_navigation_frame}
\varepsilon\triangleq\tilde{\mathcal{X}}\mathcal{X}^{-1}=\begin{bmatrix}
\tilde{C}_b^n C_n^b & \tilde{v}_{eb}^n-\tilde{C}_b^n C_n^b v_{eb}^n & \tilde{r}_{eb}^n-\tilde{C}_b^n C_n^b r_{eb}^n\\
0_{1\times 3} &1 &0\\
0_{1\times 3} &0&1
\end{bmatrix}\triangleq\begin{bmatrix}
\varepsilon^a & \varepsilon^v & \varepsilon^r\\
0_{1\times 3} &1 &0\\
0_{1\times 3} &0&1
\end{bmatrix}\in SE_2(3)
\end{equation}

where $\varepsilon^a$ is the attitude error expressed in the estimated navigation frame; $\varepsilon^v$ is the velocity error expressed in the estimated navigation frame; $\varepsilon^r$ is the position error expressed in the estimated navigation frame. 

According to the matrix exponential mapping from the Lie algebra to the matrix Lie group, the state error can be converted back to the corresponding Lie algebra as follows
\begin{equation}\label{Lie_group_to_Lie_algebra_estimated_navigation_frame}
\varepsilon\triangleq\begin{bmatrix}
\exp_G(\phi^{\tilde{n}}\times) & J\rho_v^{\tilde{n}} & J\rho_r^{\tilde{n}}\\
0_{1\times 3} &1 &0\\
0_{1\times 3} &0&1
\end{bmatrix}=\exp_G\left( \begin{bmatrix}
\phi^{\tilde{n}}\times & \rho_v^{\tilde{n}} & \rho_r^{\tilde{n}}\\
0_{1\times 3} &0 &0\\
0_{1\times 3} &0&0
\end{bmatrix}\right)=\exp_G\left(\Lambda \begin{bmatrix}
\phi^{\tilde{n}} \\ \rho_v^{\tilde{n}} \\ \rho_r^{\tilde{n}}
\end{bmatrix}\right)=\exp_G\left(\Lambda (\tilde{\rho}) \right)
\end{equation}
where $\phi^{\tilde{n}}$ is the attitude error expressed in the estimated navigation frame; 
$\tilde{\rho}=\begin{pmatrix}
(\phi^{\tilde{n}})^T & (\rho_v^{\tilde{n}})^T &(\rho_r^{\tilde{n}})^T
\end{pmatrix}^T$ represents the Lie algebra corresponding to the state error $\varepsilon$.

Comparing equation (\ref{state_error_component_attitude_velocity_position_estimated_navigation_frame}) with equation (\ref{Lie_group_to_Lie_algebra_estimated_navigation_frame}), we can get 
\begin{equation}\label{error_new_definition_attitude_estimated_navigation_frame}
\varepsilon^a=\exp_G(\phi^{\tilde{n}}\times)=\tilde{C}_b^n C_n^b\approx I_{3\times3}+\phi^{\tilde{n}}\times, \text{if $||\phi^{\tilde{n}}||$ is small}
\end{equation}
\begin{equation}\label{error_new_definition_velocity_estimated_navigation_frame}
\varepsilon^v=J\rho_v^{\tilde{n}}=\tilde{v}_{eb}^n-\tilde{C}_b^n C_n^b v_{eb}^n\approx \tilde{v}_{eb}^n-(I_{3\times3}+\phi^{\tilde{n}}\times){v}_{eb}^n=\delta v_{eb}^n -\phi^{\tilde{n}}\times {v}_{eb}^n=\delta v_{eb}^n + {v}_{eb}^n\times \phi^{\tilde{n}}
\end{equation}
\begin{equation}\label{error_new_definition_position_estimated_navigation_frame}
\varepsilon^r= J\rho_r^{\tilde{n}}=\tilde{r}_{eb}^n-\tilde{C}_b^n C_n^b r_{eb}^n\approx \tilde{r}_{eb}^n-(I_{3\times3}+\phi^{\tilde{n}}\times){r}_{eb}^n=\delta r_{eb}^n -\phi^{\tilde{n}}\times {r}_{eb}^n=\delta r_{eb}^n + {r}_{eb}^n\times \phi^{\tilde{n}}
\end{equation}

Now, we consider the differential equations for the attitude error, velocity error, and position error which can form  a element of the $SE_2(3)$ matrix Lie group. On the one hand, by taking differential of attitude error $\varepsilon^r$ with respect to time, we can get
\begin{equation}\label{attitude_n_d_right_estimated_navigation_frame}
\begin{aligned}
&\frac{d}{dt}\varepsilon^a=\frac{d}{dt}\tilde{C}_b^n C_n^b=\dot{\tilde{C}}_b^nC_n^b+\tilde{C}_b^n\dot{C}_n^b\\
=&\left( \tilde{C}_b^n(\tilde{\omega}_{ib}^b\times)-(\tilde{\omega}_{in}^n\times)\tilde{C}_b^n\right)C_n^b+\tilde{C}_b^n\left(C_n^b(\omega_{in}^n\times)-(\omega_{ib}^b\times)C_n^b \right)\\
=& \tilde{C}_b^n(\tilde{\omega}_{ib}^b\times)C_n^b-(\tilde{\omega}_{in}^n\times)\tilde{C}_b^nC_n^b+\tilde{C}_b^nC_n^b(\omega_{in}^n\times)-\tilde{C}_b^n(\omega_{ib}^b\times)C_n^b  \\
\approx&\tilde{C}_b^n((\tilde{\omega}_{ib}^b-\omega_{ib}^b)\times)C_n^b-((\omega_{in}^n+\delta\omega_{in}^n)\times)( I_{3\times3}+\phi^n\times)+( I_{3\times3}+\phi^n\times)(\omega_{in}^n\times)\\
\approx&(I+\phi^n\times)(\delta\omega_{ib}^n\times)-(\delta\omega_{in}^n\times)+(\phi^n\times\omega_{in}^n)\times-(\delta\omega_{in}^n\times)(\phi^n\times)\\
\approx &-\delta\omega_{in}^n\times+(\phi^n\times\omega_{in}^n)\times+\delta\omega_{ib}^n\times
\end{aligned}
\end{equation}
where the  2-order small quantities $(\delta\omega_{in}^n\times)(\phi^n\times)$ and $(\phi^n\times)(\delta\omega_{ib}^n\times)$ are neglected at the last step; $\delta\omega_{in}^n$ is defined as $\delta\omega_{in}^n\triangleq \tilde{\omega}_{in}^n-\omega_{in}^n$; $\delta\omega_{ib}^b$ is defined as $\delta\omega_{ib}^b\triangleq \tilde{\omega}_{ib}^b-\omega_{ib}^b$.

On the other hand, 
\begin{equation}\label{attitude_n_d_approx_estimated_navigation_frame}
\frac{d}{dt}\eta^a\approx \frac{d}{dt}(I_{3\times3}+\phi^n\times)=\dot{\phi}^n\times
\end{equation}

Therefore, the state error differential equation for the attitude error can be written as follows
\begin{equation}\label{attitude_n_d_estimated_navigation_frame}
\dot{\phi}^n=-\delta\omega_{in}^n+(\phi^n\times\omega_{in}^n)+\delta\omega_{ib}^n=-\omega_{in}^n\times\phi^n-\delta\omega_{in}^n+C_b^n\delta\omega_{ib}^b
\end{equation}

By taking differential of velocity error $\eta^v$ with respect to time and substituting equation (\ref{error_new_definition_velocity_estimated_navigation_frame}) into it, we can get
\begin{equation}\label{Velocity_n_d_estimated_navigation_frame}
\begin{aligned}
&\frac{d}{dt}\varepsilon^v=\frac{d}{dt}\left(\tilde{v}_{eb}^n-\tilde{C}_b^n C_n^b v_{eb}^n\right)=\dot{\tilde{v}}_{eb}^n-\frac{d}{dt}(\tilde{C}_b^n C_n^b)v_{eb}^n-\tilde{C}_b^n C_n^b\dot{v}_{eb}^n\\
=&\left[\tilde{C}_b^n\tilde{f}_{ib}^b-\left[(2\tilde{\omega}_{ie}^n+\tilde{\omega}_{en}^n)\times\right]\tilde{v}_{eb}^n+\tilde{g}_{ib}^n\right]
-\tilde{C}_b^n C_n^b\left[C_b^nf_{ib}^b-\left[ (2\omega_{ie}^n+\omega_{en}^n)\times\right]v_{eb}^n+g_{ib}^n\right]\\
&-\left[\tilde{C}_b^n(\tilde{\omega}_{ib}^b\times)C_n^b-(\tilde{\omega}_{in}^n\times)\tilde{C}_b^nC_n^b+\tilde{C}_b^nC_n^b(\omega_{in}^n\times)-\tilde{C}_b^n(\omega_{ib}^b\times)C_n^b \right]v_{eb}^n\\
=&\tilde{C}_b^n(\tilde{f}_{ib}^b-f_{ib}^b)-\left[ (2\tilde{\omega}_{ie}^n+\tilde{\omega}_{en}^n)\times\right](\varepsilon^v+\tilde{C}_b^nC_n^b v_{eb}^n)+\tilde{C}_b^nC_n^b\left[ (2\omega_{ie}^n+\omega_{en}^n)\times\right]v_{eb}^n\\
&-\tilde{C}_b^n(\tilde{\omega}_{ib}^b\times)C_n^bv_{eb}^n+(\tilde{\omega}_{in}^n\times)\tilde{C}_b^nC_n^bv_{eb}^n-\tilde{C}_b^nC_n^b(\omega_{in}^n\times)v_{eb}^n+\tilde{C}_b^n(\omega_{ib}^b\times)C_n^bv_{eb}^n \\
&+(\tilde{g}_{ib}^n-\tilde{C}_b^n C_n^bg_{ib}^n)\\
\approx &\tilde{C}_b^n\delta f_{ib}^b-(\tilde{\omega}_{ie}^n\times)\tilde{C}_b^nC_b^nv_{eb}^n-\left[ (2\tilde{\omega}_{ie}^n+\tilde{\omega}_{en}^n)\times\right]\varepsilon^v+\tilde{C}_b^nC_n^b(\omega_{ie}^n\times)v_{eb}^n\\
&-\tilde{C}_b^nC_n^bC_b^n(\tilde{\omega}_{ib}^b-\omega_{ib}^b)C_n^bv_{eb}^n+(\tilde{g}_{ib}^n-g_{ib}^n)-\phi^n\times g_{ib}^n\\
\approx & C_b^n\delta f_{ib}^b-\left[ (2\tilde{\omega}_{ie}^n+\tilde{\omega}_{en}^n)\times\right]\varepsilon^v-\delta\omega_{ie}^n\times v_{eb}^n-(\omega_{ie}^n\times )(\phi^n\times) v_{eb}^n+(\phi^n\times)(\omega_{ie}^n\times) v_{eb}^n\\
& -\delta\omega_{ie}^n\times\phi^n\times v_{eb}^n
-(C_b^n\delta\omega_{ib}^b)\times v_{eb}^n-\phi^n\times(C_b^n\delta\omega_{ib}^b)\times v_{eb}^n+\delta g_{ib}^n+ g_{ib}^n\times \phi^n\\
= &\tilde{C}_b^n\delta f_{ib}^b-\left[ (2\tilde{\omega}_{ie}^n+\tilde{\omega}_{en}^n)\times\right]\varepsilon^v-\delta\omega_{ie}^n\times v_{eb}^n-(\omega_{ie}^n\times \phi^n)\times v_{eb}^n -\delta\omega_{ie}^n\times\phi^n\times v_{eb}^n\\
&-(C_b^n\delta\omega_{ib}^b)\times v_{eb}^n-\phi^n\times(C_b^n\delta\omega_{ib}^b)\times v_{eb}^n+\delta g_{ib}^n+ g_{ib}^n\times \phi^n\\
\approx &\tilde{C}_b^n\delta f_{ib}^b-\left[ (2\tilde{\omega}_{ie}^n+\tilde{\omega}_{en}^n)\times\right]\varepsilon^v+v_{eb}^n\times\delta\omega_{ie}^n+(v_{eb}^n\times)(\omega_{ie}^n\times)\phi^n  \\
&+(v_{eb}^n\times)C_b^n\delta\omega_{ib}^b+\delta g_{ib}^n+ g_{ib}^n\times \phi^n
\end{aligned}
\end{equation}
where 2-order small quantities $\phi^n\times(C_b^n\delta\omega_{ib}^b)\times v_{eb}^n$ and $\omega_{ie}^n\times\phi^n\times\delta\tilde{v}_{eb}^n$ are neglected at the last step; $\delta f_{ib}^b$ is defined as $\delta f_{ib}^b \triangleq \tilde{f}_{ib}^b- f_{ib}^b$; $\delta\omega_{ie}^n$ is defined as $\delta\omega_{ie}^n\triangleq \tilde{\omega}_{ie}^n-\omega_{ie}^n$; $\delta g^n$ is defined as $\delta g_{ib}^n\triangleq \tilde{g}_{ib}^n-g_{ib}^n$ and it can be neglected as the change of $g_{ib}^n$ is quite small for carrier's local navigation.

By taking differential of position error $\eta^r$ with respect to time and substituting equation (\ref{error_new_definition_position_estimated_navigation_frame}) into it, we can get
\begin{equation}\label{Position_n_d_estimated_navigation_frame}
\begin{aligned}
&\frac{d}{dt}\varepsilon^r= \frac{d}{dt}(\tilde{r}_{eb}^n-\tilde{C}_b^n C_n^b r_{eb}^n)=\dot{\tilde{r}}_{eb}^n-\frac{d}{dt}(\tilde{C}_b^nC_n^b)r_{eb}^n-\tilde{C}_b^nC_n^b\dot{r}_{eb}^n\\
=&(-\tilde{\omega}_{en}^n\times \tilde{r}_{eb}^n+\tilde{v}_{eb}^n)-\tilde{C}_b^nC_n^b(-\omega_{en}^n\times r_{eb}^n+v_{eb}^n)\\
&-\left[\tilde{C}_b^n(\tilde{\omega}_{ib}^b\times)C_n^b-(\tilde{\omega}_{in}^n\times)\tilde{C}_b^nC_n^b+\tilde{C}_b^nC_n^b(\omega_{in}^n\times)-\tilde{C}_b^n(\omega_{ib}^b\times)C_n^b \right]r_{eb}^n\\
=&-(\tilde{\omega}_{en}^n\times)(\tilde{C}_b^nC_n^br_{eb}^n+\varepsilon^r)+(\tilde{v}_{eb}^n-\tilde{C}_b^nC_n^bv_{eb}^n)+\tilde{C}_b^nC_n^b(\omega_{en}^n\times) r_{eb}^n\\
&+(\tilde{\omega}_{in}^n\times)\tilde{C}_b^nC_n^br_{eb}^n-\tilde{C}_b^nC_n^b(\omega_{in}^n\times)r_{eb}^n
-\tilde{C}_b^n(\tilde{\omega}_{ib}^b\times)C_n^br_{eb}^n+\tilde{C}_b^n(\omega_{ib}^b\times)C_n^br_{eb}^n\\
=&-\tilde{\omega}_{en}^n\times\varepsilon^r+\varepsilon^v+(\tilde{\omega}_{ie}^n\times)\tilde{C}_b^nC_n^br_{eb}^n-\tilde{C}_b^nC_n^b(\omega_{ie}^n\times)r_{eb}^n-\tilde{C}_b^n(\delta\omega_{ib}^b\times)C_n^br_{eb}^n\\
\approx &-\tilde{\omega}_{en}^n\times\varepsilon^r+\varepsilon^v+(\omega_{ie}^n\times)(\phi^n\times)r_{eb}^n-(\phi^n\times)(\omega_{ie}^n\times)r_{eb}^n+\delta\omega_{ie}^n\times r_{eb}^n\\
&+\delta\omega_{ie}^n\times\phi^n\times r_{eb}^n-\delta\omega_{ib}^n\times r_{eb}^n-\phi^n\times\delta\omega_{ib}^n\times r_{eb}^n\\
= &-\tilde{\omega}_{en}^n\times\varepsilon^r+\varepsilon^v+(\omega_{ie}^n\times\phi^n)\times r_{eb}^n+\delta\omega_{ie}^n\times r_{eb}^n\\
&+\delta\omega_{ie}^n\times\phi^n\times r_{eb}^n-\delta\omega_{ib}^n\times r_{eb}^n-\phi^n\times\delta\omega_{ib}^n\times r_{eb}^n\\
\approx&-\tilde{\omega}_{en}^n\times\varepsilon^r+\varepsilon^v-(r_{eb}^n\times)(\omega_{ie}^n\times)\phi^n-r_{eb}^n\times \delta\omega_{ie}^n +r_{eb}^n\times(C_b^n\delta\omega_{ib}^b)
\end{aligned}
\end{equation}
where 2-order small quantities $\phi^n\times(C_b^n\delta\omega_{ib}^b)\times r_{eb}^n$ and $\delta\omega_{ie}^n\times\phi^n\times r_{eb}^n$ are neglected at the last step.

Similar to equation (\ref{perturbation_omega_ie_n_new}) and equation (\ref{perturbation_omega_in_n_new}), we can get
\begin{equation}\label{perturbation_omega_ie_n_new_estimated_navigation_frame}
\delta\omega_{ie}^n=M_1\delta r_{eb}^n=M_1(\varepsilon^r-r_{eb}^n\times \phi^n)
\end{equation}
\begin{equation}\label{perturbation_omega_in_n_new_estimated_navigation_frame}
\delta\omega_{in}^n=(M_1+M_3)\delta r_{eb}^n+M_2\delta v_{eb}^n=(M_1+M_3)(\varepsilon^r-r_{eb}^n\times \phi^n)+M_2(\varepsilon^v-v_{eb}^n\times \phi^n)
\end{equation}

Consequently, the state error dynamical equations with respect to the estimated navigation frame can be written as follows
\begin{equation}\label{attitude_n_d_new_estimated_navigation_frame}
\begin{aligned}
&\dot{\phi}^n=-\omega_{in}^n\times\phi^n-(M_1+M_3)(\varepsilon^r-r_{eb}^n\times \phi^n)-M_2(\varepsilon^v-v_{eb}^n\times \phi^n)+C_b^n(b_g+w_g)\\
=&-(M_1+M_3)\varepsilon^r-M_2\varepsilon^v-\left((\omega_{in}^n\times)-M_2(v_{eb}^n\times)-(M_1+M_3)(r_{eb}^n\times)\right)\phi^n+C_b^n(b_g+w_g)
\end{aligned}
\end{equation}
\begin{equation}\label{Velocity_n_d_new_estimated_navigation_frame}
\begin{aligned}
\frac{d}{dt}\varepsilon^v=
&C_b^n(b_a+w_a)-\left[ (2\tilde{\omega}_{ie}^n+\tilde{\omega}_{en}^n)\times\right]\varepsilon^v+(v_{eb}^n\times)M_1(\eta^r-r_{eb}^n\times \phi^n)\\
&+(v_{eb}^n\times)(\omega_{ie}^n\times)\phi^n  
+v_{eb}^n\times(C_b^n(b_g+w_g))+g_{ib}^n\times \phi^n\\
=&(v_{eb}^n\times)M_1\varepsilon^r-\left[ (2\tilde{\omega}_{ie}^n+\tilde{\omega}_{en}^n)\times\right]\varepsilon^v+v_{eb}^n\times(C_b^n(b_g+w_g))\\
&+\left(-(v_{eb}^n\times)M_1(r_{eb}^n\times) +(v_{eb}^n\times)(\omega_{ie}^n\times)+(g_{ib}^n\times) \right)\phi^n+C_b^n(b_a+w_a)
\end{aligned}
\end{equation}
\begin{equation}\label{Position_n_d_new_estimated_navigation_frame}
\begin{aligned}
\frac{d}{dt}\varepsilon^r=&-\omega_{en}^n\times\varepsilon^r+\varepsilon^v-(r_{eb}^n\times)(\omega_{ie}^n\times)\phi^n\\
&-r_{eb}^n\times M_1(\varepsilon^r-r_{eb}^n\times \phi^n) +r_{eb}^n\times(C_b^n(b_g+w_g))\\
=&-(r_{eb}^n\times M_1+(\omega_{en}^n\times))\varepsilon^r+\varepsilon^v-((r_{eb}^n\times)(\omega_{ie}^n\times)-(r_{eb}^n\times) M_1(r_{eb}^n\times))\phi^n\\
&+r_{eb}^n\times(C_b^n(b_g+w_g))
\end{aligned}
\end{equation}

If the position is represented in terms of LLH,  $\delta\omega_{ie}^n$ and $\delta\omega_{in}^n$ can be calculated as
\begin{equation}\label{perturbation_omega_ie_n_new_LLH_estimated_navigation_frame}
\delta\omega_{ie}^n=N_1\delta r_{eb}^l=N_1(\eta^r-r_{eb}^l\times \phi^n)
\end{equation}
\begin{equation}\label{perturbation_omega_in_n_new_LLH_estimated_navigation_frame}
\delta\omega_{in}^n=(N_1+N_3)\delta r_{eb}^l+N_2\delta v_{eb}^n=(N_1+N_3)(\eta^r-r_{eb}^l\times \phi^n)+N_2(\eta^v-v_{eb}^n\times \phi^n)
\end{equation}

Then, the new differential equations of  attitude error, velocity error and position error can be calculated as
\begin{equation}\label{attitude_n_d_new_LLH_estimated_navigation_frame}
\begin{aligned}
&\dot{\phi}^n=-\omega_{in}^n\times\phi^n-(N_1+N_3)(\eta^r-\tilde{r}_{eb}^l\times \phi^n)-N_2(\eta^v-\tilde{v}_{eb}^n\times \phi^n)+C_b^n(b_g+w_g)\\
=&-(N_1+N_3)\eta^r-N_2\eta^v-\left((\omega_{in}^n\times)-N_2(v_{eb}^n\times)-(N_1+N_3)(r_{eb}^l\times)\right)\phi^n+C_b^n(b_g+w_g)
\end{aligned}
\end{equation}
\begin{equation}\label{Velocity_n_d_new_LLH_estimated_navigation_frame}
\begin{aligned}
\frac{d}{dt}\varepsilon^v=
&C_b^n(b_a+w_a)-\left[ (2\tilde{\omega}_{ie}^n+\tilde{\omega}_{en}^n)\times\right]\varepsilon^v+(v_{eb}^n\times)N_1(\varepsilon^r-r_{eb}^l\times \phi^n)\\
&+(v_{eb}^n\times)(\omega_{ie}^n\times)\phi^n  
+(v_{eb}^n\times)(C_b^n(b_g+w_g))+g_{ib}^n\times \phi^n\\
=&(v_{eb}^n\times)N_1\varepsilon^r-\left[ (2\tilde{\omega}_{ie}^n+\tilde{\omega}_{en}^n)\times\right]\varepsilon^v+v_{eb}^n\times(C_b^n(b_g+w_g))\\
&+\left(-(v_{eb}^n\times)N_1(r_{eb}^l\times) +(v_{eb}^n\times)(\omega_{ie}^n\times)+(g_{ib}^n\times) \right)\phi^n+C_b^n(b_a+w_a)
\end{aligned}
\end{equation}
\begin{equation}\label{Position_n_d_new_LLH_estimated_navigation_frame}
\begin{aligned}
&\frac{d}{dt}\varepsilon^r= \frac{d}{dt}(\tilde{r}_{eb}^l-\tilde{C}_b^n C_n^b r_{eb}^l)\approx \frac{d}{dt}(\tilde{r}_{eb}^l-r_{eb}^l-\phi^n\times r_{eb}^l)=\frac{d}{dt}(\delta r_{eb}^l+r_{eb}^l\times \phi^n)\\
=&(N_{rr}\delta r_{eb}^l+N_{rv}\delta v_{eb}^n)+((N_{rv}v_{eb}^n)\times)\phi^n+(r_{eb}^l\times)(-\omega_{in}^n\times\phi^n-\delta\omega_{in}^n+C_b^n\delta\omega_{ib}^b)\\
=&\left(N_{rr}(\varepsilon^r-r_{eb}^l\times \phi^n)+N_{rv}(\varepsilon^v-v_{eb}^n\times \phi^n)\right)+\left((N_{rv} v_{eb}^n)\times\right)\phi^n\\
&-(r_{eb}^l\times)(N_1+N_3)\varepsilon^r-(r_{eb}^l\times)N_2\varepsilon^v\\
&-(r_{eb}^l\times)\left((\omega_{in}^n\times)+N_2(\tilde{v}_{eb}^n\times)+(N_1+N_3)(\tilde{r}_{eb}^l\times)\right)\phi^n+(r_{eb}^l\times)C_b^n(b_g+w_g)\\
=&\left(N_{rr}-(r_{eb}^l\times)(N_1+N_3)\right)\varepsilon^r+\left(N_{rv}-(r_{eb}^l\times)N_2\right)\varepsilon^v\\
&-\left(N_{rr}(r_{eb}^l\times)+N_{rv}(v_{eb}^n\times)-\left((N_{rv}v_{eb}^n)\times\right)+(r_{eb}^l\times)(\omega_{in}^n\times)+(r_{eb}^l\times)N_2(v_{eb}^n\times)\right.\\
&\left. +(r_{eb}^l\times)(N_1+N_3)(r_{eb}^l\times)\right)\phi^n+(r_{eb}^l\times)C_b^n(b_g+w_g)
\end{aligned}
\end{equation}
\subsection{$SE_2(3)$ based EKF for NED Navigation with Body Frame Attitude Error}
When the error state is left invariant by the left group action, this is the world-centric estimator formulation and is suitable for sensors such as GNSS, 5G, etc. If the error state is converted to the true body frame, i.e.,
$\varepsilon_l=(L{\mathcal{X}})^{-1}(L\tilde{\mathcal{X}})={\mathcal{X}}^{-1}\tilde{\mathcal{X}}\in SE_2(3)$, then
\begin{equation}\label{left_error_true_body}
\varepsilon_l={\mathcal{X}}^{-1}\tilde{\mathcal{X}}=\begin{bmatrix}
C_n^b\tilde{C}_b^n & C_n^b(\tilde{v}_{eb}^n-v_{eb}^n)& C_n^b(\tilde{r}_{eb}^n-r_{eb}^n)\\
0_{1\times 3} &1 &0\\
0_{1\times 3} &0&1
\end{bmatrix}=\begin{bmatrix}
\varepsilon^a & \varepsilon^v& \varepsilon^r\\
0_{1\times 3} &1 &0\\
0_{1\times 3} &0&1
\end{bmatrix}
\end{equation}

The error state can be converted to the corresponding Lie algebra as follows:
\begin{equation}\label{liealgebre_left_error_true_body}
\varepsilon_l\triangleq\begin{bmatrix}
\exp_G(\phi^b\times) & J\rho_v^b & J\rho_r^b\\
0_{1\times 3} &1 &0\\
0_{1\times 3} &0&1
\end{bmatrix}=\exp_G\left( \begin{bmatrix}
\phi^b\times & \rho_v^b & \rho_r^b\\
0_{1\times 3} &0 &0\\
0_{1\times 3} &0&0
\end{bmatrix}\right)=\exp_G\left(\Lambda \begin{bmatrix}
\phi^b \\ \rho_v^b \\ \rho_r^b
\end{bmatrix}\right)=\exp_G\left(\Lambda (\rho^b) \right)
\end{equation}
where $\phi^b$ is the attitude error expressed in the body frame; $\rho^b=\begin{pmatrix}
(\phi^b)^T & (\rho_v^b)^T &(\rho_r^b)^T
\end{pmatrix}^T$ represents the Lie algebra corresponding to the state error $\varepsilon_l$.

Comparing equation (\ref{left_error_true_body}) and equation (\ref{liealgebre_left_error_true_body}), we can get 
\begin{equation}\label{error_new_definition_attitude_left_body}
\varepsilon^a=\exp_G(\phi^b\times)=C_n^b\tilde{C}_b^n\approx I_{3\times3}+\phi^b\times, \text{if $||\phi^b||$ is small}
\end{equation}
\begin{equation}\label{error_new_definition_velocity_left_body}
\varepsilon^v=J\rho_v^b=C_n^b(\tilde{v}_{eb}^n-v_{eb}^n)=C_n^b \delta v_{eb}^n
\end{equation}
\begin{equation}\label{error_new_definition_position_left_body}
\varepsilon^r= J\rho_r^b=C_n^b(\tilde{r}_{eb}^n-r_{eb}^n)=C_n^b \delta r_{eb}^n
\end{equation}

Now, we consider the differential equations for the attitude error, velocity error, and position error which can form  a element of the $SE_2(3)$ matrix Lie group. On the one hand, by taking differential of attitude error $\varepsilon^a$ with respect to time, we can get
\begin{equation}\label{attitude_n_d_left_body_frame}
\begin{aligned}
&\frac{d}{dt}\varepsilon^a=\frac{d}{dt}C_n^b\tilde{C}_b^n
=\dot{C}_n^b\tilde{C}_b^n+C_n^b\dot{\tilde{C}}_b^n\\
=&\left(C_n^b(\omega_{in}^n\times)-(\omega_{ib}^b\times)C_n^b \right)\tilde{C}_b^n+C_n^b\left( \tilde{C}_b^n(\tilde{\omega}_{ib}^b\times)-(\tilde{\omega}_{in}^n\times)\tilde{C}_b^n\right)\\
=&C_n^b(\omega_{in}^n\times)\tilde{C}_b^n-(\omega_{ib}^b\times)C_n^b\tilde{C}_b^n+C_n^b\tilde{C}_b^n(\tilde{\omega}_{ib}^b\times)-C_n^b(\tilde{\omega}_{in}^n\times)\tilde{C}_b^n\\
\approx& -(\omega_{ib}^b\times)(I_{3\times3}+\phi^b\times)+(I_{3\times3}+\phi^b\times)(\tilde{\omega}_{ib}^b\times)-C_n^b((\tilde{\omega}_{in}^n-\omega_{in}^n)\times)\tilde{C}_b^n  \\
=& -\omega_{ib}^b\times-(\omega_{ib}^b\times)(\phi^b\times)+\tilde{\omega}_{ib}^b\times+(\phi^b\times)(\tilde{\omega}_{ib}^b\times)-C_n^b(\delta\omega_{in}^n\times)C_b^nC_n^b\tilde{C}_b^n\\
\approx &\delta \omega_{ib}^b\times+(\phi^b\times\omega_{ib}^b)\times +(\phi^b\times)(\delta\omega_{ib}^b\times)-\delta\omega_{in}^b\times-(\delta\omega_{in}^b\times)(\phi^b\times)\\
\approx &\delta \omega_{ib}^b\times+(\phi^b\times\omega_{ib}^b)\times -\delta\omega_{in}^b\times
\end{aligned}
\end{equation}
where the  2-order small quantities $(\phi^b\times)(\delta\omega_{ib}^b\times)$ and $(\delta\omega_{in}^b\times)(\phi^b\times)$ are neglected at the last step; $\delta\omega_{in}^n$ is defined as $\delta\omega_{in}^n\triangleq \tilde{\omega}_{in}^n-\omega_{in}^n$; $\delta\omega_{ib}^b$ is defined as $\delta\omega_{ib}^b\triangleq \tilde{\omega}_{ib}^b-\omega_{ib}^b$.

On the other hand, 
\begin{equation}\label{attitude_n_d_approx_body_frame}
\frac{d}{dt}\varepsilon^a\approx \frac{d}{dt}(I_{3\times3}+\phi^b\times)=\dot{\phi}^b\times
\end{equation}

Therefore, the state error differential equation for the attitude error can be written as follows
\begin{equation}\label{attitude_n_d_body_frame}
\dot{\phi}^b=\delta\omega_{ib}^b+(\phi^b\times\omega_{ib}^b)-\delta\omega_{in}^b=-\omega_{ib}^b\times\phi^b+\delta\omega_{ib}^b-C_n^b\delta\omega_{in}^n
\end{equation}

By taking differential of velocity error $\varepsilon^v$ with respect to time and substituting equation (\ref{error_new_definition_velocity_left_body}) into it, we can get
\begin{equation}\label{Velocity_n_d_body_frame}
\begin{aligned}
&\frac{d}{dt}\varepsilon^v=\frac{d}{dt}\left(C_n^b(\tilde{v}_{eb}^n-v_{eb}^n)\right)=\dot{C}_n^b(\tilde{v}_{eb}^n-v_{eb}^n)+C_n^b(\dot{\tilde{v}}_{eb}^n-\dot{v}_{eb}^n)\\
=&\left(C_n^b(\omega_{in}^n\times)-(\omega_{ib}^b\times)C_n^b \right) \delta v_{eb}^n\\
&+C_n^b\left\{\left[\tilde{C}_b^n\tilde{f}_{ib}^b-\left[(2\tilde{\omega}_{ie}^n+\tilde{\omega}_{en}^n)\times\right]\tilde{v}_{eb}^n+\tilde{g}_{ib}^n\right]-
\left[{C}_b^n{f}_{ib}^b-\left[(2{\omega}_{ie}^n+{\omega}_{en}^n)\times\right]{v}_{eb}^n+{g}_{ib}^n\right]\right\}\\
=&C_n^b(\omega_{in}^n\times)\delta v_{eb}^n-(\omega_{ib}^b\times)C_n^b\delta v_{eb}^n+C_n^b\tilde{C}_b^n\tilde{f}_{ib}^b-C_n^b\left[(2\tilde{\omega}_{ie}^n+\tilde{\omega}_{en}^n)\times\right]\tilde{v}_{eb}^n+C_n^b\tilde{g}_{ib}^n\\
&-C_n^b{C}_b^n{f}_{ib}^b+C_n^b\left[(2{\omega}_{ie}^n+{\omega}_{en}^n)\times\right]{v}_{eb}^n-C_n^b{g}_{ib}^n\\
\approx & C_n^b(\omega_{in}^n\times)\delta v_{eb}^n-(\omega_{ib}^b\times)C_n^b\delta v_{eb}^n+\delta f_{ib}^b+\phi^b\times \tilde{f}_{ib}^b+C_n^b\delta g_{ib}^n\\
&-C_n^b(2\omega_{ie}^n+\omega_{en}^n)\times \delta v_{eb}^n-C_n^b(2\delta\omega_{ie}^n+\delta\omega_{en}^n)\times \tilde{v}_{eb}^n\\
=&-(\omega_{ib}^b\times)\varepsilon^v+\delta f_{ib}^b+\phi^b\times \tilde{f}_{ib}^b+C_n^b\delta g_{ib}^n-C_n^b\omega_{ie}^n\times \delta v_{eb}^n-C_n^b(\delta\omega_{ie}^n+\delta\omega_{in}^n)\times \tilde{v}_{eb}^n\\
=&-(\omega_{ib}^b\times)\varepsilon^v+\delta f_{ib}^b+\phi^b\times \tilde{f}_{ib}^b+C_n^b\delta g_{ib}^n-(C_n^b\omega_{ie}^n)\times \varepsilon^v+C_n^b\tilde{v}_{eb}^n\times(\delta\omega_{ie}^n+\delta\omega_{in}^n)
\end{aligned}
\end{equation}
where $\delta f_{ib}^b$ is defined as $\delta f_{ib}^b \triangleq \tilde{f}_{ib}^b- f_{ib}^b$; $\delta\omega_{ie}^n$ is defined as $\delta\omega_{ie}^n\triangleq \tilde{\omega}_{ie}^n-\omega_{ie}^n$; $\delta g_{ib}^n$ is defined as $\delta g_{ib}^n\triangleq \tilde{g}_{ib}^n-g_{ib}^n$ and it can be neglected as the change of $g_{ib}^n$ is quite small for carrier's local navigation.

By taking differential of position error $\varepsilon^r$ with respect to time and substituting equation (\ref{error_new_definition_position_left_body}) into it, we can get
\begin{equation}\label{Position_n_d_body_frame}
\begin{aligned}
&\frac{d}{dt}\varepsilon^r= \frac{d}{dt}\left(C_n^b(\tilde{r}_{eb}^n-r_{eb}^n)\right)=\dot{C}_n^b(\tilde{r}_{eb}^n-r_{eb}^n)+C_n^b(\dot{\tilde{r}}_{eb}^n-\dot{r}_{eb}^n)\\
=&\left(C_n^b(\omega_{in}^n\times)-(\omega_{ib}^b\times)C_n^b \right) \delta r_{eb}^n+C_n^b\left[(-\tilde{\omega}_{en}^n\times \tilde{r}_{eb}^n+\tilde{v}_{eb}^n)-(-{\omega}_{en}^n\times {r}_{eb}^n+{v}_{eb}^n)\right]\\
=&C_n^b(\omega_{in}^n\times)\delta r_{eb}^n-(\omega_{ib}^b\times)C_n^b\delta r_{eb}^n+C_n^b\delta {v}_{eb}^n-C_n^b(\tilde{\omega}_{en}^n\times \tilde{r}_{eb}^n-{\omega}_{en}^n\times {r}_{eb}^n)\\
=&C_n^b(\omega_{in}^n\times)\delta r_{eb}^n-(\omega_{ib}^b\times)C_n^b\delta r_{eb}^n+C_n^b\delta {v}_{eb}^n-C_n^b(\delta\omega_{en}^n\times r_{eb}^n+\tilde{\omega}_{en}^n\times \delta r_{eb}^n)\\
=&-(\omega_{ib}^b\times)\varepsilon^r+\varepsilon^v+C_n^br_{eb}^n\times\delta\omega_{en}^n+C_n^b\omega_{ie}^n\times\delta r_{eb}^n-C_n^b(\delta{\omega}_{en}^n\times \delta r_{eb}^n)\\
\approx&-(\omega_{ib}^b\times)\varepsilon^r+\varepsilon^v+C_n^br_{eb}^n\times\delta\omega_{en}^n+(C_n^b\omega_{ie}^n)\times \varepsilon^r
\end{aligned}
\end{equation}
where 2-order small quantities $C_n^b(\delta{\omega}_{en}^n\times \delta r_{eb}^n)$ is neglected at the last step.

With the new definition of the attitude error, velocity error, and position error, we substitute equation (\ref{error_new_definition_velocity_left_body}) and equation (\ref{error_new_definition_position_left_body}) into equation (\ref{perturbation_omega_en_n}) and equation (\ref{perturbation_omega_in_n}):
\begin{equation}\label{perturbation_omega_ie_n_left_invariant}
\delta\omega_{ie}^n=M_1\delta r_{eb}^n=M_1C_b^n\varepsilon^r
\end{equation}
\begin{equation}\label{perturbation_omega_in_n_left_invariant}
\delta\omega_{in}^n=(M_1+M_3)\delta r_{eb}^n+M_2\delta v_{eb}^n=(M_1+M_3)C_b^n\varepsilon^r+M_2C_b^n\varepsilon^v
\end{equation}

Consequently, the state error dynamical equations with respect to the true body frame can be written as follows:
\begin{equation}\label{attitude_n_d_leftinvariant_body_frame}
\begin{aligned}
&\dot{\phi}^b=-\omega_{ib}^b\times\phi^b+(b_g+w_g)-C_n^b((M_1+M_3)C_b^n\varepsilon^r+M_2C_b^n\varepsilon^v)\\
=&-C_n^b(M_1+M_3)C_b^n\varepsilon^r-C_n^bM_2C_b^n\varepsilon^v-\omega_{ib}^b\times\phi^b+(b_g+w_g)
\end{aligned}
\end{equation}
\begin{equation}\label{Velocity_n_d_left_body_frame}
\begin{aligned}
\frac{d}{dt}\varepsilon^v=
&-(\omega_{ib}^b\times)\varepsilon^v+\delta f_{ib}^b+\phi^b\times \tilde{f}_{ib}^b+C_n^b\delta g_{ib}^n-(C_n^b\omega_{ie}^n)\times \varepsilon^v\\
&+C_n^b\tilde{v}_{eb}^n\times(M_1C_b^n\varepsilon^r+(M_1+M_3)C_b^n\varepsilon^r+M_2C_b^n\varepsilon^v)\\
=&C_n^b(\tilde{v}_{eb}^n\times)(2M_1+M_3)C_b^n\varepsilon^r+(C_n^b(\tilde{v}_{eb}^n\times)M_2C_b^n-(\omega_{ib}^b\times)-(C_n^b\omega_{ie}^n)\times)\varepsilon^v\\
&-\tilde{f}_{ib}^b \times\phi^b +C_n^b\delta g_{ib}^n+(b_a+w_a)
\end{aligned}
\end{equation}
\begin{equation}\label{Position_n_d_left_body_frame}
\begin{aligned}
\frac{d}{dt}\varepsilon^r=&-(\omega_{ib}^b\times)\varepsilon^r+\varepsilon^v+C_n^br_{eb}^n\times(M_3C_b^n\varepsilon^r+M_2C_b^n\varepsilon^v)+(C_n^b\omega_{ie}^n)\times \varepsilon^r\\
=&\left(C_n^b(r_{eb}^n\times)M_3C_b^n-(\omega_{ib}^b\times)+(C_n^b\omega_{ie}^n)\times\right)\varepsilon^r+\left( I+C_n^b(r_{eb}^n\times)M_2C_b^n\right)\varepsilon^v
\end{aligned}
\end{equation}
\subsection{$SE_2(3)$ based EKF for NED Navigation with estimated Body Frame Attitude Error}
If the state error is converted to the estimated body frame, i.e.,
$\varepsilon_l=(L\tilde{\mathcal{X}})^{-1}(L\mathcal{X})=\tilde{\mathcal{X}}^{-1}\mathcal{X}\in SE_2(3)$, the error is left invariant. The definition of the error on matrix Lie group as inverse of the estimated state multiplies the true state is similar to the error defined on the Euclidean space as true vector minus estimated vector.

The error state is given as
\begin{equation}\label{left_error_estimated_body}
\varepsilon_e={\tilde{\mathcal{X}}}^{-1}\mathcal{X}=\begin{bmatrix}
\tilde{C}_n^b C_b^n & \tilde{C}_n^b(v_{eb}^n-\tilde{v}_{eb}^n)& \tilde{C}_n^b(r_{eb}^n-\tilde{r}_{eb}^n)\\
0_{1\times 3} &1 &0\\
0_{1\times 3} &0&1
\end{bmatrix}=\begin{bmatrix}
\varepsilon^a & \varepsilon^v& \varepsilon^r\\
0_{1\times 3} &1 &0\\
0_{1\times 3} &0&1
\end{bmatrix}
\end{equation}

The error state can be converted to the corresponding Lie algebra as follows:
\begin{equation}\label{liealgebre_left_error_estimated_body}
\varepsilon_e\triangleq\begin{bmatrix}
\exp_G(\phi^{\tilde{b}}\times) & J\rho_v^{\tilde{b}} & J\rho_r^{\tilde{b}}\\
0_{1\times 3} &1 &0\\
0_{1\times 3} &0&1
\end{bmatrix}=\exp_G\left( \begin{bmatrix}
\phi^{\tilde{b}}\times & \rho_v^{\tilde{b}} & \rho_r^{\tilde{b}}\\
0_{1\times 3} &0 &0\\
0_{1\times 3} &0&0
\end{bmatrix}\right)=\exp_G\left(\Lambda \begin{bmatrix}
\phi^{\tilde{b}} \\ \rho_v^{\tilde{b}} \\ \rho_r^{\tilde{b}}
\end{bmatrix}\right)=\exp_G\left(\Lambda (\rho^{\tilde{b}}) \right)
\end{equation}
where $\phi^{\tilde{b}}$ is the attitude error expressed in the estimated body frame; $\rho^{\tilde{b}}=\begin{pmatrix}
(\phi^{\tilde{b}})^T & (\rho_v^{\tilde{b}})^T &(\rho_r^{\tilde{b}})^T
\end{pmatrix}^T$ represents the Lie algebra corresponding to the state error $\varepsilon_e$;

Comparing equation (\ref{left_error_estimated_body}) and equation (\ref{liealgebre_left_error_estimated_body}), we can get 
\begin{equation}\label{error_estimated_definition_attitude_left_body}
\varepsilon^a=\exp_G(\phi^{\tilde{b}}\times)=\tilde{C}_n^b C_b^n\approx I_{3\times3}+\phi^{\tilde{b}}\times, \text{if $||\phi^{\tilde{b}}||$ is small}
\end{equation}
\begin{equation}\label{error_estimated_definition_velocity_left_body}
\varepsilon^v=J\rho_v^{\tilde{b}}=\tilde{C}_n^b(v_{eb}^n-\tilde{v}_{eb}^n)=-\tilde{C}_n^b \delta v_{eb}^n
\end{equation}
\begin{equation}\label{error_estimated_definition_position_left_body}
\varepsilon^r= J\rho_r^{\tilde{b}}=\tilde{C}_n^b(r_{eb}^n-\tilde{r}_{eb}^n)=-\tilde{C}_n^b \delta r_{eb}^n
\end{equation}

Now, we consider the differential equations for the attitude error, velocity error, and position error which can form  a element of the $SE_2(3)$ matrix Lie group. On the one hand, by taking differential of attitude error $\varepsilon^a$ with respect to time, we can get
\begin{equation}\label{attitude_n_d_left_estimated_body_frame}
\begin{aligned}
&\frac{d}{dt}\varepsilon^a=\frac{d}{dt}\tilde{C}_n^b C_b^n
=\dot{\tilde{C}}_n^bC_b^n+\tilde{C}_n^b\dot{C}_b^n\\
=&\left(\tilde{C}_n^b(\tilde{\omega}_{in}^n\times)-(\tilde{\omega}_{ib}^b\times)\tilde{C}_n^b \right)C_b^n+\tilde{C}_n^b\left( C_b^n(\omega_{ib}^b\times)-(\omega_{in}^n\times)C_b^n\right)\\
=&\tilde{C}_n^b(\tilde{\omega}_{in}^n\times)C_b^n-(\tilde{\omega}_{ib}^b\times)\tilde{C}_n^bC_b^n+\tilde{C}_n^bC_b^n(\omega_{ib}^b\times)-\tilde{C}_n^b(\omega_{in}^n\times)C_b^n\\
\approx& -(\tilde{\omega}_{ib}^b\times)(I_{3\times3}+\phi^b\times)+(I_{3\times3}+\phi^b\times)(\omega_{ib}^b\times)+\tilde{C}_n^b((\tilde{\omega}_{in}^n-\omega_{in}^n)\times)C_b^n  \\
=& -\tilde{\omega}_{ib}^b\times-(\tilde{\omega}_{ib}^b\times)(\phi^b\times)+\omega_{ib}^b\times+(\phi^b\times)(\omega_{ib}^b\times)+\tilde{C}_n^bC_b^nC_n^b(\delta\omega_{in}^n\times)C_b^n\\
\approx &-\delta \omega_{ib}^b\times+(\phi^b\times\omega_{ib}^b)\times -(\delta\omega_{ib}^b\times)(\phi^b\times)+\delta\omega_{in}^b\times+\phi^b\times(\delta\omega_{in}^b\times)\\
\approx &-\delta \omega_{ib}^b\times+(\phi^b\times\omega_{ib}^b)\times +\delta\omega_{in}^b\times
\end{aligned}
\end{equation}
where the  2-order small quantities $(\delta\omega_{ib}^b\times)(\phi^b\times)$ and $(\phi^b\times)(\delta\omega_{in}^b\times)$ are neglected at the last step; $\delta\omega_{in}^n$ is defined as $\delta\omega_{in}^n\triangleq \tilde{\omega}_{in}^n-\omega_{in}^n$; $\delta\omega_{ib}^b$ is defined as $\delta\omega_{ib}^b\triangleq \tilde{\omega}_{ib}^b-\omega_{ib}^b$.

On the other hand, 
\begin{equation}\label{attitude_n_d_approx_estimated_body_frame}
\frac{d}{dt}\varepsilon^a\approx \frac{d}{dt}(I_{3\times3}+\phi^{\tilde{b}}\times)=\dot{\phi}^{\tilde{b}}\times
\end{equation}

Therefore, the state error differential equation for the attitude error can be written as follows
\begin{equation}\label{attitude_n_d_estimated_body_frame}
\dot{\phi}^{\tilde{b}}=-\delta\omega_{ib}^b+(\phi^{\tilde{b}}\times\omega_{ib}^b)+\delta\omega_{in}^b=-\omega_{ib}^b\times\phi^{\tilde{b}}-\delta\omega_{ib}^b+C_n^b\delta\omega_{in}^n
\end{equation}

By taking differential of velocity error $\varepsilon^v$ with respect to time and substituting equation (\ref{error_estimated_definition_velocity_left_body}) into it, we can get
\begin{equation}\label{Velocity_n_d_estimated_body_frame}
\begin{aligned}
&\frac{d}{dt}\varepsilon^v=\frac{d}{dt}\left(\tilde{C}_n^b(v_{eb}^n-\tilde{v}_{eb}^n)\right)=\dot{\tilde{C}}_n^b(v_{eb}^n-\tilde{v}_{eb}^n)+\tilde{C}_n^b(\dot{v}_{eb}^n-\dot{\tilde{v}}_{eb}^n)\\
=&-\left(\tilde{C}_n^b(\tilde{\omega}_{in}^n\times)-(\tilde{\omega}_{ib}^b\times)\tilde{C}_n^b \right) \delta {v}_{eb}^n\\
&+\tilde{C}_n^b\left\{\left[C_b^nf_{ib}^b-\left[(2\omega_{ie}^n+\omega_{en}^n)\times\right]v_{eb}^n+g^n\right]-
\left[\tilde{C}_b^n\tilde{f}_{ib}^b-\left[(2\tilde{\omega}_{ie}^n+\tilde{\omega}_{en}^n)\times\right]\tilde{v}_{eb}^n+\tilde{g}^n\right]\right\}\\
=&-\tilde{C}_n^b(\tilde{\omega}_{in}^n\times)\delta v_{eb}^n+(\tilde{\omega}_{ib}^b\times)\tilde{C}_n^b\delta v_{eb}^n+\tilde{C}_n^bC_b^n{f}_{ib}^b-\tilde{C}_n^b\left[(2\omega_{ie}^n+\omega_{en}^n)\times\right]v_{eb}^n+\tilde{C}_n^bg^n\\
&-\tilde{C}_n^b\tilde{C}_b^n\tilde{f}_{ib}^b+\tilde{C}_n^b\left[(2\tilde{\omega}_{ie}^n+\tilde{\omega}_{en}^n)\times\right]\tilde{v}_{eb}^n-\tilde{C}_n^b\tilde{g}^n\\
\approx & -\tilde{C}_n^b(\tilde{\omega}_{in}^n\times)\delta v_{eb}^n+(\tilde{\omega}_{ib}^b\times)\tilde{C}_n^b\delta v_{eb}^n-\delta f_{ib}^b+\phi^b\times {f}_{ib}^b-\tilde{C}_n^b\delta g^n\\
&+\tilde{C}_n^b(2\tilde{\omega}_{ie}^n+\tilde{\omega}_{en}^n)\times \delta v_{eb}^n+\tilde{C}_n^b(2\delta\omega_{ie}^n+\delta\omega_{en}^n)\times {v}_{eb}^n\\
=&-(\tilde{\omega}_{ib}^b\times)\varepsilon^v-\delta f_{ib}^b+\phi^b\times {f}_{ib}^b-\tilde{C}_n^b\delta g^n+\tilde{C}_n^b(\tilde{\omega}_{ie}^n\times )\tilde{C}_b^n\tilde{C}_n^b \delta v_{eb}^n+\tilde{C}_n^b(\delta\omega_{ie}^n+\delta\omega_{in}^n)\times {v}_{eb}^n\\
=&-(\tilde{\omega}_{ib}^b\times)\varepsilon^v-\delta f_{ib}^b+\phi^b\times {f}_{ib}^b-\tilde{C}_n^b\delta g^n-(\tilde{C}_n^b\tilde{\omega}_{ie}^n)\times \varepsilon^v-\tilde{C}_n^b({v}_{eb}^n\times) (\delta\omega_{ie}^n+\delta\omega_{in}^n)\\
\approx&-({\omega}_{ib}^b\times)\varepsilon^v-\delta f_{ib}^b+\phi^b\times {f}_{ib}^b-{C}_n^b\delta g^n-({C}_n^b{\omega}_{ie}^n)\times \varepsilon^v-{C}_n^b({v}_{eb}^n\times) (\delta\omega_{ie}^n+\delta\omega_{in}^n)
\end{aligned}
\end{equation}
where $\delta f_{ib}^b$ is defined as $\delta f_{ib}^b \triangleq \tilde{f}_{ib}^b- f_{ib}^b$; $\delta\omega_{ie}^n$ is defined as $\delta\omega_{ie}^n\triangleq \tilde{\omega}_{ie}^n-\omega_{ie}^n$; $\delta g^n$ is defined as $\delta g^n\triangleq \tilde{g}^n-g^n$ and it can be neglected as the change of $g^n$ is quite small for carrier's local navigation.

By taking differential of position error $\varepsilon^r$ with respect to time and substituting equation (\ref{error_estimated_definition_position_left_body}) into it, we can get
\begin{equation}\label{Position_n_d_estimated_body_frame}
\begin{aligned}
&\frac{d}{dt}\varepsilon^r= \frac{d}{dt}\left(\tilde{C}_n^b(r_{eb}^n-\tilde{r}_{eb}^n)\right)=\dot{\tilde{C}}_n^b(r_{eb}^n-\tilde{r}_{eb}^n)+\tilde{C}_n^b(\dot{{r}}_{eb}^n-\dot{\tilde{r}}_{eb}^n)\\
=&-\left(\tilde{C}_n^b(\tilde{\omega}_{in}^n\times)-(\tilde{\omega}_{ib}^b\times)\tilde{C}_n^b \right) \delta r_{eb}^n+\tilde{C}_n^b\left[(-{\omega}_{en}^n\times {r}_{eb}^n+{v}_{eb}^n)-(-\tilde{\omega}_{en}^n\times \tilde{r}_{eb}^n+\tilde{v}_{eb}^n)\right]\\
=&-\tilde{C}_n^b(\tilde{\omega}_{in}^n\times)\delta r_{eb}^n+(\tilde{\omega}_{ib}^b\times)\tilde{C}_n^b\delta r_{eb}^n-\tilde{C}_n^b\delta v_{eb}^n+\tilde{C}_n^b(\tilde{\omega}_{en}^n\times \tilde{r}_{eb}^n-{\omega}_{en}^n\times {r}_{eb}^n)\\
=&-\tilde{C}_n^b(\tilde{\omega}_{in}^n\times)\delta r_{eb}^n+(\tilde{\omega}_{ib}^b\times)\tilde{C}_n^b\delta r_{eb}^n-\tilde{C}_n^b\delta v_{eb}^n+\tilde{C}_n^b(\delta\omega_{en}^n\times \tilde{r}_{eb}^n+{\omega}_{en}^n\times \delta r_{eb}^n)\\
=&-(\tilde{\omega}_{ib}^b\times)\varepsilon^r+\varepsilon^v-\tilde{C}_n^b(\tilde{r}_{eb}^n\times)\delta\omega_{en}^n-\tilde{C}_n^b\omega_{ie}^n\times \tilde{C}_b^n\tilde{C}_n^b \delta r_{eb}^n-\tilde{C}_n^b(\delta{\omega}_{in}^n\times) \delta r_{eb}^n\\
\approx&-(\tilde{\omega}_{ib}^b\times)\varepsilon^r+\varepsilon^v-\tilde{C}_n^b(\tilde{r}_{eb}^n\times)\delta\omega_{en}^n+(\tilde{C}_n^b\omega_{ie}^n)\times \varepsilon^r\\
\approx&-({\omega}_{ib}^b\times)\varepsilon^r+\varepsilon^v-{C}_n^b({r}_{eb}^n\times)\delta\omega_{en}^n+({C}_n^b\omega_{ie}^n)\times \varepsilon^r
\end{aligned}
\end{equation}
where 2-order small quantities $\tilde{C}_n^b(\delta{\omega}_{in}^n\times \delta r_{eb}^n)$ is neglected at the last step.

With the new definition of the attitude error, velocity error, and position error, we substitute equation (\ref{error_estimated_definition_velocity_left_body}) and equation (\ref{error_estimated_definition_position_left_body}) into equation (\ref{perturbation_omega_en_n}) and equation (\ref{perturbation_omega_in_n}):
\begin{equation}\label{perturbation_omega_ie_n_estimated_left_invariant}
\delta\omega_{ie}^n=M_1\delta r_{eb}^n=-M_1\tilde{C}_b^n\varepsilon^r
\end{equation}
\begin{equation}\label{perturbation_omega_in_n_estimated_left_invariant}
\delta\omega_{in}^n=(M_1+M_3)\delta r_{eb}^n+M_2\delta v_{eb}^n=-(M_1+M_3)\tilde{C}_b^n\varepsilon^r-M_2\tilde{C}_b^n\varepsilon^v
\end{equation}

Consequently, the state error dynamical equations with respect to the true body frame can be written as follows:
\begin{equation}\label{attitude_n_d_left_invariant_estimated_body_frame}
\begin{aligned}
&\dot{\phi}^b=-\omega_{ib}^b\times\phi^b-(b_g+w_g)+C_n^b(-(M_1+M_3)\tilde{C}_b^n\varepsilon^r-M_2\tilde{C}_b^n\varepsilon^v)\\
=&-C_n^b(M_1+M_3)\tilde{C}_b^n\varepsilon^r-C_n^bM_2\tilde{C}_b^n\varepsilon^v-\omega_{ib}^b\times\phi^b-(b_g+w_g)
\end{aligned}
\end{equation}
\begin{equation}\label{Velocity_n_d_left_estimated_body_frame}
\begin{aligned}
\frac{d}{dt}\varepsilon^v=
&-(\omega_{ib}^b\times)\varepsilon^v-\delta f_{ib}^b+\phi^b\times {f}_{ib}^b-C_n^b\delta g^n-(C_n^b\omega_{ie}^n)\times \varepsilon^v\\
&+C_n^b{v}_{eb}^n\times(M_1C_b^n\varepsilon^r+(M_1+M_3)C_b^n\varepsilon^r+M_2C_b^n\varepsilon^v)\\
=&C_n^b({v}_{eb}^n\times)(2M_1+M_3)C_b^n\varepsilon^r+(C_n^b({v}_{eb}^n\times)M_2C_b^n-(\omega_{ib}^b\times)-(C_n^b\omega_{ie}^n)\times)\varepsilon^v\\
&-{f}_{ib}^b \times\phi^b -C_n^b\delta g^n-(b_a+w_a)
\end{aligned}
\end{equation}
\begin{equation}\label{Position_n_d_left_estimated_body_frame}
\begin{aligned}
\frac{d}{dt}\varepsilon^r=&-(\omega_{ib}^b\times)\varepsilon^r+\varepsilon^v+C_n^br_{eb}^n\times(M_3C_b^n\varepsilon^r+M_2C_b^n\varepsilon^v)+(C_n^b\omega_{ie}^n)\times \varepsilon^r\\
=&\left(C_n^b(r_{eb}^n\times)M_3C_b^n-(\omega_{ib}^b\times)+(C_n^b\omega_{ie}^n)\times\right)\varepsilon^r+\left( I+C_n^b(r_{eb}^n\times)M_2C_b^n\right)\varepsilon^v
\end{aligned}
\end{equation}

It is obvious that the only difference between the true body frame and the estimate body frame lies in the $\delta f_{ib}^b$ term and the $\delta\omega_{ib}^b$ term.
\subsection{left-invariant measurement equation}
The GNSS provides navigation information in a global frame and has the left-invariant measurement equations on matrix Lie group.
GNSS  positioning solution gives the position coordinates of the antenna phase center(or other reference point), while SINS's mechanization gives the navigation results of the IMU measurement center. The two do not coincide physically, so the integrated navigation needs to correct the lever arm effect.
In the case of the arm lever error, we rearrange every measurement from GNSS as:
\begin{equation}\label{left_measurement_equation}
y_t=\begin{bmatrix}
r_{GNSS}^n \\ 0\\1
\end{bmatrix}=\begin{bmatrix}
C_b^n & v_{eb}^n & r_{eb}^n\\
0_{1\times3} & 1 & 0\\
0_{1\times 3} & 0& 1
\end{bmatrix}\begin{bmatrix}
l^b \\0\\1
\end{bmatrix}+\begin{bmatrix}
r_t\\0\\0
\end{bmatrix}\triangleq\mathcal{X}_tb+V_t
\end{equation}
where $r_{GNSS}^n$ is the positioning result calculated by GNSS and expressed in the navigation frame; $l^b$ is the lever arm measurement vector expressed in the body frame; $r_t$ is measurement white noise with covariance $R_t$.

Then, the left-innovation can be defined as 
\begin{equation}\label{left_innovation}
\begin{aligned}
&z_t=\tilde{\mathcal{X}}_t^{-1}y_t-b=\tilde{\mathcal{X}}_t^{-1}(\mathcal{X}_tb+V_t)-b=\varepsilon_e b-b+\tilde{\mathcal{X}}_t^{-1}V_t\\
\approx & (I+\Lambda (\rho^{\tilde{b}}) )b-b+\tilde{\mathcal{X}}_t^{-1}V_t=\Lambda (\rho^{\tilde{b}}) b+\tilde{\mathcal{X}}_t^{-1}V_t\\
=&\begin{bmatrix}
\phi^{\tilde{b}}\times & \rho_v^{\tilde{b}} & \rho_r^{\tilde{b}}\\
0_{1\times 3} &0 &0\\
0_{1\times 3} &0&0
\end{bmatrix}\begin{bmatrix}
l^b \\0\\1
\end{bmatrix}+\begin{bmatrix}
\tilde{C}_n^b & -\tilde{v}_{eb}^b & -\tilde{r}_{eb}^b\\
0_{1\times 3} &1 &0\\
0_{1\times 3} &0&1
\end{bmatrix}\begin{bmatrix}
r_t\\0\\0
\end{bmatrix}=\begin{bmatrix}
\phi^{\tilde{b}}\times l^b+\rho_r^{\tilde{b}} \\0\\0
\end{bmatrix}+\begin{bmatrix}
\tilde{C}_n^b r_t \\0\\0
\end{bmatrix}\\
=&H\rho^{\tilde{b}}+\tilde{V}_t
\end{aligned}
\end{equation}
where $H$ can be abbreviated as its reduced form as $H_{rt}=\begin{bmatrix}
-l^b{\times} & 0_{3\times3}& I_{3\times 3}
\end{bmatrix}$ by considering the computational efficiency, and $H$ is independent of the system state, but only related to the known vector $b$. $\tilde{V}_t$ can be abbreviated as $\tilde{r}_t=\tilde{C}_n^b r_t=(\tilde{C}_b^n)^{-1} r_t=M_tr_t$.
It is worth noting that the invariant-innovation can be termed as innovation expressed in the body frame.
\begin{remark}
	\label{remark_left_innnovation}
	From the definition of left-innovation, the inverse of the estimated system state is used to multiply the measurement is reasonable as the we get the state-independent measurement matrix H. Meanwhile, the form of the left-innovation can be viewed as analogous to the GNSS positioning results minus the SINS predicted values which is .
\end{remark}

When the biased of the acceleration and gyroscope are considered, the innovation vector can be quantified as 
\begin{equation}\label{innovation_bias}
\tilde{z}_t=\begin{bmatrix}
-l^b{\times} & 0_{3\times3}& I_{3\times 3}& 0_{3\times3}& 0_{3\times3}
\end{bmatrix}\begin{bmatrix}
\rho^{\tilde{b}} \\ \xi^b
\end{bmatrix}+\tilde{V}_t\triangleq H_t \delta x+M_tr_t
\end{equation}
where $\xi^b$ represents error state about the bias term.

Therefore, the Kalman filter gain can be partitioned into two parts: 
\begin{equation}\label{Kalman_filter_gain_invariant}
K_t=\begin{bmatrix}
K_t^{\zeta} \\K_t^{\xi}
\end{bmatrix}=P_tH_t^T(H_tP_tH_t^T+M_tR_tM_t^T)^{-1}
\end{equation}

The covariance update can be calculated as  
\begin{equation}\label{covariance_update}
P_t^+=(I-K_tH_t)P_t(I-K_tH_t)^T+K_tM_tR_tM_t^TK_t^T
\end{equation}

\subsection{$SE_2(3)$ based EKF measurement equation}
\label{subsection_equivalence}
In the $SE_2(3)$ based EKF, the measurement vector can be represented as the difference of the position expressed in the n frame calculated by GNSS and the position expressed in the n frame calculated by SINS:
\begin{equation}\label{measurement_vector_EKF}
\begin{aligned}
&\delta z_l=\tilde{r}_{GNSS}^n-\tilde{r}_{SINS}^n=r_{GNSS}^n+r_t-(\tilde{r}_{imu}^n+\tilde{C}_b^nl^b)\\
=&r_{GNSS}^n+r_t-\left(r_{imu}^n+\delta r_{imu}^n+C_b^n(I-\phi^{\tilde{b}}\times) l^b\right)\\
=&r_{GNSS}^n-\left(r_{imu}^n+C_b^n l^b\right)-\delta r_{imu}^n+C_b^n(\phi^{\tilde{b}}\times) l^b+r_t=-\delta r_{imu}^n+C_b^n(\phi^{\tilde{b}}\times) l^b+r_t\\
=&-\delta r_{eb}^n-C_b^n(l^b\times)\phi^{\tilde{b}}+r_t =\tilde{C}_b^nJ\rho_r^{\tilde{b}}-C_b^n(l^b\times)\phi^{\tilde{b}}+r_t\\
\approx & \tilde{C}_b^n\rho_r^{\tilde{b}}-\tilde{C}_b^n(l^b\times)\phi^{\tilde{b}}+r_t\triangleq H_{new}\delta x+r_t
\end{aligned}
\end{equation}
where $H_{new}$ is the $SE_2(3)$ based measurement matrix and is defined as
\begin{equation}\label{H_new}
H_{new}=\begin{bmatrix}
-\tilde{C}_b^n(l^b{\times}) & 0_{3\times3}& \tilde{C}_b^n& 0_{3\times3}& 0_{3\times3}
\end{bmatrix}
\end{equation}
It is worth noting that the innovation is expressed in the navigation frame  which is different from the invariant-innovation defined in equation(\ref{left_innovation}).

Comparing equation(\ref{innovation_bias}) and equation(\ref{H_new}) we can find
\begin{equation}\label{H_new_H_invariant}
H_{new}=\tilde{C}_b^nH_t \Rightarrow H_t=\tilde{C}_n^bH_{new}
\end{equation}

Then, by considering the Kalman filter gain in the $SE_2(3)$-based EKF, the Kalman gain in the $SE_2(3)$ based EKF can be written as
\begin{equation}\label{Kalman_gain_new}
\begin{aligned}
&K_{new}=P_tH_{new}^T\left(H_{new}P_tH_{new}^T+R_t  \right)^{-1}\\
=&P_t H_{new}^T \tilde{C}_b^n \left(\tilde{C}_n^b H_{new} P_t H_{new}^T\tilde{C}_b^n +\tilde{C}_n^bR_t\tilde{C}_b^n\right)^{-1}\tilde{C}_n^b\\
=&P_t\left( \tilde{C}_n^bH_{new}\right)^T \left(\left(\tilde{C}_n^bH_{new}\right) P_t \left( \tilde{C}_n^bH_{new}\right)^T+M_tR_tM_t^T  \right)^{-1}\tilde{C}_n^b\\
=&P_tH_t^T(H_tP_tH_t^T+M_tR_tM_t^T)^{-1}\tilde{C}_n^b=K_t\tilde{C}_n^b
\end{aligned}
\end{equation}

As all the KF algorithms execute the reset state in closed loop after each measurement update step, the error state will be set as "zero" to indicate the nominal value is the same as the estimation~\cite{shin2001accuracy}. Consequently, there is no need to implement the error state prediction step after feedback is made, and the correction of the error state can be described as
\begin{equation}\label{error_state_correction}
\hat{x}\approx K_z\tilde{z}_t+x=K_t\tilde{z}_t
\end{equation}
Substituting equation(\ref{innovation_bias}) and equation(\ref{Kalman_filter_gain_invariant}) into the above equation, we can get
\begin{equation}\label{error_state_correction_equivalence}
\begin{aligned}
&\hat{x}=K_t\tilde{z}_t=P_tH_t^T(H_tP_tH_t^T+M_tR_tM_t^T)^{-1}(H_t \delta x+M_tr_t)\\
=& K_{new}\tilde{C}_b^n(\tilde{C}_n^bH_{new} \delta x+M_tr_t)=K_{new}\tilde{C}_b^n(\tilde{C}_n^bH_{new} \delta x+\tilde{C}_n^br_t)\\
=&K_{new}(H_{new} \delta x+r_t)=K_{new}\delta z_l
\end{aligned}
\end{equation}
\begin{remark}
It is obvious that the correction of the error state by the invariant-EKF and the $SE_2(3)$ based EKF is the same. Therefore, the invariant-EKF can be viewed as the correction step of error state in the body frame but the $SE_2(3)$ based EKF executes the correction of error state in the navigation frame.
\end{remark}
\begin{remark}
	Our $SE_2(3)$ based EKF can be used in $H_{\infty}$ filtering similar to the invariant extended $H_{\infty}$ filter~\cite{lavoie2019invariant}.
\end{remark}
\section{$SE_2(3)$ based EKF for another navigation frame}
When the state defined on the matrix Lie group is given as
\begin{equation}\label{new_navigation_frame}
\mathcal{X}=\begin{bmatrix}
C_b^n & v_{ib}^n & r_{ib}^n\\
0_{1\times 3} & 1 & 0\\
0_{1\times 3} & 0 &1
\end{bmatrix}
\end{equation}
where $C_b^n$ is the direction cosine matrix from the body frame to the navigation frame; $v_{ib}^n$ is the velocity of body relative to the ECI frame expressed in the navigation frame.

Then the dynamic equation fo the state $\mathcal{X}$ can be deduced as follows
\begin{equation}\label{group_affine_property_navigation_frame}
\begin{aligned}
&\frac{d}{dt}\mathcal{X}=f_{u_t}(\mathcal{X})=\frac{d}{dt}\begin{bmatrix}
C_b^n & v_{ib}^n & r_{ib}^n\\
0_{1\times 3} & 1 & 0\\
0_{1\times 3} & 0 &1
\end{bmatrix}=\begin{bmatrix}
\dot{C}_b^n & \dot{v}_{ib}^n & \dot{r}_{ib}^n\\
0_{1\times 3} & 0 & 0\\
0_{1\times 3} & 0 &0
\end{bmatrix}\\
=&\begin{bmatrix}
C_b^n(\omega_{ib}^b\times)-(\omega_{in}^n\times)C_b^n &-\omega_{in}^n\times v_{ib}^n+C_b^n f^b+G_{ib}^n & -\omega_{in}^n\times r_{ib}^n+v_{ib}^n\\
0_{1\times 3} & 0 & 0\\
0_{1\times 3} & 0 &0
\end{bmatrix}\\
=&\begin{bmatrix}
C_b^n & v_{ib}^n & r_{ib}^n\\
0_{1\times 3} & 1 & 0\\
0_{1\times 3} & 0 &1
\end{bmatrix}\begin{bmatrix}
\omega_{ib}^b\times & f_{ib}^b & 0_{3\times 1}\\
0_{1\times 3} & 0 & 0\\
0_{1\times 3} & 0 &0
\end{bmatrix}+\begin{bmatrix}
-\omega_{in}^n\times & G_{ib}^n & v_{ib}^n\\
0_{1\times 3} & 0 & 0\\
0_{1\times 3} & 0 &0
\end{bmatrix}\begin{bmatrix}
C_b^n & v_{ib}^n & r_{ib}^n\\
0_{1\times 3} & 1 & 0\\
0_{1\times 3} & 0 &1
\end{bmatrix}\\
=&\mathcal{X}W_1+W_2\mathcal{X}
\end{aligned}
\end{equation}

It is easy to verify that the group-affine property is satisfied. 
Now, let us consider the left invariant error state and right invariant error state respectively.
\subsection{$SE_2(3)$ based EKF for navigation frame with body frame attitude error}
The left-invariant error state is defined as
\begin{equation}\label{left_invariant_new_navigation_frame}
\begin{aligned}
&\eta^L=\mathcal{X}^{-1}\tilde{\mathcal{X}}=\begin{bmatrix}
C_n^b & -C_n^bv_{ib}^n & -C_n^br_{ib}^n\\
0_{1\times 3} & 1 & 0\\
0_{1\times 3} & 0 &1
\end{bmatrix}\begin{bmatrix}
\tilde{C}_b^n & \tilde{v}_{ib}^n & \tilde{r}_{ib}^n\\
0_{1\times 3} & 1 & 0\\
0_{1\times 3} & 0 &1
\end{bmatrix}\\
=&\begin{bmatrix}
C_n^b\tilde{C}_b^n & C_n^b(\tilde{v}_{ib}^n-v_{ib}^n) & C_n^b(\tilde{r}_{ib}^n-r_{ib}^n)\\
0_{1\times 3} & 1 & 0\\
0_{1\times 3} & 0 &1
\end{bmatrix}=\begin{bmatrix}
C_n^b\tilde{C}_b^n & C_n^b\delta v_{ib}^n & C_n^b\delta r_{ib}^n\\
0_{1\times 3} & 1 & 0\\
0_{1\times 3} & 0 &1
\end{bmatrix}
\end{aligned}
\end{equation}

According to the exponential mapping from the Lie algebra to the Lie group, the error state of attitude, velocity, and position can be defined as
\begin{equation}\label{left_error_state_attitude_velocity_position}
\begin{aligned}
C_n^b\tilde{C}_b^n=&\exp_G(\phi^b\times)\approx I+\phi^b\times\\
\eta_v^L=J\rho_{v}^b=&C_n^b(\tilde{v}_{ib}^n-v_{ib}^n)=C_n^b\delta v_{ib}^n\\
\eta_r^L=J\rho_{r}^b=&C_n^b(\tilde{r}_{ib}^n-r_{ib}^n)=C_n^b\delta r_{ib}^n
\end{aligned}
\end{equation}

Therefore, the left-invariant error state satisfies that
\begin{equation}\label{left_error_state_lie_algebra}
\eta^L=\begin{bmatrix}
\exp_G(\phi^b\times) & J\rho_{v}^b & J\rho_{r}^b\\
0_{1\times 3} & 1 & 0\\
0_{1\times 3} & 0 &1
\end{bmatrix}=\exp_G\left(\begin{bmatrix}
\phi^b\times & \rho_{v}^b & \rho_{r}^b\\
0_{1\times 3} & 0 & 0\\
0_{1\times 3} & 0 &0
\end{bmatrix} \right)=\exp_G\left( \Lambda\begin{bmatrix}
\phi^b \\ \rho_{v}^b \\ \rho_{r}^b
\end{bmatrix}\right)
\end{equation}
where $\phi^b$ is the attitude error state; $J\rho_{v}^b$ is the new definition of velocity error state defined on the Lie group; $J\rho_{r}^b$ is the new definition of position error state defined on the Lie group; $\rho_{v}$ is the velocity error state defined on the Euclidean space; $\rho_{r}$ is the position error state defined on the Euclidean space.

The differential equation of the attitude error state is given as
\begin{equation}\label{attitude_error_dynamical_left}
\dot{\phi}^b=-\omega_{ib}^b\times \phi^b+\delta\omega_{ib}^b-C_n^b\delta\omega_{in}^n
\end{equation}

The differential equation of the velocity error state is given as
\begin{equation}\label{velocity_error_dynamcal_left}
\begin{aligned}
&\frac{d}{dt}(J\rho_{v}^b)=\dot{C}_n^b\delta v_{ib}^n+C_n^b(\dot{\tilde{v}}_{ib}^n-\dot{v}_{ib}^n)\\
=&(C_n^b(\omega_{in}^n\times)-(\omega_{ib}^b\times)C_n^b)\delta v_{ib}^n\\
&+C_n^b\left((-\tilde{\omega}_{in}^n\times \tilde{v}_{ib}^n+\tilde{C}_b^n \tilde{f}^b+\tilde{G}_{ib}^n)-(-\omega_{in}^n\times v_{ib}^n+C_b^n f^b+G_{ib}^n) \right)\\
=&C_n^b(\omega_{in}^n\times)\delta v_{ib}^n-(\omega_{ib}^b\times)C_n^b\delta v_{ib}^n+C_n^b\tilde{C}_b^n \tilde{f}^b-f^b+C_n^b(\tilde{G}_{ib}^n-G_{ib}^n)\\
&-C_n^b\omega_{in}^n\times(\tilde{v}_{ib}^n-v_{ib}^n)-C_n^b(\delta\omega_{in}^n\times) \tilde{v}_{ib}^n\\
=&-(\omega_{ib}^b\times)J\rho_{v}^b-\tilde{f}^b\times\phi^b+\delta f^b+C_n^b( \tilde{v}_{ib}^n \times) \delta\omega_{in}^n+C_n^b(\tilde{G}_{ib}^n-G_{ib}^n)\\
\approx &-(\omega_{ib}^b\times)J\rho_{v}^b-{f}^b\times\phi^b+\delta f^b+C_n^b( \tilde{v}_{ib}^n \times) \delta\omega_{in}^n
\end{aligned}
\end{equation}
where the second order small quantity $\delta{f}^b\times\phi^b$ is neglected; $C_n^b(\tilde{G}_{ib}^n-G_{ib}^n)$ can also be neglected.

The differential equation of the position error state is given as
\begin{equation}\label{position_error_dynamcal_left}
\begin{aligned}
&\frac{d}{dt}(J\rho_{r}^b)=\dot{C}_n^b\delta r_{ib}^n+C_n^b(\dot{\tilde{r}}_{ib}^n-\dot{r}_{ib}^n)\\
=&(C_n^b(\omega_{in}^n\times)-(\omega_{ib}^b\times)C_n^b)\delta r_{ib}^n
+C_n^b\left((-\tilde{\omega}_{in}^n\times \tilde{r}_{ib}^n+\tilde{v}_{ib}^n)-(-\omega_{in}^n\times r_{ib}^n+v_{ib}^n)  \right)\\
=&C_n^b(\omega_{in}^n\times)\delta r_{ib}^n-(\omega_{ib}^b\times)C_n^b\delta r_{ib}^n+C_n^b(\tilde{v}_{ib}^n-v_{ib}^n)-C_n^b(\delta\omega_{in}^n\times \tilde{r}_{ib}^n)-C_n^b(\omega_{in}^n\times) \delta{r}_{ib}^n\\
=&-(\omega_{ib}^b\times)J\rho_{r}^b+J\rho_{v}^b+C_n^b(\tilde{r}_{ib}^n\times )\delta\omega_{in}^n
\end{aligned}
\end{equation}
\subsection{$SE_2(3)$ based EKF for navigation frame with NED frame attitude error}
The right-invariant error is defined as
\begin{equation}\label{right_invariant_error_new_navigation}
\begin{aligned}
&\eta^R=\mathcal{X}\tilde{\mathcal{X}}^{-1}=\begin{bmatrix}
{C}_b^n & {v}_{ib}^n & {r}_{ib}^n\\
0_{1\times 3} & 1 & 0\\
0_{1\times 3} & 0 &1
\end{bmatrix}
\begin{bmatrix}
\tilde{C}_n^b & -\tilde{C}_n^b\tilde{v}_{ib}^n & -\tilde{C}_n^b\tilde{r}_{ib}^n\\
0_{1\times 3} & 1 & 0\\
0_{1\times 3} & 0 &1
\end{bmatrix}\\
=&\begin{bmatrix}
{C}_b^n\tilde{C}_n^b &{v}_{ib}^n-{C}_b^n\tilde{C}_n^b\tilde{v}_{ib}^n &{r}_{ib}^n-{C}_b^n\tilde{C}_n^b\tilde{r}_{ib}^n\\
0_{1\times 3} & 1 & 0\\
0_{1\times 3} & 0 &1
\end{bmatrix}
\end{aligned}
\end{equation}

The new error state defined on the matrix Lie group can be denoted as
\begin{equation}\label{right_invariant_error_new_lie_group}
\begin{aligned}
{C}_b^n\tilde{C}_n^b=&\exp_G(\phi^n\times)\approx I+\phi^n\times\\
\eta_v^R=J\rho_{v}^n=&{v}_{ib}^n-{C}_b^n\tilde{C}_n^b\tilde{v}_{ib}^n\approx  {v}_{ib}^n -(I+\phi^n\times)\tilde{v}_{ib}^n=\tilde{v}_{ib}^n\times \phi^n-\delta{v}_{ib}^n \\
\eta_r^R=J\rho_{r}^n=&{r}_{ib}^n-{C}_b^n\tilde{C}_n^b\tilde{r}_{ib}^n\approx  {r}_{ib}^n -(I+\phi^n\times)\tilde{r}_{ib}^n=\tilde{r}_{ib}^n\times \phi^n-\delta{r}_{ib}^n
\end{aligned}
\end{equation}

The right invariant error state can be converted to the Euclidean space as
\begin{equation}\label{right_invariant_error_lie_algebra}
\eta^R=\begin{bmatrix}
\exp_G(\phi^n\times) &J\rho_{v}^n &J\rho_{r}^n\\
0_{1\times 3} & 1 & 0\\
0_{1\times 3} & 0 &1
\end{bmatrix}=\exp_G\left(\begin{bmatrix}
\phi^n\times &\rho_{v}^n &\rho_{r}^n\\
0_{1\times 3} & 0 & 0\\
0_{1\times 3} & 0 &0
\end{bmatrix}  \right)=\exp_G\left( \Lambda\begin{bmatrix}
\phi^n\\ \rho_{v}^n \\ \rho_{r}^n
\end{bmatrix}\right)
\end{equation}
where $\phi^n$ is the attitude error state expressed in the navigation frame; $\rho_{v}$ is the velocity error state expressed in the Euclidean space; $\rho_{r}^n$ is the position error state expressed in the Euclidean space.

The differential equation of the attitude error state is given as
\begin{equation}\label{attitude_error_dynamical_right}
\dot{\phi}^b=-\omega_{ib}^b\times \phi^b-\delta\omega_{ib}^b+C_n^b\delta\omega_{in}^n
\end{equation}

The differential equation of the velocity error state is given as
\begin{equation}\label{velocity_error_dynamcal_right}
\begin{aligned}
&\frac{d}{dt}(J\rho_{v}^n)=\dot{v}_{ib}^n-\frac{d}{dt}({C}_b^n\tilde{C}_n^b)\tilde{v}_{ib}^n- {C}_b^n\tilde{C}_n^b\dot{\tilde{v}}_{ib}^n\\
=&-\omega_{in}^n\times v_{ib}^n+C_b^n f^b+G_{ib}^n-{C}_b^n\tilde{C}_n^b(-\tilde{\omega}_{in}^n\times \tilde{v}_{ib}^n+\tilde{C}_b^n \tilde{f}^b+\tilde{G}_{ib}^n)\\
&-(C_b^n(\omega_{ib}^b\times)\tilde{C}_n^b-(\omega_{in}^n\times)C_b^n\tilde{C}_n^b+C_b^n\tilde{C}_n^b(\tilde{\omega}_{in}^n\times)-C_b^n(\tilde{\omega}_{ib}^b\times)\tilde{C}_n^b)\tilde{v}_{ib}^n\\
=&-\omega_{in}^n\times(v_{ib}^n-C_b^n\tilde{C}_n^b\tilde{v}_{ib}^n)+C_b^n(f^b-\tilde{f}^b)+G_{ib}^n-C_b^n\tilde{C}_n^b\tilde{G}_{ib}^n+C_b^n(\delta\omega_{ib}^b\times)C_n^bC_b^n\tilde{C}_n^b\tilde{v}_{ib}^n\\
=&-\omega_{in}^n\times J\rho_{v}^n-C_b^n\delta f^b+G_{ib}^n\times \phi^n-\delta G_{ib}^n+(C_b^n\delta\omega_{ib}^b)\times(v_{ib}^n-J\rho_{v}^n)\\
=&-\omega_{in}^n\times J\rho_{v}^n-C_b^n\delta f^b+G_{ib}^n\times \phi^n-(v_{ib}^n\times)C_b^n\delta\omega_{ib}^b
\end{aligned}
\end{equation}
where the second order small quantity $(C_b^n\delta\omega_{ib}^b)\times J\rho_{v}^n$ is neglected; $\delta G_{ib}^n$ can also be neglected.

The differential equation of the position error state is given as
\begin{equation}\label{position_error_dynamcal_right}
\begin{aligned}
&\frac{d}{dt}(J\rho_{r}^n)=\dot{r}_{ib}^n-\frac{d}{dt}({C}_b^n\tilde{C}_n^b)\tilde{r}_{ib}^n- {C}_b^n\tilde{C}_n^b\dot{\tilde{r}}_{ib}^n\\
=&-\omega_{in}^n\times r_{ib}^n+v_{ib}^n-{C}_b^n\tilde{C}_n^b(-\tilde{\omega}_{in}^n\times \tilde{r}_{ib}^n+\tilde{v}_{ib}^n)\\
&-(C_b^n(\omega_{ib}^b\times)\tilde{C}_n^b-(\omega_{in}^n\times)C_b^n\tilde{C}_n^b+C_b^n\tilde{C}_n^b(\tilde{\omega}_{in}^n\times)-C_b^n(\tilde{\omega}_{ib}^b\times)\tilde{C}_n^b)\tilde{r}_{ib}^n\\
=&-\omega_{in}^n\times (r_{ib}^n-C_b^n\tilde{C}_n^b\tilde{r}_{ib}^n)+(v_{ib}^n-C_b^n\tilde{C}_n^b\tilde{v}_{ib}^n)+C_b^n\delta\omega_{ib}^b\times C_n^bC_b^n\tilde{C}_n^b\tilde{r}_{ib}^n\\
=&-\omega_{in}^n\times J\rho_{r}^n+J\rho_{v}^n+(C_b^n\delta\omega_{ib}^b)\times (r_{ib}^n-J\rho_{r}^n)\\
\approx &-\omega_{in}^n\times J\rho_{r}^n+J\rho_{v}^n-(r_{ib}^n\times)C_b^n\delta\omega_{ib}^b
\end{aligned}
\end{equation}
where the second order small quantity $(C_b^n\delta\omega_{ib}^b)\times J\rho_{r}^n$ is neglected
\section{$SE_2(3)$ based EKF for transformaed INS Mechanization in NED Frame}
\label{transformed_INS}
The INS mechanization in NED frame in terms of XYZ is given as
\begin{equation}\label{C_b_n_d_e_invariant}
\dot{C}_b^n=C_b^n(\omega_{ib}^b\times)-(\omega_{in}^n\times)C_b^n
\end{equation}
\begin{equation}\label{v_eb_n_d_e_invarant}
\dot{v}_{eb}^n=C_b^nf_{ib}^b-\left[ (2\omega_{ie}^n+\omega_{en}^n)\times\right]v_{eb}^n+g_{ib}^n
\end{equation}
\begin{equation}\label{r_eb_n_d_e_invariant}
\dot{r}_{eb}^n=\frac{d}{dt}(C_e^n r_{eb}^e)=\frac{d}{dt}(C_e^n )r_{eb}^e+C_e^n\dot{r}_{eb}^e=C_e^n(\omega_{ne}^e\times)r_{eb}^e+C_e^nv_{eb}^e=-\omega_{en}^n\times r_{eb}^n+v_{eb}^n
\end{equation}
where $g_{ib}^n$ is the gravity vector, and its relationship with the gravitational vector $\overline{g}^n$ is given by
\begin{equation}\label{gravitation_gravity}
g_{ib}^n={G}_{ib}^n-(\omega_{ie}^n)^2r_{eb}^n
\end{equation}

Similar to\cite{brossard2020associating}, an auxiliary velocity is introduced as
\begin{equation}\label{auxiliary_velocity}
\overline{v}_{eb}^n=v_{eb}^n+\omega_{ie}^n\times r_{eb}^n=v_{eb}^n+C_e^n\omega_{ie}^e\times r_{eb}^n
\end{equation}

With the introduced auxiliary velocity vector, the INS mechanization is now given by
\begin{equation}\label{C_b_n_d_e_invariant1}
\dot{C}_b^n=C_b^n(\omega_{ib}^b\times)-(\omega_{in}^n\times)C_b^n
\end{equation}
\begin{equation}\label{v_eb_n_d_e_invarant1}
\dot{v}_{eb}^n=C_b^nf_{ib}^b-(\omega_{in}^n)\times \overline{v}_{eb}^n+{G}_{ib}^n
\end{equation}
\begin{equation}\label{r_eb_n_d_e_invariant1}
\dot{r}_{eb}^n=-\omega_{in}^n\times r_{eb}^n+\overline{v}_{eb}^n
\end{equation}

Then defining the state composed by the attitude $C_b^n$, the velocity $\overline{v}_{eb}^n$, and the position $r_{eb}^n$ as the elements of the matrix Lie group $SE_2(3)$, that is
\begin{equation}\label{new_state}
\mathcal{X}=\begin{bmatrix}
C_b^n & \overline{v}_{eb}^n & r_{eb}^n\\
0_{1\times 3} & 1 &0\\
0_{1\times 3} & 0 & 1
\end{bmatrix}
\end{equation}

Therefore, equation(\ref{C_b_n_d_e_invariant1}), equation(\ref{v_eb_n_d_e_invarant1}), equation(\ref{r_eb_n_d_e_invariant1}) can be rewritten in a compact form as
\begin{equation}\label{differential_invariant}
\begin{aligned}
&\frac{d}{dt}\mathcal{X}=f_{u_t}(\mathcal{X})=\frac{d}{dt}\begin{bmatrix}
C_b^n & \overline{v}_{eb}^n & r_{eb}^n\\
0_{1\times3} & 1 & 0\\
0_{1\times 3} & 0& 1
\end{bmatrix}
=\begin{bmatrix}
\dot{C}_b^n & \dot{\overline{v}}_{eb}^n & \dot{r}_{eb}^n\\
0_{1\times3} & 0 & 0\\
0_{1\times 3} & 0& 0
\end{bmatrix}=\mathcal{X}W_1+W_2\mathcal{X}\\
=&\begin{bmatrix}
C_b^n(\omega_{ib}^b\times)-(\omega_{in}^n\times)C_b^n & C_b^nf_{ib}^b-(\omega_{in}^n)\times \overline{v}_{eb}^n+{G}_{ib}^n & -\omega_{in}^n\times r_{eb}^n+\overline{v}_{eb}^n\\
0_{1\times3} & 0 & 0\\
0_{1\times 3} & 0& 0
\end{bmatrix}
\end{aligned}
\end{equation}
where $W_1$ and $W_2$ are denoted as
\begin{equation}\label{W_1_W_2_invariant}
W_1=\begin{bmatrix}
\omega_{ib}^b\times & f_{ib}^b & 0\\
0_{1\times3} & 0 & 0\\
0_{1\times 3} & 0& 0
\end{bmatrix},W_2=\begin{bmatrix}
-\omega_{in}^n\times & {G}_{ib}^n & \overline{v}_{eb}^n\\
0_{1\times3} & 0 & 0\\
0_{1\times 3} & 0& 0
\end{bmatrix}
\end{equation}

It is easy to verify that the dynamical equation(\ref{differential_invariant}) satisfies the group-affine property so that the error state dynamical equation is independent of the global state.
Next, the explicit error state dynamical equation with left invariant error and right invariant error will be derived.

\subsection{Invariant Error State Dynamical Equations}
\subsubsection{Left Invariant Error State Dynamical Equations}
The left invariant error state defined on the matrix Lie group is calculated as
\begin{equation}\label{eror_state_left_invariant}
\eta^l=\tilde{\mathcal{X}}^{-1}\mathcal{X}=\begin{bmatrix}
\tilde{C}_n^bC_b^n & \tilde{C}_n^b(\overline{v}_{eb}^n-\tilde{\overline{v}}_{eb}^n) &\tilde{C}_n^b({r}_{eb}^n-\tilde{{r}}_{eb}^n)\\
0_{1\times3} & 0 & 0\\
0_{1\times 3} & 0& 0
\end{bmatrix}=\begin{bmatrix}
\eta^a & \eta^v & \eta^r\\
0_{1\times 3} &1 &0\\
0_{1\times 3} &0&1
\end{bmatrix}
\end{equation}

The error state defined on the Lie group can be converted to the corresponding Lie algebra as follows
\begin{equation}\label{eror_state_left_invariant_algebra}
\eta^l=\begin{bmatrix}
\exp_G(\phi^b) & J\rho_v^b & J\rho_r^b\\
0_{1\times3} & 0 & 0\\
0_{1\times 3} & 0& 0
\end{bmatrix}=\exp_G\left(\begin{bmatrix}
\phi^b\times &\rho_v^b & \rho_r^b \\
0_{1\times3} & 0 & 0\\
0_{1\times 3} & 0& 0
\end{bmatrix}   \right)=\exp_G\left(\Lambda\begin{bmatrix}
\phi^b \\ \rho_v^b \\ \rho_r^b
\end{bmatrix} \right)=\exp_G\left(\Lambda(\rho^b)\right)
\end{equation}

Comparing equation(\ref{eror_state_left_invariant}) and equation(\ref{eror_state_left_invariant_algebra}), we can get
\begin{equation}\label{error_state_define}
\begin{aligned}
\eta^a&=\tilde{C}_n^bC_b^n=\exp_G(\phi^b)\approx I+\phi^b\times \\
\eta^v&=J\rho_v^b=\tilde{C}_n^b(\overline{v}_{eb}^n-\tilde{\overline{v}}_{eb}^n)=-\tilde{C}_n^b\overline{v}_{eb}^n\\
\eta^r&=J\rho_r^b=\tilde{C}_n^b({r}_{eb}^n-\tilde{{r}}_{eb}^n)=-\tilde{C}_n^b{r}_{eb}^n
\end{aligned}
\end{equation}

According to the definition of auxiliary velocity in equation(\ref{auxiliary_velocity}), the velocity error state of it is calculated as
\begin{equation}\label{perturbation_auxiliary}
\begin{aligned}
&\delta \overline{v}_{eb}^n=\tilde{\overline{v}}_{eb}^n-\overline{v}_{eb}^n=\tilde{v}_{eb}^n+\tilde{\omega}_{ie}^n\times \tilde{r}_{eb}^n-\left(v_{eb}^n+C_e^n\omega_{ie}^e\times r_{eb}^n\right)\\
=&\delta v_{eb}^n+\delta\omega_{ie}^n\times r_{eb}^n+\omega_{ie}^n\times \delta r_{eb}^n+\delta\omega_{ie}^n\times \delta r_{eb}^n
\approx \delta v_{eb}^n+\delta\omega_{ie}^n\times r_{eb}^n+\omega_{ie}^n\times \delta r_{eb}^n
\end{aligned}
\end{equation}
where the second order small quantity $\delta\omega_{ie}^n\times \delta r_{eb}^n$ is neglected.

Now, we consider the differential equations for the attitude error, velocity error, and position error which can form  a element of the $SE_2(3)$ matrix Lie group. On the one hand, by taking differential of attitude error $\eta^a$ with respect to time, we can get
\begin{equation}\label{attitude_n_d_left_estimated_body_frame_invariant}
\begin{aligned}
&\frac{d}{dt}\eta^a=\frac{d}{dt}\tilde{C}_n^b C_b^n
=\dot{\tilde{C}}_n^bC_b^n+\tilde{C}_n^b\dot{C}_b^n\\
=&\left(\tilde{C}_n^b(\tilde{\omega}_{in}^n\times)-(\tilde{\omega}_{ib}^b\times)\tilde{C}_n^b \right)C_b^n+\tilde{C}_n^b\left( C_b^n(\omega_{ib}^b\times)-(\omega_{in}^n\times)C_b^n\right)\\
=&\tilde{C}_n^b(\tilde{\omega}_{in}^n\times)C_b^n-(\tilde{\omega}_{ib}^b\times)\tilde{C}_n^bC_b^n+\tilde{C}_n^bC_b^n(\omega_{ib}^b\times)-\tilde{C}_n^b(\omega_{in}^n\times)C_b^n\\
\approx& -(\tilde{\omega}_{ib}^b\times)(I_{3\times3}+\phi^b\times)+(I_{3\times3}+\phi^b\times)(\omega_{ib}^b\times)+\tilde{C}_n^b((\tilde{\omega}_{in}^n-\omega_{in}^n)\times)C_b^n  \\
=& -\tilde{\omega}_{ib}^b\times-(\tilde{\omega}_{ib}^b\times)(\phi^b\times)+\omega_{ib}^b\times+(\phi^b\times)(\omega_{ib}^b\times)+\tilde{C}_n^bC_b^nC_n^b(\delta\omega_{in}^n\times)C_b^n\\
\approx &-\delta \omega_{ib}^b\times+(\phi^b\times\omega_{ib}^b)\times -(\delta\omega_{ib}^b\times)(\phi^b\times)+\delta\omega_{in}^b\times+\phi^b\times(\delta\omega_{in}^b\times)\\
\approx &-\delta \omega_{ib}^b\times+(\phi^b\times\omega_{ib}^b)\times +\delta\omega_{in}^b\times
\end{aligned}
\end{equation}
where the  2-order small quantities $(\delta\omega_{ib}^b\times)(\phi^b\times)$ and $(\phi^b\times)(\delta\omega_{in}^b\times)$ are neglected at the last step; $\delta\omega_{in}^n$ is defined as $\delta\omega_{in}^n\triangleq \tilde{\omega}_{in}^n-\omega_{in}^n$; $\delta\omega_{ib}^b$ is defined as $\delta\omega_{ib}^b\triangleq \tilde{\omega}_{ib}^b-\omega_{ib}^b$.

On the other hand, 
\begin{equation}\label{attitude_n_d_approx_estimated_body_frame_invariant}
\frac{d}{dt}\varepsilon^a\approx \frac{d}{dt}(I_{3\times3}+\phi^{{b}}\times)=\dot{\phi}^{{b}}\times
\end{equation}

Therefore, the state error differential equation for the attitude error can be written as follows
\begin{equation}\label{attitude_n_d_estimated_body_frame_invariant}
\dot{\phi}^{{b}}=-\delta\omega_{ib}^b+(\phi^{{b}}\times\omega_{ib}^b)+\delta\omega_{in}^b=-\omega_{ib}^b\times\phi^{{b}}-\delta\omega_{ib}^b+C_n^b\delta\omega_{in}^n
\end{equation}

By taking differential of velocity error $\eta^v$ with respect to time and substituting equation (\ref{error_state_define}) into it, we can get
\begin{equation}\label{Velocity_n_d_estimated_body_frame_invariant}
\begin{aligned}
&\frac{d}{dt}\eta^v=\frac{d}{dt}\left(\tilde{C}_n^b(\overline{v}_{eb}^n-\tilde{\overline{v}}_{eb}^n)\right)
=\dot{\tilde{C}}_n^b(\overline{v}_{eb}^n-\tilde{\overline{v}}_{eb}^n)+\tilde{C}_n^b(\dot{\overline{v}}_{eb}^n-\dot{\tilde{\overline{v}}}_{eb}^n)\\
=&\left(\tilde{C}_n^b(\tilde{\omega}_{in}^n\times)-(\tilde{\omega}_{ib}^b\times)\tilde{C}_n^b \right) (\overline{v}_{eb}^n-\tilde{\overline{v}}_{eb}^n)\\
&+\tilde{C}_n^b\left[ \left(C_b^nf_{ib}^b-\omega_{in}^n\times\overline{v}_{eb}^n+{G}_{ib}^n\right)-
\left(\tilde{C}_b^n\tilde{f}_{ib}^b-\tilde{\omega}_{in}^n\times\tilde{\overline{v}}_{eb}^n+\tilde{G}_{ib}^n\right)\right]\\
=&\tilde{C}_n^b(\tilde{\omega}_{in}^n\times)\overline{v}_{eb}^n-(\tilde{\omega}_{ib}^b\times)\tilde{C}_n^b\overline{v}_{eb}^n-\tilde{C}_n^b(\tilde{\omega}_{in}^n\times)\tilde{\overline{v}}_{eb}^n+(\tilde{\omega}_{ib}^b\times)\tilde{C}_n^b\tilde{\overline{v}}_{eb}^n\\
&-\tilde{C}_n^b({\omega}_{in}^n\times)\overline{v}_{eb}^n+\tilde{C}_n^b(\tilde{\omega}_{in}^n\times)\tilde{\overline{v}}_{eb}^n+\tilde{C}_n^bC_b^nf_{ib}^b-\tilde{f}_{ib}^b+\tilde{C}_n^b(G_{ib}^n-\tilde{G}_{ib}^n)\\
\approx &\tilde{C}_n^b(\delta {\omega}_{in}^n\times)\overline{v}_{eb}^n-(\tilde{\omega}_{ib}^b\times)\tilde{C}_n^b(\overline{v}_{eb}^n-\tilde{\overline{v}}_{eb}^n)+\phi^b\times \tilde{f}_{ib}^b+\phi^b\times \delta f_{ib}^b-\delta f_{ib}^b-\tilde{C}_n^b\delta G_{ib}^n\\
\approx &-\tilde{C}_n^b(\overline{v}_{eb}^n \times)\delta {\omega}_{in}^n-\tilde{\omega}_{ib}^b\times J\rho_v^b-\tilde{f}_{ib}^b\times \phi^b-\delta f_{ib}^b
\end{aligned}
\end{equation}
where $\delta f_{ib}^b$ is defined as $\delta f_{ib}^b \triangleq \tilde{f}_{ib}^b- f_{ib}^b$; $\delta\omega_{ie}^n$ is defined as $\delta\omega_{ie}^n\triangleq \tilde{\omega}_{ie}^n-\omega_{ie}^n$; $\delta G_{ib}^n$ is defined as $\delta G_{ib}^n\triangleq \tilde{G}_{ib}^n-G_{ib}^n$ and it can be neglected as the change of $G_{ib}^n$ is quite small for carrier's local navigation; the second order small quantity $\phi^b\times \delta f_{ib}^b$ is neglected.

By taking differential of position error $\varepsilon^r$ with respect to time and substituting equation (\ref{error_state_define}) into it, we can get
\begin{equation}\label{Position_n_d_estimated_body_frame_invariant}
\begin{aligned}
&\frac{d}{dt}\eta^r= \frac{d}{dt}\left(\tilde{C}_n^b(r_{eb}^n-\tilde{r}_{eb}^n)\right)=\dot{\tilde{C}}_n^b(r_{eb}^n-\tilde{r}_{eb}^n)+\tilde{C}_n^b(\dot{{r}}_{eb}^n-\dot{\tilde{r}}_{eb}^n)\\
=&\left(\tilde{C}_n^b(\tilde{\omega}_{in}^n\times)-(\tilde{\omega}_{ib}^b\times)\tilde{C}_n^b \right)(r_{eb}^n-\tilde{r}_{eb}^n)+\tilde{C}_n^b\left[(-{\omega}_{in}^n\times {r}_{eb}^n+\overline{v}_{eb}^n)-(-\tilde{\omega}_{in}^n\times \tilde{r}_{eb}^n+\tilde{\overline{v}}_{eb}^n)\right]\\
=&\tilde{C}_n^b(\tilde{\omega}_{in}^n\times)r_{eb}^n-(\tilde{\omega}_{ib}^b\times)\tilde{C}_n^br_{eb}^n-\tilde{C}_n^b(\tilde{\omega}_{in}^n\times)\tilde{r}_{eb}^n+(\tilde{\omega}_{ib}^b\times)\tilde{C}_n^b\tilde{r}_{eb}^n\\
&-\tilde{C}_n^b({\omega}_{in}^n\times) {r}_{eb}^n+\tilde{C}_n^b\overline{v}_{eb}^n+\tilde{C}_n^b(\tilde{\omega}_{in}^n\times) \tilde{r}_{eb}^n-\tilde{C}_n^b\tilde{\overline{v}}_{eb}^n\\
=&\tilde{C}_n^b(\delta{\omega}_{in}^n\times)r_{eb}^n-(\tilde{\omega}_{ib}^b\times)\tilde{C}_n^b(r_{eb}^n-\tilde{r}_{eb}^n)+(\tilde{C}_n^b\overline{v}_{eb}^n-\tilde{C}_n^b\tilde{\overline{v}}_{eb}^n)\\
=&-\tilde{C}_n^b(r_{eb}^n\times)\delta{\omega}_{in}^n-\tilde{\omega}_{ib}^b\times J\rho_r^b+J\rho_v^b
\end{aligned}
\end{equation}

With the new definition of the attitude error, velocity error, and position error, we substitute equation (\ref{error_state_define}) into equation (\ref{perturbation_omega_en_n}):
\begin{equation}\label{perturbation_omega_ie_n_estimated_left_invariant1}
\delta\omega_{ie}^n=M_1\delta r_{eb}^n=-M_1\tilde{C}_b^nJ\rho_r^b
\end{equation}

Substituting equation(\ref{perturbation_auxiliary}) into equation(\ref{perturbation_omega_in_n}), we can get
\begin{equation}\label{perturbation_omega_in_n_estimated_left_invariant1}
\begin{aligned}
&\delta\omega_{in}^n=(M_1+M_3)\delta r_{eb}^n+M_2\delta v_{eb}^n\\
=& (M_1+M_3)\delta r_{eb}^n+M_2\left(\delta \overline{v}_{eb}^n-\delta\omega_{ie}^n\times r_{eb}^n-\omega_{ie}^n\times \delta r_{eb}^n\right)\\
=&(M_1+M_3)\delta r_{eb}^n+M_2\delta \overline{v}_{eb}^n+M_2(r_{eb}^n\times)\delta\omega_{ie}^n-M_2\omega_{ie}^n\times \delta r_{eb}^n\\
=&(M_1+M_3)\delta r_{eb}^n+M_2\delta \overline{v}_{eb}^n+M_2(r_{eb}^n\times)M_1\delta r_{eb}^n-M_2\omega_{ie}^n\times \delta r_{eb}^n\\
=&\left(M_1+M_3 +M_2(r_{eb}^n\times)M_1-M_2(\omega_{ie}^n\times)  \right)\delta r_{eb}^n+M_2\delta \overline{v}_{eb}^n\\
=&K_1\delta r_{eb}^n+K_2\delta \overline{v}_{eb}^n
=K_1(-\tilde{C}_b^nJ\rho_r^b)+K_2(-\tilde{C}_b^nJ\rho_v^b)\triangleq  -K_1\tilde{C}_b^nJ\rho_r^b-K_2\tilde{C}_b^nJ\rho_v^b
\end{aligned}
\end{equation}

Consequently, the state error dynamical equations with respect to the true body frame can be written as follows:
\begin{equation}\label{attitude_n_d_left_invariant_estimated_body_frame1}
\begin{aligned}
&\dot{\phi}^b=-\tilde{\omega}_{ib}^b\times\phi^b-(b_g+w_g)+\tilde{C}_n^b(-K_1\tilde{C}_b^nJ\rho_r^b-K_2\tilde{C}_b^nJ\rho_v^b)\\
=&-\tilde{\omega}_{ib}^b\times\phi^b-\tilde{C}_n^bK_1\tilde{C}_b^nJ\rho_r^b-\tilde{C}_n^bK_2\tilde{C}_b^nJ\rho_v^b-(b_g+w_g)
\end{aligned}
\end{equation}
\begin{equation}\label{Velocity_n_d_left_estimated_body_frame1}
\begin{aligned}
\frac{d}{dt}\eta^v=
&-(\tilde{\omega}_{ib}^b\times)J\rho_v^b -\delta f_{ib}^b- \tilde{f}_{ib}^b\times \phi^b-\tilde{C}_n^b\delta \overline{g}^n-\tilde{C}_n^b{v}_{eb}^n\times(-K_1\tilde{C}_b^nJ\rho_r^b-K_2\tilde{C}_b^nJ\rho_v^b)\\
=&-\left((\tilde{\omega}_{ib}^b\times)-\tilde{C}_n^b({v}_{eb}^n\times)K_2\tilde{C}_b^n \right)J\rho_v^b+\tilde{C}_n^b({v}_{eb}^n\times)K_1\tilde{C}_b^nJ\rho_r^b\\
&-\tilde{f}_{ib}^b \times\phi^b -\tilde{C}_n^b\delta \overline{g}^n-(b_a+w_a)
\end{aligned}
\end{equation}
\begin{equation}\label{Position_n_d_left_estimated_body_frame1}
\begin{aligned}
\frac{d}{dt}\varepsilon^r=&-\tilde{C}_n^b(r_{eb}^n\times)(-K_1\tilde{C}_b^nJ\rho_r^b-K_2\tilde{C}_b^nJ\rho_v^b)-\tilde{\omega}_{ib}^b\times J\rho_r^b+J\rho_v^b\\
=&(\tilde{C}_n^b(r_{eb}^n\times)K_1\tilde{C}_b^n-(\tilde{\omega}_{ib}^b\times))J\rho_r^b+(\tilde{C}_n^b(r_{eb}^n\times)K_2\tilde{C}_b^n+I_{3\times 3})J\rho_v^b
\end{aligned}
\end{equation}
\subsubsection{Right Invariant Error State Dynamical Equations}
The left invariant error state defined on the matrix Lie group is calculated as
\begin{equation}\label{eror_state_right_invariant}
\eta^l=\mathcal{X}\tilde{\mathcal{X}}^{-1}=\begin{bmatrix}
C_b^n\tilde{C}_n^b & \overline{v}_{eb}^n-C_b^n\tilde{C}_n^b\tilde{\overline{v}}_{eb}^n &{r}_{eb}^n-C_b^n\tilde{C}_n^b\tilde{{r}}_{eb}^n\\
0_{1\times3} & 0 & 0\\
0_{1\times 3} & 0& 0
\end{bmatrix}=\begin{bmatrix}
\eta^a & \eta^v & \eta^r\\
0_{1\times 3} &1 &0\\
0_{1\times 3} &0&1
\end{bmatrix}
\end{equation}

The error state defined on the Lie group can be converted to the corresponding Lie algebra as follows
\begin{equation}\label{eror_state_right_invariant_algebra}
\eta^r=\begin{bmatrix}
\exp_G(\phi^n) & J\rho_v^n & J\rho_r^n\\
0_{1\times3} & 0 & 0\\
0_{1\times 3} & 0& 0
\end{bmatrix}=\exp_G\left(\begin{bmatrix}
\phi^n\times &\rho_v^n & \rho_r^n \\
0_{1\times3} & 0 & 0\\
0_{1\times 3} & 0& 0
\end{bmatrix}   \right)=\exp_G\left(\Lambda\begin{bmatrix}
\phi^n \\ \rho_v^n \\ \rho_r^n
\end{bmatrix} \right)=\exp_G\left(\Lambda(\rho^n)\right)
\end{equation}

Comparing equation(\ref{eror_state_right_invariant}) and equation(\ref{eror_state_right_invariant_algebra}), we can get
\begin{equation}\label{error_state_define_right}
\begin{aligned}
\eta^a&=C_b^n\tilde{C}_n^b=\exp_G(\phi^n)\approx I+\phi^n\times \\
\eta^v&=J\rho_v^n=\overline{v}_{eb}^n-C_b^n\tilde{C}_n^b\tilde{\overline{v}}_{eb}^n\approx \overline{v}_{eb}^n-(I+\phi^n\times)\tilde{\overline{v}}_{eb}^n=\tilde{\overline{v}}_{eb}^n\times\phi^n-\delta \overline{v}_{eb}^n\\
\eta^r&=J\rho_r^n={r}_{eb}^n-C_b^n\tilde{C}_n^b\tilde{{r}}_{eb}^n\approx {r}_{eb}^n-(I+\phi^n\times)\tilde{{r}}_{eb}^n=\tilde{r}_{eb}^n\times\phi^n-\delta r_{eb}^n
\end{aligned}
\end{equation}

Now, we consider the differential equations for the attitude error, velocity error, and position error which can form  a element of the $SE_2(3)$ matrix Lie group. On the one hand, by taking differential of attitude error $\eta^a$ with respect to time, we can get
\begin{equation}\label{attitude_n_d_right_invariant}
\begin{aligned}
&\frac{d}{dt}\eta^a=\frac{d}{dt}C_b^n\tilde{C}_n^b=\dot{C}_b^n\tilde{C}_n^b+C_b^n\dot{\tilde{C}}_n^b\\
=&\left( C_b^n(\omega_{ib}^b\times)-(\omega_{in}^n\times)C_b^n\right)\tilde{C}_n^b+C_b^n\left(\tilde{C}_n^b(\tilde{\omega}_{in}^n\times)-(\tilde{\omega}_{ib}^b\times)\tilde{C}_n^b \right)\\
=&C_b^n(\omega_{ib}^b\times)\tilde{C}_n^b-(\omega_{in}^n\times)C_b^n\tilde{C}_n^b+C_b^n\tilde{C}_n^b(\tilde{\omega}_{in}^n\times)-C_b^n(\tilde{\omega}_{ib}^b\times)\tilde{C}_n^b \\
\approx&-(\omega_{in}^n\times)( I_{3\times3}+\phi^n\times)+( I_{3\times3}+\phi^n\times)((\omega_{in}^n+\delta\omega_{in}^n)\times)-C_b^n\left(\tilde{\omega}_{ib}^b-\omega_{ib}^b)\times\right)\tilde{C}_n^b\\
\approx&\delta\omega_{in}^n\times+(\phi^n\times\omega_{in}^n)\times+(\phi^n\times)(\delta\omega_{in}^n\times)-\delta\omega_{ib}^n\times(I+\phi^n\times)\\
\approx &\delta\omega_{in}^n\times+(\phi^n\times\omega_{in}^n)\times-\delta\omega_{ib}^n\times
\end{aligned}
\end{equation}
where the  2-order small quantities $(\phi^n\times)(\delta\omega_{in}^n\times)$ and $(\delta\omega_{ib}^n\times)(\phi^n\times)$ are neglected at the last step; $\delta\omega_{in}^n$ is defined as $\delta\omega_{in}^n\triangleq \tilde{\omega}_{in}^n-\omega_{in}^n$; $\delta\omega_{ib}^b$ is defined as $\delta\omega_{ib}^b\triangleq \tilde{\omega}_{ib}^b-\omega_{ib}^b$.

On the other hand, 
\begin{equation}\label{attitude_n_d_approx_invariant}
\frac{d}{dt}\eta^a\approx \frac{d}{dt}(I_{3\times3}+\phi^n\times)=\dot{\phi}^n\times
\end{equation}

Therefore, the state error differential equation for the attitude error can be written as follows
\begin{equation}\label{attitude_n_d_invariant}
\dot{\phi}^n=\delta\omega_{in}^n+(\phi^n\times\omega_{in}^n)-\delta\omega_{ib}^n=-\omega_{in}^n\times\phi^n+\delta\omega_{in}^n-{C}_b^n\delta\omega_{ib}^b
\end{equation}

By taking differential of velocity error $\eta^v$ with respect to time and substituting equation (\ref{error_state_define_right}) into it, we can get
\begin{equation}\label{Velocity_n_d_estimated_body_frame_invariant_right}
\begin{aligned}
&\frac{d}{dt}\eta^v=\frac{d}{dt}\left(\overline{v}_{eb}^n-C_b^n\tilde{C}_n^b\tilde{\overline{v}}_{eb}^n\right)
=\dot{\overline{v}}_{eb}^n-C_b^n\tilde{C}_n^b\dot{\tilde{\overline{v}}}_{eb}^n-\frac{d}{dt}(C_b^n\tilde{C}_n^b)\tilde{\overline{v}}_{eb}^n\\
=&\left(C_b^nf_{ib}^b-\omega_{in}^n\times\overline{v}_{eb}^n+G_{ib}^n\right)-C_b^n\tilde{C}_n^b
\left(\tilde{C}_b^n\tilde{f}_{ib}^b-\tilde{\omega}_{in}^n\times\tilde{\overline{v}}_{eb}^n+\tilde{G}_{ib}^n\right)\\
&-\left(C_b^n(\omega_{ib}^b\times)\tilde{C}_n^b-(\omega_{in}^n\times)C_b^n\tilde{C}_n^b+C_b^n\tilde{C}_n^b(\tilde{\omega}_{in}^n\times)-C_b^n(\tilde{\omega}_{ib}^b\times)\tilde{C}_n^b\right)\tilde{\overline{v}}_{eb}^n\\
=&-C_b^n\delta f_{ib}^b-\omega_{in}^n\times(\overline{v}_{eb}^n-C_b^n\tilde{C}_n^b\tilde{\overline{v}}_{eb}^n)+C_b^n(\delta\omega_{ib}^b\times)\tilde{C}_n^b\tilde{\overline{v}}_{eb}^n+G_{ib}^n-C_b^n\tilde{C}_n^b\tilde{G}_{ib}^n\\
=&-C_b^n\delta f_{ib}^b-\omega_{in}^n\times J\rho_v^n+C_b^n(\delta\omega_{ib}^b\times)C_n^bC_b^n\tilde{C}_n^b\tilde{\overline{v}}_{eb}^n+G_{ib}^n-C_b^n\tilde{C}_n^b\tilde{G}_{ib}^n\\
\approx& -C_b^n\delta f_{ib}^b-\omega_{in}^n\times J\rho_v^n+C_b^n(\delta\omega_{ib}^b\times)C_n^b(\overline{v}_{eb}^n-J\rho_v^n)+G_{ib}^n-(I+\phi^n\times)\tilde{G}_{ib}^n\\
\approx &-C_b^n\delta f_{ib}^b-\omega_{in}^n\times J\rho_v^n+(C_b^n\delta\omega_{ib}^b)\times\overline{v}_{eb}^n+\tilde{G}_{ib}^n\times \phi^n-\delta G_{ib}^n\\
\approx &-C_b^n\delta f_{ib}^b-\omega_{in}^n\times J\rho_v^n-(\overline{v}_{eb}^n\times)C_b^n\delta\omega_{ib}^b+\tilde{G}_{ib}^n\times \phi^n
\end{aligned}
\end{equation}
where $\delta f_{ib}^b$ is defined as $\delta f_{ib}^b \triangleq \tilde{f}_{ib}^b- f_{ib}^b$; $\delta\omega_{ie}^n$ is defined as $\delta\omega_{ie}^n\triangleq \tilde{\omega}_{ie}^n-\omega_{ie}^n$; $\delta \overline{g}^n$ is defined as $\delta G_{ib}^n\triangleq \tilde{G}_{ib}^n-G_{ib}^n$ and it can be neglected as the change of $G_{ib}^n$ is quite small for carrier's local navigation; the second order small quantity $(C_b^n\delta\omega_{ib}^b)\times J\rho_v^n$ is also neglected.

By taking differential of position error $\varepsilon^r$ with respect to time and substituting equation (\ref{error_state_define_right}) into it, we can get
\begin{equation}\label{Position_n_d_estimated_body_frame_invariant_right}
\begin{aligned}
&\frac{d}{dt}\eta^r= \frac{d}{dt}\left({r}_{eb}^n-C_b^n\tilde{C}_n^b\tilde{{r}}_{eb}^n\right)=\dot{r}_{eb}^n-C_b^n\tilde{C}_n^b\dot{\tilde{r}}_{eb}^n-\frac{d}{dt}(C_b^n\tilde{C}_n^b)\tilde{r}_{eb}^n\\
=&\left[(-{\omega}_{in}^n\times {r}_{eb}^n+\overline{v}_{eb}^n)-C_b^n\tilde{C}_n^b(-\tilde{\omega}_{in}^n\times \tilde{r}_{eb}^n+\tilde{\overline{v}}_{eb}^n)\right]\\
&-
\left(C_b^n(\omega_{ib}^b\times)\tilde{C}_n^b-(\omega_{in}^n\times)C_b^n\tilde{C}_n^b+C_b^n\tilde{C}_n^b(\tilde{\omega}_{in}^n\times)-C_b^n(\tilde{\omega}_{ib}^b\times)\tilde{C}_n^b\right)\tilde{r}_{eb}^n\\
=&-{\omega}_{in}^n\times ({r}_{eb}^n-C_b^n\tilde{C}_n^b\tilde{r}_{eb}^n)+\overline{v}_{eb}^n-C_b^n\tilde{C}_n^b\tilde{\overline{v}}_{eb}^n+C_b^n(\delta\omega_{ib}^b\times)\tilde{C}_n^b\tilde{r}_{eb}^n\\
=&-{\omega}_{in}^n\times J\rho_r^n+J\rho_v^n+C_b^n(\delta{\omega}_{ib}^b\times)C_n^bC_b^n\tilde{C}_n^b\tilde{r}_{eb}^n\\
=&-{\omega}_{in}^n\times J\rho_r^n+J\rho_v^n+(C_b^n\delta{\omega}_{ib}^b)\times(r_{eb}^n-J\rho_r^n)\\
\approx &-(r_{eb}^n\times )C_b^n\delta{\omega}_{ib}^b-{\omega}_{in}^n\times J\rho_r^n+J\rho_v^n
\end{aligned}
\end{equation}
where the second order small quantity $(C_b^n\delta{\omega}_{ib}^b)\times J\rho_r^n$ is neglected.

With the new definition of the attitude error, velocity error, and position error, we substitute equation (\ref{error_state_define_right}) into equation (\ref{perturbation_omega_en_n}):
\begin{equation}\label{perturbation_omega_ie_n_estimated_right_invariant1}
\delta\omega_{ie}^n=M_1\delta r_{eb}^n=M_1(\tilde{r}_{eb}^n\times \phi^n-J\rho_r^n)
\end{equation}

Substituting equation(\ref{perturbation_auxiliary}) into equation(\ref{perturbation_omega_in_n}), we can get
\begin{equation}\label{perturbation_omega_in_n_estimated_right_invariant1}
\begin{aligned}
&\delta\omega_{in}^n=K_1\delta r_{eb}^n+K_2\delta \overline{v}_{eb}^n
=K_1(\tilde{r}_{eb}^n\times \phi^n-J\rho_r^n)+K_2(\tilde{\overline{v}}_{eb}^n\times \phi^n-J\rho_v^n)\\
=&(K_1(\tilde{r}_{eb}^n\times)+K_2(\tilde{\overline{v}}_{eb}^n\times))\phi^n-K_1J\rho_r^n-K_2J\rho_v^n
\triangleq  Q_1\phi^n+Q_2J\rho_r^n+Q_3J\rho_v^n
\end{aligned}
\end{equation}

Consequently, the state error dynamical equations with respect to the true body frame can be written as follows:
\begin{equation}\label{attitude_n_d_right_invariant_estimated_body_frame1}
\begin{aligned}
&\dot{\phi}^b=-\omega_{in}^n\times\phi^n-{C}_b^n(b_g+w_g)+(Q_1\phi^n+Q_2J\rho_r^n+Q_3J\rho_v^n)\\
=&(Q_1-\omega_{ib}^b\times)\phi^n+Q_2J\rho_r^n+Q_3J\rho_v^n-{C}_b^n(b_g+w_g)
\end{aligned}
\end{equation}
\begin{equation}\label{Velocity_n_d_right_estimated_body_frame1}
\begin{aligned}
&\frac{d}{dt}\eta^v=
-(\tilde{\omega}_{ib}^b\times)J\rho_v^b -\delta f_{ib}^b- \tilde{f}_{ib}^b\times \phi^b-\tilde{C}_n^b\delta \overline{g}^n-\tilde{C}_n^b{v}_{eb}^n\times(Q_1\phi^n+Q_2J\rho_r^n+Q_3J\rho_v^n)\\
=&-(\tilde{C}_n^b({v}_{eb}^n\times)Q_1)\phi^n -\delta f_{ib}^b- \tilde{f}_{ib}^b\times \phi^b-\tilde{C}_n^b\delta \overline{g}^n-\tilde{C}_n^b({v}_{eb}^n\times)Q_2J\rho_r^n\\
&-((\tilde{\omega}_{ib}^b\times)+\tilde{C}_n^b({v}_{eb}^n\times)Q_3)J\rho_v^n
\end{aligned}
\end{equation}
\begin{equation}\label{Position_n_d_right_estimated_body_frame1}
\begin{aligned}
&\frac{d}{dt}\varepsilon^r=-\tilde{C}_n^b(r_{eb}^n\times)(Q_1\phi^n+Q_2J\rho_r^n+Q_3J\rho_v^n)-\tilde{\omega}_{ib}^b\times J\rho_r^n+J\rho_v^n\\
=&-\tilde{C}_n^b(r_{eb}^n\times)Q_1\phi^n-(\tilde{C}_n^b(r_{eb}^n\times)Q_2+(\tilde{\omega}_{ib}^b\times)) J\rho_r^n-{(\tilde{C}_n^b(r_{eb}^n\times)Q_3-I_{3\times3})}J\rho_v^n
\end{aligned}
\end{equation}
\subsection{Modified Error State Dynamical Equations for MEMS IMU-integrated Navigation}
On the one hand, when MEMS IMU are used for the inertial-integrated navigation, the errors of the gyroscope are greatly exceed the Earth rate and the transport rate so that the associated perturbation terms can be neglected in the dynamical equations\cite{groves2013principles}. On the other hand, the Earth rate term and the transport rate term can be calculated precisely by precise velocity and position information provided by GNSS in some applications such as initial alignment~\cite{yan2008on}.
Therefore, $\delta\omega_{in}^n$ term can be neglected in the left invariant error state dynamical equations and the right invariant error state dynamical.
\subsubsection{Modified Left Invariant Error State Dynamical Equations for consumer-grade inertial sensors}
Neglecting the $\delta\omega_{in}^n$ term in equation(\ref{attitude_n_d_estimated_body_frame_invariant}), equation(\ref{Velocity_n_d_estimated_body_frame_invariant}), and equation(\ref{Position_n_d_estimated_body_frame_invariant}), the modified left invariant error state dynamical equations for consumer-grade inertial sensors are given as
\begin{equation}\label{attitude_n_d_estimated_body_frame_invariant_modified}
\dot{\phi}^{{b}}=-\tilde{\omega}_{ib}^b\times\phi^{{b}}-\delta\omega_{ib}^b
\end{equation}
\begin{equation}\label{Velocity_n_d_estimated_body_frame_invariant_modified}
\frac{d}{dt}\eta^v=\frac{d}{dt}\left(\tilde{C}_n^b(\overline{v}_{eb}^n-\tilde{\overline{v}}_{eb}^n)\right)
\approx -\tilde{\omega}_{ib}^b\times J\rho_v^b-\tilde{f}_{ib}^b\times \phi^b-\delta f_{ib}^b
\end{equation}
\begin{equation}\label{Position_n_d_estimated_body_frame_invariant_modified}
\frac{d}{dt}\eta^r= \frac{d}{dt}\left(\tilde{C}_n^b(r_{eb}^n-\tilde{r}_{eb}^n)\right)=\dot{\tilde{C}}_n^b(r_{eb}^n-\tilde{r}_{eb}^n)+\tilde{C}_n^b(\dot{{r}}_{eb}^n-\dot{\tilde{r}}_{eb}^n)
\approx -\tilde{\omega}_{ib}^b\times J\rho_r^b+J\rho_v^b
\end{equation}

Therefor, the error state transition matrix F and the noise driven matrix G are represented as
\begin{equation}\label{left_invariant_modified}
F=\begin{bmatrix}
-\tilde{\omega}_{ib}^b\times &  0&0 & -I& 0\\
-\tilde{f}_{ib}^b & -\tilde{\omega}_{ib}^b\times & 0& 0&-I\\
0& I& -\tilde{\omega}_{ib}^b\times &0 &0\\
0&0&0&-\frac{1}{\tau_g}&0\\
0&0&0&0&-\frac{1}{\tau_a}
\end{bmatrix},
G=\begin{bmatrix}
-I &0&0&0\\
0&-I&0&0\\
0 &0&0&0\\
0&0&I&0\\
0&0&0&I
\end{bmatrix}
\end{equation}
\begin{remark}
	In fact, the perturbation term $\delta\omega_{in}^n$ is introduced in the attitude error state differential equation and then substituted into the velocity error state and the position error state. Therefore, if this term is neglected in the attitude error state differential equation, it will disappear in the error state differential equations for velocity and position.
\end{remark}
\subsubsection{Modified Right Invariant Error State Dynamical Equations for consumer-grade inertial sensors}
Neglecting the $\delta\omega_{in}^n$ term in equation(\ref{attitude_n_d_invariant}), the modified left invariant error state dynamical equations for consumer-grade inertial sensors are given as
\begin{equation}\label{attitude_n_d_invariant_modified}
\dot{\phi}^n=\delta\omega_{in}^n+(\phi^n\times\omega_{in}^n)-\delta\omega_{ib}^n\approx -\omega_{in}^n\times\phi^n-{C}_b^n\delta\omega_{ib}^b
\end{equation}
\begin{equation}\label{Velocity_n_d_estimated_body_frame_invariant_right_modified}
\frac{d}{dt}\eta^v=\frac{d}{dt}\left(\overline{v}_{eb}^n-C_b^n\tilde{C}_n^b\tilde{\overline{v}}_{eb}^n\right)
\approx -C_b^n\delta f_{ib}^b-\omega_{in}^n\times J\rho_v^n-(\overline{v}_{eb}^n\times)C_b^n\delta\omega_{ib}^b+\tilde{\overline{g}}^n\times \phi^n
\end{equation}
\begin{equation}\label{Position_n_d_estimated_body_frame_invariant_right_modified}
\frac{d}{dt}\eta^r= \frac{d}{dt}\left({r}_{eb}^n-C_b^n\tilde{C}_n^b\tilde{{r}}_{eb}^n\right)
\approx -(r_{eb}^n\times )C_b^n\delta{\omega}_{ib}^b-{\omega}_{in}^n\times J\rho_r^n+J\rho_v^n
\end{equation}

Therefor, the error state transition matrix F and the noise driven matrix G are represented as
\begin{equation}\label{right_invariant_modified}
F=\begin{bmatrix}
-\omega_{in}^n\times &  0&0 & -C_b^n& 0\\
\tilde{\overline{g}}^n\times & -\omega_{in}^n\times & 0& -(\overline{v}_{eb}^n\times)C_b^n&-C_b^n\\
0& I& -\omega_{in}^n\times & -(r_{eb}^n\times)C_b^n &0\\
0&0&0&-\frac{1}{\tau_g}&0\\
0&0&0&0&-\frac{1}{\tau_a}
\end{bmatrix},
G=\begin{bmatrix}
-C_b^n &0&0&0\\
- (\overline{v}_{eb}^n\times)C_b^n&-C_b^n&0&0\\
-(r_{eb}^n\times)C_b^n &0&0&0\\
0&0&I&0\\
0&0&0&I
\end{bmatrix}
\end{equation}

\begin{remark}
	In fact, for the initial alignment problem with the velocity and position measurement from the GNSS, the perturbation term $\delta\omega_{in}^n$ can also be neglected in the $SE_2(3)$ based EKF algorithms that have been derived above, especially for the motion alignment problem and quasi-static alignment problem..
\end{remark}
\section{$SE_2(3)$ based EKF for ECEF Navigation}
When the system  state is defined as 
	\begin{equation}\label{new_state_eb_e}
\mathcal{X}=\begin{bmatrix}
C_b^e & v_{eb}^e & r_{eb}^e\\
0_{1\times 3} & 1 & 0\\
0_{1\times 3} & 0& 1
\end{bmatrix}\in SE_2(3)
\end{equation}
	where $C_b^e$ is the direction cosine matrix from the body frame to the ECEF frame; $v_{eb}^e$ is the velocity of body frame relative to the ECEF frame expressed in the ECEF frame; $r_{eb}^e$ is the position of body frame relative to the ECEF frame expressed in the ECEF frame.
	
	Then the dynamic equation of the state $\mathcal{X}$ can be deduced as follows
		\begin{equation}\label{lie_group_state_differential_eb_e}
	\begin{aligned}
	&\frac{d}{dt}\mathcal{X}=f_{u_t}(\mathcal{X})=\frac{d}{dt}\begin{bmatrix}
	C_b^e & v_{eb}^e & r_{eb}^e\\0_{1\times 3}&1&0\\0_{1\times 3}&0&1
	\end{bmatrix}=\begin{bmatrix}
	\dot{C}_b^e & \dot{v}_{eb}^e & \dot{r}_{eb}^e\\0_{1\times 3}&0&0\\0_{1\times 3}&0&0
	\end{bmatrix}\\
	=&\begin{bmatrix}
	C_b^e(\omega_{ib}^b\times)-(\omega_{ie}^e\times)C_b^e & (-2\omega_{ie}^e\times)v_{eb}^e+C_b^ef^b+g_{ib}^e & v_{eb}^e\\
	0_{1\times 3}&0&0\\0_{1\times 3}&0&0
	\end{bmatrix}\\
	=&\begin{bmatrix}
	C_b^e & v_{eb}^e & r_{eb}^e\\0_{1\times 3}&1&0\\0_{1\times 3}&0&1
	\end{bmatrix}\begin{bmatrix}
	\omega_{ib}^b\times & f^b & 0_{3\times 1}\\ 0_{1\times 3} &0&0\\ 0_{1\times 3}&0&0
	\end{bmatrix}+\\
	&\begin{bmatrix}
	-\omega_{ie}^e\times & g_{ib}^e-\omega_{ie}^e\times v_{eb}^e & v_{eb}^e+\omega_{ie}^e\times r_{eb}^e\\ 0_{1\times 3} &0&0\\ 0_{1\times 3}&0&0
	\end{bmatrix}\begin{bmatrix}
	C_b^e & v_{eb}^e & r_{eb}^e\\0_{1\times 3}&1&0\\0_{1\times 3}&0&1
	\end{bmatrix}
	\triangleq \mathcal{X}W_1+W_2\mathcal{X}
	\end{aligned}
	\end{equation}
	
	It is easy to verify that the dynamical equation is group-affine property similar to equation(\ref{proof_invariance}). 
	
	\subsection{Left $SE_2(3)$ based EKF for ECEF Navigation with Body Frame Attitude Error}
	Considering the measurements of the GNSS are left-invariant observations for the world-centric observer, we first give the left-invariant error state differential equations.
	The left-invariant error is defined as
	\begin{equation}\label{left_invariant_error_ECEF}
	\eta^L=\mathcal{X}^{-1}\tilde{X}=\begin{bmatrix}
	C_e^b & -v_{eb}^b & -r_{eb}^b
	\\ 0_{1\times 3}&1&0\\0_{1\times 3}&0&1
	\end{bmatrix}\begin{bmatrix}
	\tilde{C}_b^e & \tilde{v}_{eb}^e & \tilde{r}_{eb}^e\\
	0_{1\times 3}&1&0\\0_{1\times 3}&0&1
	\end{bmatrix}=\begin{bmatrix}
	C_e^b\tilde{C}_b^e & C_e^b\tilde{v}_{eb}^e -v_{eb}^b& C_e^b\tilde{r}_{eb}^e-r_{eb}^b\\
	0_{1\times 3}&1&0\\0_{1\times 3}&0&1
	\end{bmatrix}
	\end{equation}
	
	According to the map form the Lie algebra to the Lie group, the error states of attitude, velocity, and position can de derived as
	\begin{equation}\label{attitude_velocity_position_ECEF}
	\begin{aligned}
	C_e^b\tilde{C}_b^e =&\exp_G(\phi^b\times)\approx I+\phi^b\times\\
	\eta_v^L=J\rho_{v}^b=&C_e^b\tilde{v}_{eb}^e -v_{eb}^b=C_e^b\tilde{v}_{eb}^e-C_e^bv_{eb}^e=C_e^b(\tilde{v}_{eb}^e -v_{eb}^e)=C_e^b\delta v_{eb}^e \\ 
	\eta_r^L=J\rho_{r}^b=&C_e^b\tilde{r}_{eb}^e-r_{eb}^b =C_e^b\tilde{r}_{eb}^e-C_e^br_{eb}^e=C_e^b(\tilde{r}_{eb}^e -r_{eb}^e)=C_e^b\delta r_{eb}^e
	\end{aligned}
	\end{equation}
	
	Meanwhile, the left-invariant error satisfies that
	\begin{equation}\label{lie_algebra_left_error_ECEF}
	\eta^L=\begin{bmatrix}
	\exp_G(\phi^b\times) & J\rho_{v}^b & J\rho_{r}^b\\
	0_{1\times 3} & 1& 0\\ 0_{1\times 3}&0&1
	\end{bmatrix}=\exp_G\left(\begin{bmatrix}
	(\phi^b)\times & \rho_{v}^b & \rho_{r}^b\\0_{1\times 3}&0&0\\0_{1\times 3}&0&0
	\end{bmatrix} \right)=\exp_G\left(\Lambda\begin{bmatrix}
	\phi^b \\ \rho_{v}^b \\ \rho_{r}^b
	\end{bmatrix} \right)
	\end{equation}
	where $\phi^b$ is the attitude error state, $J\rho_{v}^b$ is the new definition of velocity error state, $J\rho_{r}^b$ is the new definition of position error state;
	$J$ is the left Jacobian matrix given in equation(\ref{left_Jacobian_n}).
	
	The differential equation of the attitude error state is given as
	\begin{equation}\label{attitude_error_ECEF}
	\begin{aligned}
	\frac{d}{dt}(C_e^b\tilde{C}_b^e)&=\dot{C}_e^b\tilde{C}_b^e+C_e^b\dot{\tilde{C}}_b^e\\
	&=\left[C_e^b(\omega_{ie}^e\times)-(\omega_{ib}^b\times)C_e^b\right]\tilde{C}_b^e+C_e^b\left[\tilde{C}_b^e(\tilde{\omega}_{ib}^b\times)-(\tilde{\omega}_{ie}^e\times)\tilde{C}_b^e\right]\\
	&=C_e^b(\omega_{ie}^e\times)\tilde{C}_b^e-(\omega_{ib}^b\times)C_e^b\tilde{C}_b^e+C_e^b\tilde{C}_b^e(\tilde{\omega}_{ib}^b\times)-C_e^b({\omega}_{ie}^e\times)\tilde{C}_b^e\\
	&\approx -(\omega_{ib}^b\times)(I+\phi^b\times)+(I+\phi^b\times)(({\omega}_{ib}^b+\delta \omega_{ib}^b)\times)\\
	&=-(\omega_{ib}^b\times)(\phi^b\times)+(\delta \omega_{ib}^b)\times+ (\phi^b\times)(\omega_{ib}^b\times)+\phi^b\times(\delta \omega_{ib}^b)\times \\
	&\approx (\phi^b\times\omega_{ib}^b)\times+\delta \omega_{ib}^b\times= (\phi^b\times\omega_{ib}^b)\times+(\delta b_g^b+w_g^b)\times
	\end{aligned}
	\end{equation}
	where the angular velocity error of the earth's rotation can be neglected, i.e., $\tilde{\omega}_{ie}^e=\omega_{ie}^e$; and second order small quantity $(\phi^b\times)(\delta \omega_{ib}^b\times)$ is also neglected. Therefore, the equation(\ref{attitude_error_ECEF}) can be simplified as
	\begin{equation}\label{attitude_error_differential_ECEF}
	\dot{\phi}^b=\phi^b\times\omega_{ib}^b+\delta \omega_{ib}^b=-\omega_{ib}^b \times \phi^b+\delta \omega_{ib}^b=-\omega_{ib}^b\times\phi^b+\delta b_g^b+w_g^b
	\end{equation}
	
	The differential equation of the velocity error state is given as
	\begin{equation}\label{new_velocity_error_ECEF}
	\begin{aligned}
	&\frac{d}{dt}(J\rho_{v}^b)=\dot{C}_e^b\delta v_{eb}^e+C_e^b(\dot{\tilde{v}}_{eb}^e-\dot{v}_{eb}^e)\\
	=&\left[C_e^b(\omega_{ie}^e\times)-(\omega_{ib}^b\times)C_e^b\right]\delta v_{eb}^e\\
	&+C_e^b\left( \left[(-2\tilde{\omega}_{ie}^e\times)\tilde{v}_{eb}^e+\tilde{C}_b^e\tilde{f}^b+\tilde{g}_{ib}^e\right] -\left[(-2\omega_{ie}^e\times)v_{eb}^e+C_b^ef^b+g_{ib}^e\right]  \right)\\
	=&C_e^b(\omega_{ie}^e\times)\delta v_{eb}^e-(\omega_{ib}^b\times)\textcolor{red}{C_e^b\delta v_{eb}^e}+C_e^b\tilde{C}_b^e\tilde{f}^b-C_e^bC_b^ef^b-2C_e^b\omega_{ie}^e\times(\tilde{v}_{eb}^e-v_{eb}^e)\\
	&+C_e^b(\tilde{g}_{ib}^e-g_{ib}^e)\\
	\approx&-C_e^b(\omega_{ie}^e\times)\delta v_{eb}^e-(\omega_{ib}^b\times)J\rho_{v}^b+(I+\phi^b\times)({f}^b+\delta b_a^b+w_a^b)-f^b+C_e^b(\tilde{g}_{ib}^e-g_{ib}^e)\\
	=&-((C_e^b\omega_{ie}^e)\times)J\rho_{v}^b-(\omega_{ib}^b\times)J\rho_{v}^b+\phi^b\times f^b+\phi^b\times\delta f^b +C_e^b(\tilde{g}_{ib}^e-g_{ib}^e)+\delta f^b \\
	\approx &-((C_e^b\omega_{ie}^e)\times)J\rho_{v}^b -(\omega_{ib}^b\times)J\rho_{v}^b-f^b\times \phi^b+C_e^b(\tilde{g}_{ib}^e-g_{ib}^e)+\delta f^b\\
	= & -((C_e^b\omega_{ie}^e)\times)J\rho_{v}^b-(\omega_{ib}^b\times)J\rho_{v}^b-f^b\times \phi^b+C_e^b(\tilde{g}_{ib}^e-g_{ib}^e)+\delta b_a^b+w_a^b
	\end{aligned}
	\end{equation}
	where the second order small quantity $\phi^b\times\delta f^b $ is neglected; and as $g_{ib}^e$ can be approximated as constant, $C_e^b(\tilde{g}_{ib}^e-g_{ib}^e)$ can also be neglected.
	
	In the same way, the differential equation of the position error state is given as
	\begin{equation}\label{new_position_error_ECEF}
	\begin{aligned}
	&\frac{d}{dt}(J\rho_{r}^b)=\dot{C}_e^b\delta r_{eb}^e+C_e^b(\dot{\tilde{r}}_{eb}^e-\dot{r}_{eb}^e)\\
	=&\left[C_e^b(\omega_{ie}^e\times)-(\omega_{ib}^b\times)C_e^b\right]\delta r_{eb}^e
	+C_e^b\left( \tilde{v}_{eb}^e -v_{eb}^e  \right)\\
	=&C_e^b(\omega_{ie}^e\times)\delta r_{eb}^e-(\omega_{ib}^b\times)\textcolor{red}{C_e^b\delta r_{eb}^e}+C_e^b\delta v_{eb}^e\\
	=&((C_e^b\omega_{ie}^e)\times)J\rho_{r}^b-\omega_{ib}^b\times J\rho_{r}^b+J\rho_{v}^b
	\end{aligned}
	\end{equation}
	
	Thus, the inertial-integrated error state dynamic equation for the $SE_2(3)$ based EKF can be obtained 
	\begin{equation}\label{invariant_ekf_ECEF}
	\delta\dot{x}=F\delta x+Gw
	\end{equation}
	where $F$ is the error state transition matrix; $\delta x$ is the error state including the terms about bias; G is the noise driven matrix. Their definition is given as
	\begin{equation}\label{state_x_ECEF}
	\begin{aligned}
	&x=\begin{bmatrix}
	\phi^b\\ J\rho_{v}^b \\J\rho_{r}^b \\\delta b_g^b \\\delta b_a^b
	\end{bmatrix}, 
	F=\begin{bmatrix}
	-\omega_{ib}^b\times & 0 & 0& I &0\\
	-f^b\times &-(C_e^b\omega_{ie}^e)\times-\omega_{ib}^b\times&0 & 0& I\\
	0&I&(C_e^b\omega_{ie}^e)\times-\omega_{ib}^b\times&0&0\\
	0&0&0&-\frac{1}{\tau_g}&0\\
	0&0&0&0&-\frac{1}{\tau_a}
	\end{bmatrix},\\
	&G=\begin{bmatrix}
	I&0&0&0\\0&I&0&0\\0&0&0&0\\0&0&I&0\\0&0&0&I
	\end{bmatrix},
	w=\begin{bmatrix}w_g^b\\w_a^b \\ w_{b_g}^b \\ w_{b_a}^b\end{bmatrix}
	\end{aligned}
	\end{equation}
	\subsection{Left $SE_2(3)$ based EKF for ECEF Navigation with Estimated Body Frame Attitude Error}
	Considering the measurements of the GNSS are left-invariant observations for the world-centric observer, we first give the left-invariant error state differential equations.
	The left-invariant error is defined as
	\begin{equation}\label{left_invariant_error_estimated_ECEF}
	\eta^L=\tilde{\mathcal{X}}^{-1}\mathcal{X}=\begin{bmatrix}
	\tilde{C}_e^b & -\tilde{v}_{eb}^b & -\tilde{r}_{eb}^b
	\\ 0_{1\times 3}&1&0\\0_{1\times 3}&0&1
	\end{bmatrix}\begin{bmatrix}
	{C}_b^e & {v}_{eb}^e & {r}_{eb}^e\\
	0_{1\times 3}&1&0\\0_{1\times 3}&0&1
	\end{bmatrix}=\begin{bmatrix}
	\tilde{C}_e^bC_b^e & \tilde{C}_e^bv_{eb}^e -\tilde{v}_{eb}^b& \tilde{C}_e^b{r}_{eb}^e-\tilde{r}_{eb}^b\\
	0_{1\times 3}&1&0\\0_{1\times 3}&0&1
	\end{bmatrix}
	\end{equation}
	
	According to the map form the Lie algebra to the Lie group, the error states of attitude, velocity, and position can de derived as
	\begin{equation}\label{attitude_velocity_position_estimated_ECEF}
	\begin{aligned}
	\tilde{C}_e^b{C}_b^e =&\exp_G(\phi^b\times)\approx I+\phi^b\times\\
	\eta_v^L=J\rho_{v}^b=&\tilde{C}_e^b{v}_{eb}^e -\tilde{v}_{eb}^b=\tilde{C}_e^b{v}_{eb}^e-\tilde{C}_e^b\tilde{v}_{eb}^e=\tilde{C}_e^b({v}_{eb}^e -\tilde{v}_{eb}^e)=-\tilde{C}_e^b\delta v_{eb}^e \\ 
	\eta_r^L=J\rho_{r}^b=&\tilde{C}_e^b{r}_{eb}^e-\tilde{r}_{eb}^b =\tilde{C}_e^b{r}_{eb}^e-\tilde{C}_e^b\tilde{r}_{eb}^e=\tilde{C}_e^b({r}_{eb}^e -\tilde{r}_{eb}^e)=-\tilde{C}_e^b\delta r_{eb}^e
	\end{aligned}
	\end{equation}
	
	Meanwhile, the left-invariant error satisfies that
	\begin{equation}\label{lie_algebra_left_error_estimated_ECEF}
	\eta^L=\begin{bmatrix}
	\exp_G(\phi^b\times) & J\rho_{v}^b & J\rho_{r}^b\\
	0_{1\times 3} & 1& 0\\ 0_{1\times 3}&0&1
	\end{bmatrix}=\exp_G\left(\begin{bmatrix}
	(\phi^b)\times & \rho_{v}^b & \rho_{r}^b\\0_{1\times 3}&0&0\\0_{1\times 3}&0&0
	\end{bmatrix} \right)=\exp_G\left(\Lambda\begin{bmatrix}
	\phi^b \\ \rho_{v}^b \\ \rho_{r}^b
	\end{bmatrix} \right)
	\end{equation}
	where $\phi^b$ is the attitude error state, $J\rho_{v}^b$ is the new definition of velocity error state, $J\rho_{r}^b$ is the new definition of position error state;
	$J$ is the left Jacobian matrix given in equation(\ref{left_Jacobian_n}).
	
	The differential equation of the attitude error state is given as
	\begin{equation}\label{attitude_error_estimated_ECEF}
	\begin{aligned}
	\frac{d}{dt}(\tilde{C}_e^b{C}_b^e)&=\dot{\tilde{C}}_e^b{C}_b^e+\tilde{C}_e^b\dot{{C}}_b^e\\
	&=\left[\tilde{C}_e^b(\tilde{\omega}_{ie}^e\times)-(\tilde{\omega}_{ib}^b\times)\tilde{C}_e^b\right]{C}_b^e+\tilde{C}_e^b\left[{C}_b^e({\omega}_{ib}^b\times)-({\omega}_{ie}^e\times){C}_b^e\right]\\
	&=\tilde{C}_e^b(\omega_{ie}^e\times){C}_b^e-(\tilde{\omega}_{ib}^b\times)\tilde{C}_e^b{C}_b^e+\tilde{C}_e^b{C}_b^e({\omega}_{ib}^b\times)-\tilde{C}_e^b({\omega}_{ie}^e\times){C}_b^e\\
	&\approx -(\tilde{\omega}_{ib}^b\times)(I+\phi^b\times)+(I+\phi^b\times)((\tilde{\omega}_{ib}^b-\delta \omega_{ib}^b)\times)\\
	&=-(\tilde{\omega}_{ib}^b\times)(\phi^b\times)-(\delta \omega_{ib}^b)\times+ (\phi^b\times)(\tilde{\omega}_{ib}^b\times)-\phi^b\times(\delta \omega_{ib}^b)\times \\
	&\approx (\phi^b\times \tilde{\omega}_{ib}^b)\times-\delta \omega_{ib}^b\times= (\phi^b\times \tilde{\omega}_{ib}^b)\times-(\delta b_g^b+w_g^b)\times
	\end{aligned}
	\end{equation}
	where the angular velocity error of the earth's rotation can be neglected, i.e., $\tilde{\omega}_{ie}^e=\omega_{ie}^e$; and second order small quantity $(\phi^b\times)(\delta \omega_{ib}^b\times)$ is also neglected. Therefore, the equation(\ref{attitude_error_ECEF}) can be simplified as
	\begin{equation}\label{attitude_error_differential_estimated_ECEF}
	\dot{\phi}^b=\phi^b\times \tilde{\omega}_{ib}^b-\delta \omega_{ib}^b=-\tilde{\omega}_{ib}^b \times \phi^b-\delta \omega_{ib}^b=-\tilde{\omega}_{ib}^b\times\phi^b-\delta b_g^b-w_g^b
	\end{equation}
	
	The differential equation of the velocity error state is given as
	\begin{equation}\label{new_velocity_error_estimated_ECEF}
	\begin{aligned}
	&\frac{d}{dt}(J\rho_{v}^b)=-\dot{\tilde{C}}_e^b\delta v_{eb}^e+\tilde{C}_e^b(\dot{{v}}_{eb}^e-\dot{\tilde{v}}_{eb}^e)\\
	=&-\left[\tilde{C}_e^b(\tilde{\omega}_{ie}^e\times)-(\tilde{\omega}_{ib}^b\times)\tilde{C}_e^b\right]\delta v_{eb}^e\\
	&+\tilde{C}_e^b\left(\left[(-2\omega_{ie}^e\times)v_{eb}^e+C_b^ef^b+g_{ib}^e\right]- \left[(-2\tilde{\omega}_{ie}^e\times)\tilde{v}_{eb}^e+\tilde{C}_b^e\tilde{f}^b+\tilde{g}_{ib}^e\right]\right)\\
	=&-\tilde{C}_e^b(\omega_{ie}^e\times)\delta v_{eb}^e+(\tilde{\omega}_{ib}^b\times)\textcolor{red}{\tilde{C}_e^b\delta v_{ib}^e}+\tilde{C}_e^b{C}_b^e{f}^b-\tilde{f}^b-2\tilde{C}_e^b\omega_{ie}^e\times({v}_{eb}^e-\tilde{v}_{eb}^e)\\
	&+\tilde{C}_e^b({g}_{ib}^e-\tilde{g}_{ib}^e)\\
	\approx&\tilde{C}_e^b(\omega_{ie}^e\times)\delta v_{eb}^e-(\tilde{\omega}_{ib}^b\times)J\rho_{v}^b+(I+\phi^b\times)(\tilde{f}^b-\delta b_a^b-w_a^b)-\tilde{f}^b+\tilde{C}_e^b({g}_{ib}^e-\tilde{g}_{ib}^e)\\
	=&-((\tilde{C}_e^b\omega_{ie}^e)\times)J\rho_{v}^b-(\tilde{\omega}_{ib}^b\times)J\rho_{v}^b+\phi^b\times \tilde{f}^b-\phi^b\times\delta f^b +\tilde{C}_e^b({g}_{ib}^e-\tilde{g}_{ib}^e)-\delta f^b \\
	\approx &-((\tilde{C}_e^b\omega_{ie}^e)\times)J\rho_{v}^b -(\omega_{ib}^b\times)J\rho_{v}^b-f^b\times \phi^b+\tilde{C}_e^b({g}_{ib}^e-\tilde{g}_{ib}^e)+\delta f^b\\
	= &-((\tilde{C}_e^b\omega_{ie}^e)\times)J\rho_{v}^b -(\omega_{ib}^b\times)J\rho_{v}^b-f^b\times \phi^b+\tilde{C}_e^b({g}_{ib}^e-\tilde{g}_{ib}^e)-\delta b_a^b-w_a^b
	\end{aligned}
	\end{equation}
	where the second order small quantity $\phi^b\times\delta f^b $ is neglected; and as $g_{ib}^e$ can be approximated as constant, $\tilde{C}_e^b({g}_{ib}^e-\tilde{g}_{ib}^e)$ can also be neglected.
	
	In the same way, the differential equation of the position error state is given as
	\begin{equation}\label{new_position_error_estimated_ECEF}
	\begin{aligned}
	&\frac{d}{dt}(J\rho_{r}^b)=-\dot{\tilde{C}}_e^b\delta r_{eb}^e+\tilde{C}_e^b(\dot{{r}}_{eb}^e-\dot{\tilde{r}}_{eb}^e)\\
	=&-\left[\tilde{C}_e^b(\omega_{ie}^e\times)-(\tilde{\omega}_{ib}^b\times)\tilde{C}_e^b\right]\delta r_{eb}^e
	+\tilde{C}_e^b\left(v_{eb}^e- \tilde{v}_{eb}^e\right)\\
	=&-\tilde{C}_e^b(\omega_{ie}^e\times)\delta r_{eb}^e+(\tilde{\omega}_{ib}^b\times)\textcolor{red}{\tilde{C}_e^b\delta r_{eb}^e}-\tilde{C}_e^b\delta {v}_{eb}^e\\
	=&((\tilde{C}_e^b\omega_{ie}^e)\times)J\rho_{r}^b-\tilde{\omega}_{ib}^b\times J\rho_{r}^b+J\rho_{v}^b
	\end{aligned}
	\end{equation}
	
	Thus, the inertial-integrated error state dynamic equation for the $SE_2(3)$ based EKF can be obtained 
	\begin{equation}\label{invariant_ekf_estimated_ECEF}
	\delta\dot{x}=F\delta x+Gw
	\end{equation}
	where $F$ is the error state transition matrix; $\delta x$ is the error state including the terms about bias; G is the noise driven matrix. Their definition is given as
	\begin{equation}\label{state_x_estimated_ECEF}
	\begin{aligned}
	&\delta x=\begin{bmatrix}
	\phi^b\\ J\rho_{v}^b \\J\rho_{r}^b \\\delta b_g^b \\\delta b_a^b
	\end{bmatrix}, 
	F=\begin{bmatrix}
	-\tilde{\omega}_{ib}^b\times & 0 & 0& -I_{3\times 3} &0\\
	-\tilde{f}^b\times &-(\tilde{C}_e^b\omega_{ie}^e)\times-\tilde{\omega}_{ib}^b\times&0 & 0& -I_{3\times 3}\\
	0&I_{3\times 3}&(\tilde{C}_e^b\omega_{ie}^e)\times-\tilde{\omega}_{ib}^b\times&0&0\\
	0&0&0&-\frac{1}{\tau_g}&0\\
	0&0&0&0&-\frac{1}{\tau_a}
	\end{bmatrix},\\
	&
	G=\begin{bmatrix}
	-I_{3\times 3}&0&0&0\\0&-I_{3\times 3}&0&0\\0&0&0&0\\0&0&I_{3\times 3}&0\\0&0&0&I_{3\times 3}
	\end{bmatrix},
	w=\begin{bmatrix}w_g^b\\w_a^b \\ w_{b_g}^b \\ w_{b_a}^b\end{bmatrix}
	\end{aligned}
	\end{equation}
	
	\subsection{Right $SE_2(3)$ based EKF with estimated ECEF frame attitude error}
We give the $SE_2(3)$ based EKF with ECEF frame attitude error here.
	If the error state is converted to the estimated ECEF frame, i.e., $\eta=(\tilde{\mathcal{X}}R)({\mathcal{X}}R)^{-1}=\tilde{\mathcal{X}}{\mathcal{X}}^{-1}\in SE_2(3)$, then the right invariant error is defined as
	\begin{equation}\label{key_ECFF}
	\begin{aligned}
	\eta^R&=\tilde{\mathcal{X}}\mathcal{X}^{-1}=
	\begin{bmatrix}
	\tilde{C}_b^e & \tilde{v}_{eb}^e & \tilde{r}_{eb}^e\\0_{1\times 3}&1&0\\0_{1\times 3}&0&1
	\end{bmatrix}
	\begin{bmatrix}
	C_e^b & -v_{eb}^b & -r_{eb}^b\\0_{1\times 3}&1&0\\0_{1\times 3}&0&1
	\end{bmatrix}
	=\begin{bmatrix}
	\tilde{C}_b^eC_e^b & \tilde{v}_{eb}^e -\tilde{C}_b^ev_{eb}^b& \tilde{r}_{eb}^e-\tilde{C}_b^er_{eb}^b\\
	0_{1\times 3}&1&0\\0_{1\times 3}&0&1
	\end{bmatrix}
	\end{aligned}
	\end{equation}
	
	Similarity, the new error state defined on the matrix Lie group $SE_2(3)$ can be denoted as
	\begin{equation}\label{attitude_velocity_position_right_ECEF}
	\begin{aligned}
	\tilde{C}_b^eC_e^b =&\exp_G(\phi^e\times)\approx I+\phi^e\times\\
	\eta_v^R=J\rho_{v}^e=&\tilde{v}_{eb}^e -\tilde{C}_b^ev_{eb}^b=\tilde{v}_{eb}^e- v_{eb}^e+v_{eb}^e-\tilde{C}_b^eC_e^bv_{eb}^e=\delta v_{eb}^e+(I-\exp_G(\phi^e\times)) v_{eb}^e\\ 
	\eta_r^R=J\rho_{r}^e=&\tilde{r}_{eb}^e-\tilde{C}_b^er_{eb}^b=\tilde{r}_{eb}^e- r_{eb}^e+r_{eb}^e-\tilde{C}_b^eC_e^br_{eb}^e=\delta r_{eb}^e+(I-\exp_G(\phi^e\times)) r_{eb}^e
	\end{aligned}
	\end{equation}
	
	Meanwhile, the right invariant error satisfies that
	\begin{equation}\label{lie_algebra_right_error_ECEF}
	\eta^R=\begin{bmatrix}
	\exp_G(\phi^e\times) & J\rho_{v}^e & J\rho_{r}^e\\
	0_{1\times 3} & 1& 0\\ 0_{1\times 3}&0&1
	\end{bmatrix}=\exp_G\left( \begin{bmatrix}
	(\phi^e\times) & \rho_{v}^e & \rho_{r}^e\\
	0_{1\times 3} & 0& 0\\ 0_{1\times 3}&0&0
	\end{bmatrix}\right)=\exp_G\left( \Lambda\begin{bmatrix}
	\phi^e \\ \rho_{v}^e \\ \rho_{r}^e
	\end{bmatrix} \right)
	\end{equation}
	where $\phi^e$ is the attitude error state; $J\rho_v^e$ is the new definition of velocity error state; $J\rho_r^e$ is the new definition of position error state.
	
	The differential equation of the attitude error state is given as
	\begin{equation}\label{attitude_error_right_ECEF}
	\begin{aligned}
	&\frac{d}{dt}(\tilde{C}_b^eC_e^b)=\dot{\tilde{C}}_b^eC_e^b+\tilde{C}_b^e\dot{C}_e^b\\
	=&\left[\tilde{C}_b^e(\tilde{\omega}_{ib}^b\times)-(\tilde{\omega}_{ie}^e\times)\tilde{C}_b^e\right]C_e^b+\tilde{C}_b^e\left[C_e^b(\omega_{ie}^e\times)-(\omega_{ib}^b\times)C_e^b\right]\\
	=&\tilde{C}_b^e(\tilde{{\omega}}_{ib}^b\times)C_e^b-({\omega}_{ie}^e\times)\tilde{C}_b^eC_e^b+\tilde{C}_b^eC_e^b(\omega_{ie}^e\times)-\tilde{C}_b^e(\omega_{ib}^b\times)C_e^b\\
	\approx &\tilde{C}_b^e(\delta{\omega}_{ib}^b\times)C_e^b-({\omega}_{ie}^e\times)(I+\phi^e\times)+(I+\phi^e\times)(\omega_{ie}^e\times)\\
	=&\tilde{C}_b^e(\delta \omega_{ib}^b\times)\textcolor{red}{\tilde{C}_e^b\tilde{C}_b^eC_e^b}-({\omega}_{ie}^e\times)(\phi^e\times)+(\phi^e\times)(\omega_{ie}^e\times)\\
	\approx& ((\tilde{C}_b^e\delta \omega_{ib}^b)\times)(I+\phi^e\times)+((\phi^e\times\omega_{ie}^e)\times)\\
	\approx&(\tilde{C}_b^e\delta \omega_{ib}^b)\times+(\phi^e\times\omega_{ie}^e)\times=(\tilde{C}_b^e(\delta b_g^b+w_g^b))\times+(\phi^e\times\omega_{ie}^e)\times
	\end{aligned}
	\end{equation}
	where the second order small quantity $\left((\tilde{C}_b^e\delta \omega_{ib}^b)\times\right)(\phi^e\times)$ is neglected.
	Therefore, the equation(\ref{attitude_error_right_ECEF}) can be simplified as
	\begin{equation}\label{attitude_error_differential_right_ECEF}
	\dot{\phi}^e=\phi^e\times\omega_{ie}^e+\tilde{C}_b^e\delta b_g^b+\tilde{C}_b^ew_g^b=\textcolor{red}{-\omega_{ie}^e\times\phi^e}+\tilde{C}_b^e\delta b_g^b+\tilde{C}_b^ew_g^b
	\end{equation}
	
	The differential equation of the velocity error state is given as
	\begin{equation}\label{new_velocity_error_right_ECEF}
	\begin{aligned}
	&\frac{d}{dt}(J\rho_{v}^e)=\frac{d}{dt}(\tilde{v}_{eb}^e-\tilde{C}_b^eC_e^bv_{eb}^e)=\dot{\tilde{v}}_{eb}^e -\tilde{C}_b^eC_e^b\dot{v}_{eb}^e-\frac{d}{dt}(\tilde{C}_b^eC_e^b)v_{eb}^e\\
	=&\left[(-2\tilde{\omega}_{ie}^e\times)\tilde{v}_{eb}^e+\tilde{C}_b^e\tilde{f}^b+\tilde{g}_{ib}^e\right] -\tilde{C}_b^eC_e^b\left[(-2\omega_{ie}^e\times)v_{eb}^e+C_b^ef^b+g_{ib}^e\right]\\
	&-\left(\tilde{C}_b^e(\delta \omega_{ib}^b\times)\tilde{C}_e^b\tilde{C}_b^eC_e^b-({\omega}_{ie}^e\times)\tilde{C}_b^eC_e^b+\tilde{C}_b^eC_e^b(\omega_{ie}^e\times)\right)v_{eb}^e\\
	=&\textcolor{red}{\tilde{C}_b^e\delta f^b}-2({\omega}_{ie}^e\times)(\tilde{v}_{eb}^e-\tilde{C}_b^eC_e^bv_{eb}^e)
	-\left((\tilde{C}_b^e\delta \omega_{ib}^b)\times\right)\textcolor{red}{\tilde{C}_b^eC_e^b v_{eb}^e}+\tilde{g}_{ib}^e-\tilde{C}_b^eC_e^bg_{ib}^e\\
	&-(v_{eb}^e\times)(\omega_{ie}^e\times)\phi^e\\
	\approx&\tilde{C}_b^e\delta f^b-2({\omega}_{ie}^e\times)J\rho_v^e
	-(\tilde{C}_b^e\delta \omega_{ib}^b)\times \textcolor{red}{(\tilde{v}_{eb}^e-J\rho_v^e)}+\tilde{g}_{ib}^e-(I+\phi^e\times)g_{ib}^e\\
	&-(v_{eb}^e\times)(\omega_{ie}^e\times)\phi^e\\
	\approx&g_{ib}^e\times \phi^e\textcolor{red}{-2({\omega}_{ie}^e\times)J\rho_v^e}+ \tilde{v}_{eb}^e\times(\tilde{C}_b^e\delta \omega_{ib}^b)+\tilde{C}_b^e\delta f^b+\tilde{g}_{ib}^e-g_{ib}^e
	-(v_{eb}^e\times)(\omega_{ie}^e\times)\phi^e\\
	=&g_{ib}^e\times \phi^e\textcolor{red}{-2({\omega}_{ie}^e\times)J\rho_v^e}+ \tilde{v}_{eb}^e\times(\tilde{C}_b^e(\delta b_g^b+w_g^b))+\tilde{C}_b^e(\delta b_a^b+w_a^b)+\tilde{g}_{ib}^e-g_{ib}^e\\
	&-(v_{eb}^e\times)(\omega_{ie}^e\times)\phi^e\\
	\end{aligned}
	\end{equation}
	where the second order small quantity $(J\rho_v^e\times)(\tilde{C}_b^e\delta \omega_{ib}^b)$ is neglected; and as $g_{ib}^e$ can be approximated as constant, $\tilde{g}_{ib}^e-g_{ib}^e$ can also be neglected.
	
	In the same way,the differential equation of the position error state is given as
	\begin{equation}\label{new_position_error_right_ECEF}
	\begin{aligned}
	&\frac{d}{dt}(J\rho_{r}^e)=\frac{d}{dt}(\tilde{r}_{eb}^e-\tilde{C}_b^eC_e^br_{eb}^e)=\dot{\tilde{r}}_{eb}^e -\tilde{C}_b^eC_e^b\dot{r}_{eb}^e - \frac{d}{dt}(\tilde{C}_b^eC_e^b)r_{eb}^e\\
	=&\tilde{v}_{eb}^e -\tilde{C}_b^eC_e^bv_{eb}^e
	-\left(\tilde{C}_b^e(\delta \omega_{ib}^b\times)\tilde{C}_e^b\tilde{C}_b^eC_e^b-({\omega}_{ie}^e\times)\tilde{C}_b^eC_e^b+\tilde{C}_b^eC_e^b(\omega_{ie}^e\times)\right)r_{eb}^e\\
	\approx&-(r_{eb}^e\times)(\omega_{ie}^e\times)\phi^e+(\tilde{v}_{eb}^e-\tilde{C}_b^eC_e^bv_{eb}^e)-((\tilde{C}_b^e\delta \omega_{ib}^b)\times)\textcolor{red}{\tilde{C}_b^eC_e^br_{eb}^e}\\
	\approx &-(r_{eb}^e\times)(\omega_{ie}^e\times)\phi^e+J\rho_v^e+((\tilde{C}_b^e\delta \omega_{ib}^b)\times) \textcolor{red}{(\tilde{r}_{ib}^e-J\rho_r^e)}\\
	\approx&-(r_{eb}^e\times)(\omega_{ie}^e\times)\phi^e+J\rho_v^e+\tilde{r}_{eb}^e\times(\tilde{C}_b^e\delta \omega_{ib}^b)\\
	=&-(r_{eb}^e\times)(\omega_{ie}^e\times)\phi^e+J\rho_v^e+\tilde{r}_{eb}^e\times(\tilde{C}_b^e(\delta b_g^b+w_g^b))
	\end{aligned}
	\end{equation}
	where the second order small quantity $(J\rho_r^e\times)(C_b^e\delta \omega_{ib}^b)$ is neglected.
	
	The difference of the error state differential equations between the $SE_2(3)$ based EKF with ECEF frame attitude and the $SE_2(3)$ based EKF with estimated ECEF frame attitude lies in the $\delta f_{ib}^b$ term and the $\delta \omega_{ib}^b$ term.
	Thus, the error state $\delta x$, the error state transition matrix $F$, and the noise driven matrix $G$ of the inertial-integrated error state dynamic equation for $SE_2(3)$ based EKF with estimated body frame attitude are represented as
	\begin{equation}\label{state_x_right_ECEF}
	\begin{aligned}
	&x=\begin{bmatrix}
	\phi^e\\ J\rho_{v}^e \\J\rho_{r}^e \\\delta b_g^b \\\delta b_a^b
	\end{bmatrix}, 
	F=\begin{bmatrix}
	-\omega_{ie}^e\times & 0 & 0& \tilde{C}_b^e &0\\
	-(v_{eb}^e\times)(\omega_{ie}^e\times)+g_{ib}^e\times &-2\omega_{ie}^e\times&0 & \tilde{v}_{eb}^e\times \tilde{C}_b^e& \tilde{C}_b^e\\
	-(r_{eb}^e\times)(\omega_{ie}^e\times)&I&0&\tilde{r}_{eb}^e\times \tilde{C}_b^e&0\\
	0&0&0&-\frac{1}{\tau_g}&0\\
	0&0&0&0&-\frac{1}{\tau_a}
	\end{bmatrix},\\
	&G=\begin{bmatrix}
	\tilde{C}_b^e&0&0&0\\ \tilde{v}_{eb}^e\times \tilde{C}_b^e &\tilde{C}_b^e&0&0\\ \tilde{r}_{eb}^e\times \tilde{C}_b^e&0&0&0\\0&0&I&0\\0&0&0&I
	\end{bmatrix}
	\end{aligned}
	\end{equation}
	
		\subsection{Right $SE_2(3)$ based EKF with ECEF frame attitude error}
	We give the $SE_2(3)$ based EKF with estimated ECEF frame attitude error here.
	If the error state is converted to the true ECEF frame, i.e., $\eta=({\mathcal{X}}R)(\tilde{\mathcal{X}}R)^{-1}={\mathcal{X}}\tilde{\mathcal{X}}^{-1}\in SE_2(3)$, then the right invariant error is defined as
	\begin{equation}\label{key_ECFF_r}
	\begin{aligned}
	\eta^R&={\mathcal{X}}\tilde{\mathcal{X}}^{-1}=
	\begin{bmatrix}
	{C}_b^e & {v}_{eb}^e & {r}_{eb}^e\\0_{1\times 3}&1&0\\0_{1\times 3}&0&1
	\end{bmatrix}
	\begin{bmatrix}
	\tilde{C}_e^b & -\tilde{v}_{eb}^b & -\tilde{r}_{eb}^b\\0_{1\times 3}&1&0\\0_{1\times 3}&0&1
	\end{bmatrix}
	=\begin{bmatrix}
	{C}_b^e\tilde{C}_e^b & {v}_{eb}^e -{C}_b^e\tilde{v}_{eb}^b& {r}_{eb}^e-{C}_b^e\tilde{r}_{eb}^b\\
	0_{1\times 3}&1&0\\0_{1\times 3}&0&1
	\end{bmatrix}
	\end{aligned}
	\end{equation}
	
	Similarity, the new error state defined on the matrix Lie group $SE_2(3)$ can be denoted as
	\begin{equation}\label{attitude_velocity_position_right_ECEF_r}
	\begin{aligned}
	{C}_b^e\tilde{C}_e^b =&\exp_G(\phi^e\times)\approx I+\phi^e\times\\
	\eta_v^R=J\rho_{v}^e=&{v}_{eb}^e-{C}_b^e\tilde{v}_{eb}^b={C}_b^e{v}_{eb}^b-{C}_b^e\tilde{v}_{eb}^b=-{C}_b^e\delta {v}_{eb}^b \\ 
	\eta_r^R=J\rho_{r}^e=&{r}_{eb}^e-{C}_b^e\tilde{r}_{eb}^b={C}_b^e{r}_{eb}^b-{C}_b^e\tilde{r}_{eb}^b=-{C}_b^e\delta {r}_{eb}^b
	\end{aligned}
	\end{equation}
	
	Meanwhile, the right invariant error satisfies that
	\begin{equation}\label{lie_algebra_right_error_ECEF_r}
	\eta^R=\begin{bmatrix}
	\exp_G(\phi^e\times) & J\rho_{v}^e & J\rho_{r}^e\\
	0_{1\times 3} & 1& 0\\ 0_{1\times 3}&0&1
	\end{bmatrix}=\exp_G\left( \begin{bmatrix}
	(\phi^e\times) & \rho_{v}^e & \rho_{r}^e\\
	0_{1\times 3} & 0& 0\\ 0_{1\times 3}&0&0
	\end{bmatrix}\right)=\exp_G\left( \Lambda\begin{bmatrix}
	\phi^e \\ \rho_{v}^e \\ \rho_{r}^e
	\end{bmatrix} \right)
	\end{equation}
	where $\phi^e$ is the attitude error state; $J\rho_v^e$ is the new definition of velocity error state; $J\rho_r^e$ is the new definition of position error state.
	
	The differential equation of the attitude error state is given as
	\begin{equation}\label{attitude_error_right_ECEF_r}
	\begin{aligned}
	&\frac{d}{dt}({C}_b^e\tilde{C}_e^b)=\dot{{C}}_b^e\tilde{C}_e^b+{C}_b^e\dot{\tilde{C}}_e^b\\
	=&\left[{C}_b^e({\omega}_{ib}^b\times)-({\omega}_{ie}^e\times){C}_b^e\right]\tilde{C}_e^b+{C}_b^e\left[\tilde{C}_e^b(\tilde{\omega}_{ie}^e\times)-(\tilde{\omega}_{ib}^b\times)\tilde{C}_e^b\right]\\
	=&{C}_b^e({{\omega}}_{ib}^b\times)\tilde{C}_e^b-(\omega_{ie}^e\times){C}_b^e\tilde{C}_e^b+{C}_b^e\tilde{C}_e^b(\omega_{ie}^e\times)-{C}_b^e(\tilde{\omega}_{ib}^b\times)\tilde{C}_e^b\\
	\approx &-{C}_b^e(\delta{\omega}_{ib}^b\times)\tilde{C}_e^b-({\omega}_{ie}^e\times)(I+\phi^e\times)+(I+\phi^e\times)(\omega_{ie}^e\times)\\
	=&-{C}_b^e(\delta \omega_{ib}^b\times)\textcolor{red}{{C}_e^b{C}_b^e\tilde{C}_e^b}-({\omega}_{ie}^e\times)(\phi^e\times)+(\phi^e\times)(\omega_{ie}^e\times)\\
	\approx& -(({C}_b^e\delta \omega_{ib}^b)\times)(I+\phi^e\times)+((\phi^e\times\omega_{ie}^e)\times)\\
	\approx&-({C}_b^e\delta \omega_{ib}^b)\times+(\phi^e\times\omega_{ie}^e)\times
	=-({C}_b^e(\delta b_g^b+w_g^b))\times+(\phi^e\times\omega_{ie}^e)\times
	\end{aligned}
	\end{equation}
	where the second order small quantity $\left(({C}_b^e\delta \omega_{ib}^b)\times\right)(\phi^e\times)$ is neglected.
	Therefore, the equation(\ref{attitude_error_right_ECEF_r}) can be simplified as
	\begin{equation}\label{attitude_error_differential_right_ECEF_r}
	\dot{\phi}^e=\phi^e\times\omega_{ie}^e-{C}_b^e\delta b_g^b-{C}_b^ew_g^b=\textcolor{red}{-\omega_{ie}^e\times\phi^e}-{C}_b^e\delta b_g^b-{C}_b^ew_g^b
	\end{equation}
	
	The differential equation of the velocity error state is given as
	\begin{equation}\label{new_velocity_error_right_ECEF_r}
	\begin{aligned}
	&\frac{d}{dt}(J\rho_{v}^e)=\frac{d}{dt}({v}_{eb}^e-{C}_b^e\tilde{C}_e^b\tilde{v}_{eb}^e)=\dot{{v}}_{eb}^e -{C}_b^e\tilde{C}_e^b\dot{\tilde{v}}_{eb}^e-\frac{d}{dt}({C}_b^e\tilde{C}_e^b)\tilde{v}_{eb}^e\\
	=&\left[(-2{\omega}_{ie}^e\times){v}_{eb}^e+{C}_b^e{f}^b+{g}_{ib}^e\right] -{C}_b^e\tilde{C}_e^b\left[(-2\tilde{\omega}_{ie}^e\times)\tilde{v}_{eb}^e+\tilde{C}_b^e\tilde{f}^b+\tilde{g}_{ib}^e\right]\\
	&-\left(-{C}_b^e(\delta \omega_{ib}^b\times){C}_e^b{C}_b^e\tilde{C}_e^b-({\omega}_{ie}^e\times){C}_b^e\tilde{C}_e^b+{C}_b^e\tilde{C}_e^b(\omega_{ie}^e\times)\right)\tilde{v}_{eb}^e\\
	=&-\textcolor{red}{{C}_b^e\delta f^b}-2({\omega}_{ie}^e\times)({v}_{eb}^e-{C}_b^e\tilde{C}_e^b\tilde{v}_{eb}^e)
	+\left(({C}_b^e\delta \omega_{ib}^b)\times\right)\textcolor{red}{{C}_b^e\tilde{C}_e^b \tilde{v}_{eb}^e}+{g}_{ib}^e-{C}_b^e\tilde{C}_e^b\tilde{g}_{ib}^e\\
	&-(\tilde{v}_{eb}^e\times)(\omega_{ie}^e\times)\phi^e\\
	\approx&-{C}_b^e\delta f^b-2({\omega}_{ie}^e\times)J\rho_v^e
	+({C}_b^e\delta \omega_{ib}^b)\times \textcolor{red}{({v}_{eb}^e-J\rho_v^e)}+{g}_{ib}^e-(I+\phi^e\times)\tilde{g}_{ib}^e\\
	&-(\tilde{v}_{eb}^e\times)(\omega_{ie}^e\times)\phi^e\\
	\approx&\tilde{G}_{ib}^e\times \phi^e\textcolor{red}{-2({\omega}_{ie}^e\times)J\rho_v^e}- {v}_{eb}^e\times({C}_b^e\delta \omega_{ib}^b)-{C}_b^e\delta f^b-\tilde{g}_{ib}^e+g_{ib}^e
	-(v_{eb}^e\times)(\omega_{ie}^e\times)\phi^e\\
	=&\tilde{g}_{ib}^e\times \phi^e\textcolor{red}{-2({\omega}_{ie}^e\times)J\rho_v^e}- {v}_{eb}^e\times({C}_b^e(\delta b_g^b+w_g^b))-{C}_b^e(\delta b_a^b+w_a^b)-\tilde{g}_{ib}^e+g_{ib}^e\\
	&-(\tilde{v}_{eb}^e\times)(\omega_{ie}^e\times)\phi^e\\
	\end{aligned}
	\end{equation}
	where the second order small quantity $(J\rho_v^e\times)({C}_b^e\delta \omega_{ib}^b)$ is neglected; and as $g_{ib}^e$ can be approximated as constant, $\tilde{g}_{ib}^e-g_{ib}^e$ can also be neglected.
	
	In the same way,the differential equation of the position error state is given as
	\begin{equation}\label{new_position_error_right_ECEF_r}
	\begin{aligned}
	&\frac{d}{dt}(J\rho_{r}^e)=\frac{d}{dt}({r}_{eb}^e-{C}_b^e\tilde{C}_e^b\tilde{r}_{eb}^e)=\dot{{r}}_{eb}^e -{C}_b^e\tilde{C}_e^b\dot{\tilde{r}}_{eb}^e - \frac{d}{dt}({C}_b^e\tilde{C}_e^b)\tilde{r}_{eb}^e\\
	=&{v}_{eb}^e -{C}_b^e\tilde{C}_e^b\tilde{v}_{eb}^e
	-\left(-{C}_b^e(\delta \omega_{ib}^b\times){C}_e^b{C}_b^e\tilde{C}_e^b-({\omega}_{ie}^e\times){C}_b^e\tilde{C}_e^b+{C}_b^e\tilde{C}_e^b(\omega_{ie}^e\times)\right)\tilde{r}_{eb}^e\\
	\approx&-(\tilde{r}_{eb}^e\times)(\omega_{ie}^e\times)\phi^e+({v}_{eb}^e-{C}_b^e\tilde{C}_e^b\tilde{v}_{eb}^e)+(({C}_b^e\delta \omega_{ib}^b)\times)\textcolor{red}{{C}_b^e\tilde{C}_e^b\tilde{r}_{eb}^e}\\
	\approx &-(\tilde{r}_{eb}^e\times)(\omega_{ie}^e\times)\phi^e+J\rho_v^e+(({C}_b^e\delta \omega_{ib}^b)\times) \textcolor{red}{({r}_{ib}^e-J\rho_r^e)}\\
	\approx&-(\tilde{r}_{eb}^e\times)(\omega_{ie}^e\times)\phi^e+J\rho_v^e-{r}_{eb}^e\times({C}_b^e\delta \omega_{eb}^b)\\
	=&-(\tilde{r}_{eb}^e\times)(\omega_{ie}^e\times)\phi^e+J\rho_v^e-{r}_{eb}^e\times({C}_b^e(\delta b_g^b+w_g^b))
	\end{aligned}
	\end{equation}
	where the second order small quantity $(J\rho_r^e\times)(C_b^e\delta \omega_{ib}^b)$ is neglected.
	
	The difference of the error state differential equations between the $SE_2(3)$ based EKF with ECEF frame attitude and the $SE_2(3)$ based EKF with estimated ECEF frame attitude lies in the $\delta f_{ib}^b$ term and the $\delta \omega_{ib}^b$ term.
	Thus, the error state $\delta x$, the error state transition matrix $F$, and the noise driven matrix $G$ of the inertial-integrated error state dynamic equation for $SE_2(3)$ based EKF with estimated body frame attitude are represented as
	\begin{equation}\label{state_x_right_ECEF_r}
	\begin{aligned}
	&x=\begin{bmatrix}
	\phi^e\\ J\rho_{v}^e \\J\rho_{r}^e \\\delta b_g^b \\\delta b_a^b
	\end{bmatrix}, 
	F=\begin{bmatrix}
	-\omega_{ie}^e\times & 0 & 0& -{C}_b^e &0\\
	-(\tilde{v}_{eb}^e\times)(\omega_{ie}^e\times)+\tilde{g}_{ib}^e\times &-2\omega_{ie}^e\times&0 & -{v}_{eb}^e\times {C}_b^e& -{C}_b^e\\
	-(\tilde{r}_{eb}^e\times)(\omega_{ie}^e\times)&I&0&-{r}_{eb}^e\times {C}_b^e&0\\
	0&0&0&-\frac{1}{\tau_g}&0\\
	0&0&0&0&-\frac{1}{\tau_a}
	\end{bmatrix},\\
	&G=\begin{bmatrix}
	-{C}_b^e&0&0&0\\ -{v}_{eb}^e\times {C}_b^e &-{C}_b^e&0&0\\ -{r}_{eb}^e\times {C}_b^e&0&0&0\\0&0&I&0\\0&0&0&I
	\end{bmatrix}
	\end{aligned}
	\end{equation}
		\begin{remark}
		In fact, the perturbation term $\delta\omega_{in}^n$ can be neglected for many applications in the $SE_2(3)$ based EKF algorithms that have been derived above, including the algorithms for NED navigation with consumer-grade IMU and initial alignment such as motion alignment and quasi-static alignment with the GNSS velocity and position measurements. This trick will greatly simplifies the error state dynamic equations for all the $SE_2(3) $ based EKF framework.
	\end{remark}
	\begin{remark}
		In~\cite{yan2008on}, a dampling SINS differential equations is proposed to reduce the complexity of the corresponding error state equations by neglecting the $\delta \omega_{in}^n$ as $\omega_{in}^n$ can be calculated by the velocity and position values provided by GNSS in the initial alignment problem. This idea can be applied to every error dynamic model in this paper when confronting with initial alignment problems, such as motion alignment and quasi-static alignment. Furthermore, more tricks can be applied to the dynamic models according to different applications.
	\end{remark}
\section{$SE_2(3)$ based EKF for another ECEF Navigation}

	When the system state is defined as
	\begin{equation}\label{new_state_ie_n}
	\mathcal{X}=\begin{bmatrix}
	C_b^e & v_{ib}^e & r_{ib}^e\\
	0_{1\times3} & 1 & 0\\
	0_{1\times 3} & 0& 1
	\end{bmatrix}\in SE_2(3)
	\end{equation}
	where $C_b^e$ is the direction cosine matrix from the body frame to the ECEF frame; $v_{ib}^e$ is the velocity of body frame relative to the ECI frame expressed in the ECEF frame; $r_{ib}^e$ is the position of body frame relative to the ECI frame expressed in the ECEF frame.
	
	When the state is defined as in equation(\ref{new_state_ie_n}), the invariant property of the dynamic can also be proofed in a similar approach. However, they just used the right invariant error state prediction for the inertial-integrated navigation which is not consistent with the right invariant-EKF from the perspective of invariance as the GNSS measurements are left-invariant. 
	
	Therefore, we give the left-invariant EKF in detail for the first time, which is also the framework of the $SE_2(3)$ based EKF.
	Firstly, the state and its inverse defined on the matrix Lie group are given as
	\begin{equation}\label{matrix_lie_group}
	\mathcal{X}=\begin{bmatrix}
	C_b^e & v_{ib}^e & r_{ib}^e\\0_{1\times 3}&1&0\\0_{1\times 3}&0&1
	\end{bmatrix},\mathcal{X}^{-1}=\begin{bmatrix}
	C_e^b & -v_{ib}^b & -r_{ib}^b\\0_{1\times 3}&1&0\\0_{1\times 3}&0&1
	\end{bmatrix}
	\end{equation}
	
	Then the dynamic equation of the state $\mathcal{X}$  can be deduced as follows
	\begin{equation}\label{lie_group_state_differential}
	\begin{aligned}
	&\frac{d}{dt}\mathcal{X}=f_{u_t}(\mathcal{X})=\frac{d}{dt}\begin{bmatrix}
	C_b^e & v_{ib}^e & r_{ib}^e\\0_{1\times 3}&1&0\\0_{1\times 3}&0&1
	\end{bmatrix}=\begin{bmatrix}
	\dot{C}_b^e & \dot{v}_{ib}^e & \dot{r}_{ib}^e\\0_{1\times 3}&0&0\\0_{1\times 3}&0&0
	\end{bmatrix}\\
	=&\begin{bmatrix}
	C_b^e(\omega_{ib}^b\times)-(\omega_{ie}^e\times)C_b^e & (-\omega_{ie}^e\times)v_{ib}^e+C_b^ef^b+G_{ib}^e & (-\omega_{ie}^e\times)r_{ib}^e+v_{ib}^e\\
	0_{1\times 3}&0&0\\0_{1\times 3}&0&0
	\end{bmatrix}\\
	=&\begin{bmatrix}
	C_b^e & v_{ib}^e & r_{ib}^e\\0_{1\times 3}&1&0\\0_{1\times 3}&0&1
	\end{bmatrix}\begin{bmatrix}
	\omega_{ib}^b\times & f^b & 0_{3\times 1}\\ 0_{1\times 3} &0&0\\ 0_{1\times 3}&0&0
	\end{bmatrix}+\begin{bmatrix}
	-\omega_{ie}^e\times & G_{ib}^e & v_{ib}^e\\ 0_{1\times 3} &0&0\\ 0_{1\times 3}&0&0
	\end{bmatrix}\begin{bmatrix}
	C_b^e & v_{ib}^e & r_{ib}^e\\0_{1\times 3}&1&0\\0_{1\times 3}&0&1
	\end{bmatrix}\\
	=&\mathcal{X}W_1+W_2\mathcal{X}
	\end{aligned}
	\end{equation}
	
	It is easy to verify that the dynamical equation(\ref{lie_group_state_differential}) satisfies the group-affine property in the same way as equation(\ref{proof_invariance}).
	As the error can be defined as the multiplication of the element and its inverse on matrix manifold. The error state can be defined in one of four ways:
	$\eta=\tilde{\mathcal{X}}\mathcal{X}^{-1}$, $\eta={\mathcal{X}}\tilde{\mathcal{X}}^{-1}$, $\eta=\tilde{\mathcal{X}}^{-1}\mathcal{X}$, and $\eta={\mathcal{X}}^{-1}\tilde{\mathcal{X}}$. The first two error states are left invariant, the last two error states are right invariant. While the first and fourth error state definitions are similar to the error definition in Euclidean space, that is the estimated value minus the true value, and the second and third error state definitions are similar to the error definition in Euclidean space, that is the true value minus the estimated value.
\subsection{Left $SE_2(3)$ based EKF for ECEF Navigation with Body Frame Attitude Error}
Considering the measurements of the GNSS are left-invariant observations for the world-centric observer, we first give the left-invariant error state differential equations.
The left-invariant error is defined as
\begin{equation}\label{left_invariant_error}
\eta^L=\mathcal{X}^{-1}\tilde{X}=\begin{bmatrix}
C_e^b & -v_{ib}^b & -r_{ib}^b
\\ 0_{1\times 3}&1&0\\0_{1\times 3}&0&1
\end{bmatrix}\begin{bmatrix}
\tilde{C}_b^e & \tilde{v}_{ib}^e & \tilde{r}_{ib}^e\\
0_{1\times 3}&1&0\\0_{1\times 3}&0&1
\end{bmatrix}=\begin{bmatrix}
C_e^b\tilde{C}_b^e & C_e^b\tilde{v}_{ib}^e -v_{ib}^b& C_e^b\tilde{r}_{ib}^e-r_{ib}^b\\
0_{1\times 3}&1&0\\0_{1\times 3}&0&1
\end{bmatrix}
\end{equation}

According to the map form the Lie algebra to the Lie group, the error states of attitude, velocity, and position can de derived as
\begin{equation}\label{attitude_velocity_position}
\begin{aligned}
C_e^b\tilde{C}_b^e =&\exp_G(\phi^b\times)\approx I+\phi^b\times\\
\eta_v^L=J\rho_{v}^b=&C_e^b\tilde{v}_{ib}^e -v_{ib}^b=C_e^b\tilde{v}_{ib}^e-C_e^bv_{ib}^e=C_e^b(\tilde{v}_{ib}^e -v_{ib}^e)=C_e^b\delta v_{ib}^e \\ 
\eta_r^L=J\rho_{r}^b=&C_e^b\tilde{r}_{ib}^e-r_{ib}^b =C_e^b\tilde{r}_{ib}^e-C_e^br_{ib}^e=C_e^b(\tilde{r}_{ib}^e -r_{ib}^e)=C_e^b\delta r_{ib}^e
\end{aligned}
\end{equation}

Meanwhile, the left-invariant error satisfies that
\begin{equation}\label{lie_algebra_left_error}
\eta^L=\begin{bmatrix}
\exp_G(\phi^b\times) & J\rho_{v}^b & J\rho_{r}^b\\
0_{1\times 3} & 1& 0\\ 0_{1\times 3}&0&1
\end{bmatrix}=\exp_G\left(\begin{bmatrix}
(\phi^b)\times & \rho_{v}^b & \rho_{r}^b\\0_{1\times 3}&0&0\\0_{1\times 3}&0&0
\end{bmatrix} \right)=\exp_G\left(\Lambda\begin{bmatrix}
\phi^b \\ \rho_{v}^b \\ \rho_{r}^b
\end{bmatrix} \right)
\end{equation}
where $\phi^b$ is the attitude error state, $J\rho_{v}^b$ is the new definition of velocity error state, $J\rho_{r}^b$ is the new definition of position error state;
$J$ is the left Jacobian matrix given in equation(\ref{left_Jacobian_n}).

The differential equation of the attitude error state is given as
\begin{equation}\label{attitude_error}
\begin{aligned}
\frac{d}{dt}(C_e^b\tilde{C}_b^e)&=\dot{C}_e^b\tilde{C}_b^e+C_e^b\dot{\tilde{C}}_b^e\\
&=\left[C_e^b(\omega_{ie}^e\times)-(\omega_{ib}^b\times)C_e^b\right]\tilde{C}_b^e+C_e^b\left[\tilde{C}_b^e(\tilde{\omega}_{ib}^b\times)-(\tilde{\omega}_{ie}^e\times)\tilde{C}_b^e\right]\\
&=C_e^b(\omega_{ie}^e\times)\tilde{C}_b^e-(\omega_{ib}^b\times)C_e^b\tilde{C}_b^e+C_e^b\tilde{C}_b^e(\tilde{\omega}_{ib}^b\times)-C_e^b({\omega}_{ie}^e\times)\tilde{C}_b^e\\
&\approx -(\omega_{ib}^b\times)(I+\phi^b\times)+(I+\phi^b\times)(({\omega}_{ib}^b+\delta \omega_{ib}^b)\times)\\
&=-(\omega_{ib}^b\times)(\phi^b\times)+(\delta \omega_{ib}^b)\times+ (\phi^b\times)(\omega_{ib}^b\times)+\phi^b\times(\delta \omega_{ib}^b)\times \\
&\approx (\phi^b\times\omega_{ib}^b)\times+\delta \omega_{ib}^b\times= (\phi^b\times\omega_{ib}^b)\times+(\delta b_g^b+w_g^b)\times
\end{aligned}
\end{equation}
where the angular velocity error of the earth's rotation can be neglected, i.e., $\tilde{\omega}_{ie}^e=\omega_{ie}^e$; and second order small quantity $(\phi^b\times)(\delta \omega_{ib}^b\times)$ is also neglected. Therefore, the equation(\ref{attitude_error}) can be simplified as
\begin{equation}\label{attitude_error_differential}
\dot{\phi}^b=\phi^b\times\omega_{ib}^b+\delta \omega_{ib}^b=-\omega_{ib}^b \times \phi^b+\delta \omega_{ib}^b=-\omega_{ib}^b\times\phi^b+\delta b_g^b+w_g^b
\end{equation}

The differential equation of the velocity error state is given as
\begin{equation}\label{new_velocity_error}
\begin{aligned}
&\frac{d}{dt}(J\rho_{v}^b)=\dot{C}_e^b\delta v_{ib}^e+C_e^b(\dot{\tilde{v}}_{ib}^e-\dot{v}_{ib}^e)\\
=&\left[C_e^b(\omega_{ie}^e\times)-(\omega_{ib}^b\times)C_e^b\right]\delta v_{ib}^e\\
&+C_e^b\left( \left[(-\tilde{\omega}_{ie}^e\times)\tilde{v}_{ib}^e+\tilde{C}_b^e\tilde{f}^b+\tilde{G}_{ib}^e\right] -\left[(-\omega_{ie}^e\times)v_{ib}^e+C_b^ef^b+G_{ib}^e\right]  \right)\\
=&C_e^b(\omega_{ie}^e\times)\delta v_{ib}^e-(\omega_{ib}^b\times)\textcolor{red}{C_e^b\delta v_{ib}^e}+C_e^b\tilde{C}_b^e\tilde{f}^b-C_e^bC_b^ef^b-C_e^b\omega_{ie}^e\times(\tilde{v}_{ib}^e-v_{ib}^e)\\
&+C_e^b(\tilde{G}_{ib}^e-G_{ib}^e)\\
\approx&-(\omega_{ib}^b\times)J\rho_{v}^b+(I+\phi^b\times)({f}^b+\delta b_a^b+w_a^b)-f^b+C_e^b(\tilde{G}_{ib}^e-G_{ib}^e)\\
=&-(\omega_{ib}^b\times)J\rho_{v}^b+\phi^b\times f^b+\phi^b\times\delta f^b +C_e^b(\tilde{G}_{ib}^e-G_{ib}^e)+\delta f^b \\
\approx & -(\omega_{ib}^b\times)J\rho_{v}^b-f^b\times \phi^b+C_e^b(\tilde{G}_{ib}^e-G_{ib}^e)+\delta f^b\\
= & -(\omega_{ib}^b\times)J\rho_{v}^b-f^b\times \phi^b+C_e^b(\tilde{G}_{ib}^e-G_{ib}^e)+\delta b_a^b+w_a^b
\end{aligned}
\end{equation}
where the second order small quantity $\phi^b\times\delta f^b $ is neglected; and as $G_{ib}^e$ can be approximated as constant, $C_e^b(\tilde{G}_{ib}^e-G_{ib}^e)$ can also be neglected.

In the same way, the differential equation of the position error state is given as
\begin{equation}\label{new_position_error}
\begin{aligned}
&\frac{d}{dt}(J\rho_{r}^b)=\dot{C}_e^b\delta r_{ib}^e+C_e^b(\dot{\tilde{r}}_{ib}^e-\dot{r}_{ib}^e)\\
=&\left[C_e^b(\omega_{ie}^e\times)-(\omega_{ib}^b\times)C_e^b\right]\delta r_{ib}^e
+C_e^b\left( \left[(-\tilde{\omega}_{ie}^e\times)\tilde{r}_{ib}^e+\tilde{v}_{ib}^e\right] -\left[(-\omega_{ie}^e\times)r_{ib}^e+v_{ib}^e\right]  \right)\\
=&C_e^b(\omega_{ie}^e\times)\delta r_{ib}^e-(\omega_{ib}^b\times)\textcolor{red}{C_e^b\delta r_{ib}^e}-C_e^b\omega_{ie}^e\times(\tilde{r}_{ib}^e-r_{ib}^e)+C_e^b(\tilde{v}_{ib}^e-v_{ib}^e)\\
=&-\omega_{ib}^b\times J\rho_{r}^b+J\rho_{v}^b
\end{aligned}
\end{equation}

Thus, the inertial-integrated error state dynamic equation for the $SE_2(3)$ based EKF can be obtained 
\begin{equation}\label{invariant_ekf}
\delta\dot{x}=F\delta x+Gw
\end{equation}
where $F$ is the error state transition matrix; $\delta x$ is the error state including the terms about bias; G is the noise driven matrix. Their definition is given as
\begin{equation}\label{state_x}
x=\begin{bmatrix}
\phi^b\\ J\rho_{v}^b \\J\rho_{r}^b \\\delta b_g^b \\\delta b_a^b
\end{bmatrix}, 
F=\begin{bmatrix}
-\omega_{ib}^b\times & 0 & 0& I &0\\
-f^b\times &-\omega_{ib}^b\times&0 & 0& I\\
0&I&-\omega_{ib}^b\times&0&0\\
0&0&0&-\frac{1}{\tau_g}&0\\
0&0&0&0&-\frac{1}{\tau_a}
\end{bmatrix},G=\begin{bmatrix}
I&0\\0&I\\0&0\\0&0\\0&0
\end{bmatrix},
w=\begin{bmatrix}w_g^b\\w_a^b \\ w_{b_g}^b \\ w_{b_a}^b\end{bmatrix}
\end{equation}

Comparing with the error state transition matrix in the navigation frame, the error state transition matrix in the ECEF frame owns more sparse matrix form which is beneficial to improve the calculation speed and stability of the inertial-integrated navigation system. 
\subsection{Left $SE_2(3)$ based EKF measurement equation}
If the lever arm error is taken into account, the measurement error vector is expressed in the ECEF frame as the difference between the position calculated by INS and the position calculated by GNSS:
\begin{equation}\label{measurement_arm_error_left}
\begin{aligned}
\delta z_r&=\tilde{r}_{SINS}^e-\tilde{r}_{GNSS}^e=\tilde{r}_{IMU}^e+\tilde{C}_b^el^b-r_{GNSS}^e+n_{GNSS}\\
&\approx r_{IMU}^e+\delta r_{IMU}^e+C_b^e(I+\phi^b\times)l^b-r_{GNSS}^e+n_{GNSS}\\
&=r_{IMU}^e+C_b^el^b-r_{GNSS}^e+\delta r_{IMU}^e+C_b^e(\phi^b\times)l^b+n_{GNSS}\\
&=\delta r_{IMU}^e-C_b^e(l^b\times) \phi^b+n_{GNSS}\\
&=\delta \tilde{r}_{ib}^e-C_b^e(l^b\times) \phi^b+n_{GNSS}\\
&\approx \tilde{C}_b^eJ\rho_{r}^b-C_b^e(l^b\times) \phi^b+n_{GNSS}
\approx \tilde{C}_b^eJ\rho_{r}^b-\tilde{C}_b^e(I-\phi^b\times)(l^b\times) \phi^b+n_{GNSS}\\
&\approx \tilde{C}_b^eJ\rho_{r}^b-\tilde{C}_b^e(l^b\times) \phi^b+n_{GNSS}\\
&\approx C_b^e(I+\phi^b\times)J\rho_{r}^b-C_b^e(l^b\times) \phi^b+n_{GNSS}\approx C_b^eJ\rho_{r}^b-C_b^e(l^b\times) \phi^b+n_{GNSS}
\end{aligned}
\end{equation}

Then the measurement matrix can be written as
\begin{equation}\label{measurement_matrix_left}
H=\begin{bmatrix}
-C_b^e(l^b\times) & 0& {C}_{b}^e &0&0
\end{bmatrix}
\end{equation} 
where $C_b^e$ can be replaced by $\tilde{C}_b^e$ when implement the algorithm, because the resulting error can be eliminated by a small second order quantity.

\begin{remark}
	It is worth noting that all the left Jacobian matrix $J$ can be approximated as $J\approx I_{3\times3}$ if $||\phi^b||$ is small enough.
\end{remark}

\subsection{Left $SE_2(3)$ based EKF for ECEF Navigation with Estimated Body Frame Attitude Error}
Considering the measurements of the GNSS are left-invariant observations for the world-centric observer, we first give the left-invariant error state differential equations.
The left-invariant error is defined as
\begin{equation}\label{left_invariant_error_estimated}
\eta^L=\tilde{\mathcal{X}}^{-1}\mathcal{X}=\begin{bmatrix}
\tilde{C}_e^b & -\tilde{v}_{ib}^b & -\tilde{r}_{ib}^b
\\ 0_{1\times 3}&1&0\\0_{1\times 3}&0&1
\end{bmatrix}\begin{bmatrix}
{C}_b^e & {v}_{ib}^e & {r}_{ib}^e\\
0_{1\times 3}&1&0\\0_{1\times 3}&0&1
\end{bmatrix}=\begin{bmatrix}
\tilde{C}_e^bC_b^e & \tilde{C}_e^bv_{ib}^e -\tilde{v}_{ib}^b& \tilde{C}_e^b{r}_{ib}^e-\tilde{r}_{ib}^b\\
0_{1\times 3}&1&0\\0_{1\times 3}&0&1
\end{bmatrix}
\end{equation}

According to the map form the Lie algebra to the Lie group, the error states of attitude, velocity, and position can de derived as
\begin{equation}\label{attitude_velocity_position_estimated}
\begin{aligned}
\tilde{C}_e^b{C}_b^e =&\exp_G(\phi^b\times)\approx I+\phi^b\times\\
\eta_v^L=J\rho_{v}^b=&\tilde{C}_e^b{v}_{ib}^e -\tilde{v}_{ib}^b=\tilde{C}_e^b{v}_{ib}^e-\tilde{C}_e^b\tilde{v}_{ib}^e=\tilde{C}_e^b({v}_{ib}^e -\tilde{v}_{ib}^e)=-\tilde{C}_e^b\delta v_{ib}^e \\ 
\eta_r^L=J\rho_{r}^b=&\tilde{C}_e^b{r}_{ib}^e-\tilde{r}_{ib}^b =\tilde{C}_e^b{r}_{ib}^e-\tilde{C}_e^b\tilde{r}_{ib}^e=\tilde{C}_e^b({r}_{ib}^e -\tilde{r}_{ib}^e)=-\tilde{C}_e^b\delta r_{ib}^e
\end{aligned}
\end{equation}

Meanwhile, the left-invariant error satisfies that
\begin{equation}\label{lie_algebra_left_error_estimated}
\eta^L=\begin{bmatrix}
\exp_G(\phi^b\times) & J\rho_{v}^b & J\rho_{r}^b\\
0_{1\times 3} & 1& 0\\ 0_{1\times 3}&0&1
\end{bmatrix}=\exp_G\left(\begin{bmatrix}
(\phi^b)\times & \rho_{v}^b & \rho_{r}^b\\0_{1\times 3}&0&0\\0_{1\times 3}&0&0
\end{bmatrix} \right)=\exp_G\left(\Lambda\begin{bmatrix}
\phi^b \\ \rho_{v}^b \\ \rho_{r}^b
\end{bmatrix} \right)
\end{equation}
where $\phi^b$ is the attitude error state, $J\rho_{v}^b$ is the new definition of velocity error state, $J\rho_{r}^b$ is the new definition of position error state;
$J$ is the left Jacobian matrix given in equation(\ref{left_Jacobian_n}).

The differential equation of the attitude error state is given as
\begin{equation}\label{attitude_error_estimated}
\begin{aligned}
\frac{d}{dt}(\tilde{C}_e^b{C}_b^e)&=\dot{\tilde{C}}_e^b{C}_b^e+\tilde{C}_e^b\dot{{C}}_b^e\\
&=\left[\tilde{C}_e^b(\tilde{\omega}_{ie}^e\times)-(\tilde{\omega}_{ib}^b\times)\tilde{C}_e^b\right]{C}_b^e+\tilde{C}_e^b\left[{C}_b^e({\omega}_{ib}^b\times)-({\omega}_{ie}^e\times){C}_b^e\right]\\
&=\tilde{C}_e^b(\omega_{ie}^e\times){C}_b^e-(\tilde{\omega}_{ib}^b\times)\tilde{C}_e^b{C}_b^e+\tilde{C}_e^b{C}_b^e({\omega}_{ib}^b\times)-\tilde{C}_e^b({\omega}_{ie}^e\times){C}_b^e\\
&\approx -(\tilde{\omega}_{ib}^b\times)(I+\phi^b\times)+(I+\phi^b\times)((\tilde{\omega}_{ib}^b-\delta \omega_{ib}^b)\times)\\
&=-(\tilde{\omega}_{ib}^b\times)(\phi^b\times)-(\delta \omega_{ib}^b)\times+ (\phi^b\times)(\tilde{\omega}_{ib}^b\times)-\phi^b\times(\delta \omega_{ib}^b)\times \\
&\approx (\phi^b\times \tilde{\omega}_{ib}^b)\times-\delta \omega_{ib}^b\times= (\phi^b\times \tilde{\omega}_{ib}^b)\times-(\delta b_g^b+w_g^b)\times
\end{aligned}
\end{equation}
where the angular velocity error of the earth's rotation can be neglected, i.e., $\tilde{\omega}_{ie}^e=\omega_{ie}^e$; and second order small quantity $(\phi^b\times)(\delta \omega_{ib}^b\times)$ is also neglected. Therefore, the equation(\ref{attitude_error}) can be simplified as
\begin{equation}\label{attitude_error_differential_estimated}
\dot{\phi}^b=\phi^b\times \tilde{\omega}_{ib}^b-\delta \omega_{ib}^b=-\tilde{\omega}_{ib}^b \times \phi^b-\delta \omega_{ib}^b=-\tilde{\omega}_{ib}^b\times\phi^b-\delta b_g^b-w_g^b
\end{equation}

The differential equation of the velocity error state is given as
\begin{equation}\label{new_velocity_error_estimated}
\begin{aligned}
&\frac{d}{dt}(J\rho_{v}^b)=-\dot{\tilde{C}}_e^b\delta v_{ib}^e+\tilde{C}_e^b(\dot{{v}}_{ib}^e-\dot{\tilde{v}}_{ib}^e)\\
=&-\left[\tilde{C}_e^b(\tilde{\omega}_{ie}^e\times)-(\tilde{\omega}_{ib}^b\times)\tilde{C}_e^b\right]\delta v_{ib}^e\\
&+\tilde{C}_e^b\left(\left[(-\omega_{ie}^e\times)v_{ib}^e+C_b^ef^b+G_{ib}^e\right]- \left[(-\tilde{\omega}_{ie}^e\times)\tilde{v}_{ib}^e+\tilde{C}_b^e\tilde{f}^b+\tilde{G}_{ib}^e\right]\right)\\
=&-\tilde{C}_e^b(\omega_{ie}^e\times)\delta v_{ib}^e+(\tilde{\omega}_{ib}^b\times)\textcolor{red}{\tilde{C}_e^b\delta v_{ib}^e}+\tilde{C}_e^b{C}_b^e{f}^b-\tilde{f}^b-\tilde{C}_e^b\omega_{ie}^e\times({v}_{ib}^e-\tilde{v}_{ib}^e)\\
&+\tilde{C}_e^b({G}_{ib}^e-\tilde{G}_{ib}^e)\\
\approx&-(\tilde{\omega}_{ib}^b\times)J\rho_{v}^b+(I+\phi^b\times)(\tilde{f}^b-\delta b_a^b-w_a^b)-\tilde{f}^b+\tilde{C}_e^b({G}_{ib}^e-\tilde{G}_{ib}^e)\\
=&-(\tilde{\omega}_{ib}^b\times)J\rho_{v}^b+\phi^b\times \tilde{f}^b-\phi^b\times\delta f^b +\tilde{C}_e^b({G}_{ib}^e-\tilde{G}_{ib}^e)-\delta f^b \\
\approx & -(\omega_{ib}^b\times)J\rho_{v}^b-f^b\times \phi^b+\tilde{C}_e^b({G}_{ib}^e-\tilde{G}_{ib}^e)+\delta f^b\\
= & -(\omega_{ib}^b\times)J\rho_{v}^b-f^b\times \phi^b+\tilde{C}_e^b({G}_{ib}^e-\tilde{G}_{ib}^e)-\delta b_a^b-w_a^b
\end{aligned}
\end{equation}
where the second order small quantity $\phi^b\times\delta f^b $ is neglected; and as $G_{ib}^e$ can be approximated as constant, $\tilde{C}_e^b({G}_{ib}^e-\tilde{G}_{ib}^e)$ can also be neglected.

In the same way, the differential equation of the position error state is given as
\begin{equation}\label{new_position_error_estimated}
\begin{aligned}
&\frac{d}{dt}(J\rho_{r}^b)=-\dot{\tilde{C}}_e^b\delta r_{ib}^e+\tilde{C}_e^b(\dot{{r}}_{ib}^e-\dot{\tilde{r}}_{ib}^e)\\
=&-\left[\tilde{C}_e^b(\omega_{ie}^e\times)-(\tilde{\omega}_{ib}^b\times)\tilde{C}_e^b\right]\delta r_{ib}^e
+\tilde{C}_e^b\left(\left[(-\omega_{ie}^e\times)r_{ib}^e+v_{ib}^e\right]- \left[(-\tilde{\omega}_{ie}^e\times)\tilde{r}_{ib}^e+\tilde{v}_{ib}^e\right]\right)\\
=&-\tilde{C}_e^b(\omega_{ie}^e\times)\delta r_{ib}^e+(\tilde{\omega}_{ib}^b\times)\textcolor{red}{\tilde{C}_e^b\delta r_{ib}^e}-\tilde{C}_e^b\omega_{ie}^e\times({r}_{ib}^e-\tilde{r}_{ib}^e)+\tilde{C}_e^b({v}_{ib}^e-\tilde{v}_{ib}^e)\\
=&-\tilde{\omega}_{ib}^b\times J\rho_{r}^b+J\rho_{v}^b
\end{aligned}
\end{equation}

Thus, the inertial-integrated error state dynamic equation for the $SE_2(3)$ based EKF can be obtained 
\begin{equation}\label{invariant_ekf_estimated}
\delta\dot{x}=F\delta x+Gw
\end{equation}
where $F$ is the error state transition matrix; $\delta x$ is the error state including the terms about bias; G is the noise driven matrix. Their definition is given as
\begin{equation}\label{state_x_estimated}
\begin{aligned}
&\delta x=\begin{bmatrix}
\phi^b\\ J\rho_{v}^b \\J\rho_{r}^b \\\delta b_g^b \\\delta b_a^b
\end{bmatrix}, 
F=\begin{bmatrix}
-\tilde{\omega}_{ib}^b\times & 0 & 0& -I_{3\times 3} &0\\
-\tilde{f}^b\times &-\tilde{\omega}_{ib}^b\times&0 & 0& -I_{3\times 3}\\
0&I_{3\times 3}&-\tilde{\omega}_{ib}^b\times&0&0\\
0&0&0&-\frac{1}{\tau_g}&0\\
0&0&0&0&-\frac{1}{\tau_a}
\end{bmatrix},\\
&
G=\begin{bmatrix}
-I_{3\times 3}&0&0&0\\0&-I_{3\times 3}&0&0\\0&0&0&0\\0&0&I_{3\times 3}&0\\0&0&0&I_{3\times 3}
\end{bmatrix},
w=\begin{bmatrix}w_g^b\\w_a^b \\ w_{b_g}^b \\ w_{b_a}^b\end{bmatrix}
\end{aligned}
\end{equation}
\subsection{Left $SE_2(3)$ based EKF measurement equation}
If the lever arm error is taken into account, the measurement error vector is expressed in the ECEF frame as the difference between the position calculated by GNSS and the position calculated by SINS:
\begin{equation}\label{measurement_arm_error_left_true_minus_estimated}
\begin{aligned}
\delta z_r&=\tilde{r}_{GNSS}^e-\tilde{r}_{SINS}^e=r_{GNSS}^e+n_{GNSS}-(\tilde{r}_{IMU}^e+\tilde{C}_b^el^b)\\
&\approx -r_{IMU}^e-\delta r_{IMU}^e-C_b^e(I-\phi^b\times)l^b+r_{GNSS}^e+n_{GNSS}\\
&=-r_{IMU}^e-C_b^el^b+r_{GNSS}^e-\delta r_{IMU}^e+C_b^e(\phi^b\times)l^b+n_{GNSS}\\
&=-\delta r_{IMU}^e-C_b^e(l^b\times) \phi^b+n_{GNSS}\\
&=-\delta \tilde{r}_{ib}^e-C_b^e(l^b\times) \phi^b+n_{GNSS}\\
&\approx \tilde{C}_b^eJ\rho_{r}^b-C_b^e(l^b\times) \phi^b+n_{GNSS}
\approx \tilde{C}_b^eJ\rho_{r}^b-\tilde{C}_b^e(I-\phi^b\times)(l^b\times) \phi^b+n_{GNSS}\\
&\approx \tilde{C}_b^eJ\rho_{r}^b-\tilde{C}_b^e(l^b\times) \phi^b+n_{GNSS}\\
&\approx C_b^e(I+\phi^b\times)J\rho_{r}^b-C_b^e(l^b\times) \phi^b+n_{GNSS}\approx C_b^eJ\rho_{r}^b-C_b^e(l^b\times) \phi^b+n_{GNSS}
\end{aligned}
\end{equation}

Then the measurement matrix can be written as
\begin{equation}\label{measurement_matrix_left_estimated}
H_{new}=\begin{bmatrix}
-\tilde{C}_b^e(l^b\times) & 0& \tilde{C}_{b}^e &0&0
\end{bmatrix}
\end{equation} 
where $\tilde{C}_b^e$ can be replaced by ${C}_b^e$ when implement the algorithm, because the resulting error can be eliminated by a small second order quantity.
\begin{remark}
	The equivalence of the $SE_2(3)$ based measurement equation and the left invariant EKF measurement euation can be proofed similar to Subsection \ref{subsection_equivalence}. From the above two equivalent relationships we can draw the conclusion that the error definition $\eta=\tilde{\mathcal{X}}^{-1}\mathcal{X}$ is more reasonable for the global navigation such as GNSS and 5G whose measurements have the left-invariant observation form.
\end{remark}
\subsection{Right $SE_2(3)$ based EKF with ECEF frame attitude error}
As the $SE_2(3)$ based EKF with estimated ECEF frame attitude error has been derived in Wang's dissertation~\cite{wang2018inertial}, we only give the $SE_2(3)$ based EKF with ECEF frame attitude error here.
If the error state is converted to the true ECEF frame, i.e., $\eta=(\tilde{\mathcal{X}}R)({\mathcal{X}}R)^{-1}=\tilde{\mathcal{X}}{\mathcal{X}}^{-1}\in SE_2(3)$, then the right invariant error is defined as
\begin{equation}\label{key}
\begin{aligned}
\eta^R&=\tilde{\mathcal{X}}\mathcal{X}^{-1}=
\begin{bmatrix}
\tilde{C}_b^e & \tilde{v}_{ib}^e & \tilde{r}_{ib}^e\\0_{1\times 3}&1&0\\0_{1\times 3}&0&1
\end{bmatrix}
\begin{bmatrix}
C_e^b & -v_{ib}^b & -r_{ib}^b\\0_{1\times 3}&1&0\\0_{1\times 3}&0&1
\end{bmatrix}\\
&=\begin{bmatrix}
\tilde{C}_b^eC_e^b & \tilde{v}_{ib}^e -\tilde{C}_b^ev_{ib}^b& \tilde{r}_{ib}^e-\tilde{C}_b^er_{ib}^b\\
0_{1\times 3}&1&0\\0_{1\times 3}&0&1
\end{bmatrix}
\end{aligned}
\end{equation}

Similarity, the new error state defined on the matrix Lie group $SE_2(3)$ can be denoted as
\begin{equation}\label{attitude_velocity_position_right}
\begin{aligned}
\tilde{C}_b^eC_e^b =&\exp_G(\phi^e\times)\approx I+\phi^e\times\\
\eta_v^R=J\rho_{v}^e=&\tilde{v}_{ib}^e -\tilde{C}_b^ev_{ib}^b=\tilde{v}_{ib}^e- v_{ib}^e+v_{ib}^e-\tilde{C}_b^eC_e^bv_{ib}^e=\delta v_{ib}^e+(I-\exp_G(\phi^e\times)) v_{ib}^e\\ 
\eta_r^R=J\rho_{r}^e=&\tilde{r}_{ib}^e-\tilde{C}_b^er_{ib}^b=\tilde{r}_{ib}^e- r_{ib}^e+r_{ib}^e-\tilde{C}_b^eC_e^br_{ib}^e=\delta r_{ib}^e+(I-\exp_G(\phi^e\times)) r_{ib}^e
\end{aligned}
\end{equation}

Meanwhile, the right invariant error satisfies that
\begin{equation}\label{lie_algebra_right_error}
\eta^R=\begin{bmatrix}
\exp_G(\phi^e\times) & J\rho_{v}^e & J\rho_{r}^e\\
0_{1\times 3} & 1& 0\\ 0_{1\times 3}&0&1
\end{bmatrix}=\exp_G\left( \begin{bmatrix}
(\phi^e\times) & \rho_{v}^e & \rho_{r}^e\\
0_{1\times 3} & 0& 0\\ 0_{1\times 3}&0&0
\end{bmatrix}\right)=\exp_G\left( \Lambda\begin{bmatrix}
\phi^e \\ \rho_{v}^e \\ \rho_{r}^e
\end{bmatrix} \right)
\end{equation}
where $\phi^e$ is the attitude error state; $J\rho_v^e$ is the new definition of velocity error state; $J\rho_r^e$ is the new definition of position error state.

The differential equation of the attitude error state is given as
\begin{equation}\label{attitude_error_right}
\begin{aligned}
&\frac{d}{dt}(\tilde{C}_b^eC_e^b)=\dot{\tilde{C}}_b^eC_e^b+\tilde{C}_b^e\dot{C}_e^b\\
=&\left[\tilde{C}_b^e(\tilde{\omega}_{ib}^b\times)-(\tilde{\omega}_{ie}^e\times)\tilde{C}_b^e\right]C_e^b+\tilde{C}_b^e\left[C_e^b(\omega_{ie}^e\times)-(\omega_{ib}^b\times)C_e^b\right]\\
=&\tilde{C}_b^e(\tilde{{\omega}}_{ib}^b\times)C_e^b-({\omega}_{ie}^e\times)\tilde{C}_b^eC_e^b+\tilde{C}_b^eC_e^b(\omega_{ie}^e\times)-\tilde{C}_b^e(\omega_{ib}^b\times)C_e^b\\
\approx &\tilde{C}_b^e(\delta{\omega}_{ib}^b\times)C_e^b-({\omega}_{ie}^e\times)(I+\phi^e\times)+(I+\phi^e\times)(\omega_{ie}^e\times)\\
=&\tilde{C}_b^e(\delta \omega_{ib}^b\times)\textcolor{red}{\tilde{C}_e^b\tilde{C}_b^eC_e^b}-({\omega}_{ie}^e\times)(\phi^e\times)+(\phi^e\times)(\omega_{ie}^e\times)\\
\approx& ((\tilde{C}_b^e\delta \omega_{ib}^b)\times)(I+\phi^e\times)+((\phi^e\times\omega_{ie}^e)\times)\\
\approx&(\tilde{C}_b^e\delta \omega_{ib}^b)\times+(\phi^e\times\omega_{ie}^e)\times=(\tilde{C}_b^e(\delta b_g^b+w_g^b))\times+(\phi^e\times\omega_{ie}^e)\times
\end{aligned}
\end{equation}
where the second order small quantity $\left((\tilde{C}_b^e\delta \omega_{ib}^b)\times\right)(\phi^e\times)$ is neglected.
Therefore, the equation(\ref{attitude_error_right}) can be simplified as
\begin{equation}\label{attitude_error_differential_right}
\dot{\phi}^e=\phi^e\times\omega_{ie}^e+\tilde{C}_b^e\delta b_g^b+\tilde{C}_b^ew_g^b=\textcolor{red}{-\omega_{ie}^e\times\phi^e}+\tilde{C}_b^e\delta b_g^b+\tilde{C}_b^ew_g^b
\end{equation}

The differential equation of the velocity error state is given as
\begin{equation}\label{new_velocity_error_right}
\begin{aligned}
&\frac{d}{dt}(J\rho_{v}^e)=\frac{d}{dt}(\tilde{v}_{ib}^e-\tilde{C}_b^eC_e^bv_{ib}^e)=\dot{\tilde{v}}_{ib}^e -\tilde{C}_b^eC_e^b\dot{v}_{ib}^e-\frac{d}{dt}(\tilde{C}_b^eC_e^b)v_{ib}^e\\
=&\left[(-\tilde{\omega}_{ie}^e\times)\tilde{v}_{ib}^e+\tilde{C}_b^e\tilde{f}^b+\tilde{G}_{ib}^e\right] -\tilde{C}_b^eC_e^b\left[(-\omega_{ie}^e\times)v_{ib}^e+C_b^ef^b+G_{ib}^e\right]\\
&-\left(\tilde{C}_b^e(\delta \omega_{ib}^b\times)\tilde{C}_e^b\tilde{C}_b^eC_e^b-({\omega}_{ie}^e\times)\tilde{C}_b^eC_e^b+\tilde{C}_b^eC_e^b(\omega_{ie}^e\times)\right)v_{ib}^e\\
=&\textcolor{red}{\tilde{C}_b^e\delta f^b}-({\omega}_{ie}^e\times)(\tilde{v}_{ib}^e-\tilde{C}_b^eC_e^bv_{ib}^e)
-\left((\tilde{C}_b^e\delta \omega_{ib}^b)\times\right)\textcolor{red}{\tilde{C}_b^eC_e^b v_{ib}^e}+\tilde{G}_{ib}^e-\tilde{C}_b^eC_e^bG_{ib}^e\\
\approx&\tilde{C}_b^e\delta f^b-({\omega}_{ie}^e\times)J\rho_v^e
-(\tilde{C}_b^e\delta \omega_{ib}^b)\times \textcolor{red}{(\tilde{v}_{ib}^e-J\rho_v^e)}+\tilde{G}_{ib}^e-(I+\phi^e\times)G_{ib}^e\\
\approx&G_{ib}^e\times \phi^e\textcolor{red}{-({\omega}_{ie}^e\times)J\rho_v^e}+ \tilde{v}_{ib}^e\times(\tilde{C}_b^e\delta \omega_{ib}^b)+\tilde{C}_b^e\delta f^b+\tilde{G}_{ib}^e-G_{ib}^e\\
=&G_{ib}^e\times \phi^e\textcolor{red}{-({\omega}_{ie}^e\times)J\rho_v^e}+ \tilde{v}_{ib}^e\times(\tilde{C}_b^e(\delta b_g^b+w_g^b))+\tilde{C}_b^e(\delta b_a^b+w_a^b)+\tilde{G}_{ib}^e-G_{ib}^e
\end{aligned}
\end{equation}
where the second order small quantity $(J\rho_v^e\times)(\tilde{C}_b^e\delta \omega_{ib}^b)$ is neglected; and as $G_{ib}^e$ can be approximated as constant, $\tilde{G}_{ib}^e-G_{ib}^e$ can also be neglected.

In the same way,the differential equation of the position error state is given as
\begin{equation}\label{new_position_error_right}
\begin{aligned}
&\frac{d}{dt}(J\rho_{r}^e)=\frac{d}{dt}(\tilde{r}_{ib}^e-\tilde{C}_b^eC_e^br_{ib}^e)=\dot{\tilde{r}}_{ib}^e -\tilde{C}_b^eC_e^b\dot{r}_{ib}^e - \frac{d}{dt}(\tilde{C}_b^eC_e^b)r_{ib}^e\\
=&\left[(-\tilde{\omega}_{ie}^e\times)\tilde{r}_{ib}^e+\tilde{v}_{ib}^e\right] -\tilde{C}_b^eC_e^b\left[(-\omega_{ie}^e\times)r_{ib}^e+v_{ib}^e\right]\\
&-\left(\tilde{C}_b^e(\delta \omega_{ib}^b\times)\tilde{C}_e^b\tilde{C}_b^eC_e^b-({\omega}_{ie}^e\times)\tilde{C}_b^eC_e^b+\tilde{C}_b^eC_e^b(\omega_{ie}^e\times)\right)r_{ib}^e\\
\approx&(-\tilde{\omega}_{ie}^e\times)(\tilde{r}_{ib}^e-\tilde{C}_b^eC_e^br_{ib}^e)+(\tilde{v}_{ib}^e-\tilde{C}_b^eC_e^bv_{ib}^e)-((\tilde{C}_b^e\delta \omega_{ib}^b)\times)\textcolor{red}{\tilde{C}_b^eC_e^br_{ib}^e}\\
\approx &{(-\tilde{\omega}_{ie}^e\times)J\rho_r^e}+J\rho_v^e+((\tilde{C}_b^e\delta \omega_{ib}^b)\times) \textcolor{red}{(\tilde{r}_{ib}^e-J\rho_r^e)}\\
\approx&\textcolor{red}{(-\tilde{\omega}_{ie}^e\times)J\rho_r^e}+J\rho_v^e+\tilde{r}_{ib}^e\times(\tilde{C}_b^e\delta \omega_{ib}^b)\\
=&\textcolor{red}{(-{\omega}_{ie}^e\times)J\rho_r^e}+J\rho_v^e+\tilde{r}_{ib}^e\times(\tilde{C}_b^e(\delta b_g^b+w_g^b))
\end{aligned}
\end{equation}
where the second order small quantity $(J\rho_r^e\times)(C_b^e\delta \omega_{ib}^b)$ is neglected.

The difference of the error state differential equations between the $SE_2(3)$ based EKF with ECEF frame attitude and the $SE_2(3)$ based EKF with estimated ECEF frame attitude lies in the $\delta f_{ib}^b$ term and the $\delta \omega_{ib}^b$ term.
Thus, the error state $\delta x$, the error state transition matrix $F$, and the noise driven matrix $G$ of the inertial-integrated error state dynamic equation for $SE_2(3)$ based EKF with estimated body frame attitude are represented as
\begin{equation}\label{state_x_right}
x=\begin{bmatrix}
\phi^e\\ J\rho_{v}^e \\J\rho_{r}^e \\\delta b_g^b \\\delta b_a^b
\end{bmatrix}, F=\begin{bmatrix}
-\omega_{ie}^e\times & 0 & 0& \tilde{C}_b^e &0\\
G_{ib}^e\times &-\omega_{ie}^e\times&0 & \tilde{v}_{ib}^e\times \tilde{C}_b^e& \tilde{C}_b^e\\
0&I&-\omega_{ie}^e\times&\tilde{r}_{ib}^e\times \tilde{C}_b^e&0\\
0&0&0&0&0\\
0&0&0&0&0
\end{bmatrix},G=\begin{bmatrix}
\tilde{C}_b^e&0\\ \tilde{v}_{ib}^e\times \tilde{C}_b^e &\tilde{C}_b^e\\ \tilde{r}_{ib}^e\times \tilde{C}_b^e&0\\0&0\\0&0
\end{bmatrix}
\end{equation}
\subsection{Right $SE_2(3)$ based EKF measurement equation}
If the lever arm error is taken into account, the measurement error vector is expressed in the ECEF frame as the difference between the position calculated by GNSS and the position calculated by SINS:
\begin{equation}\label{measurement_arm_error_right}
\begin{aligned}
\delta z_r&=\tilde{r}_{SINS}^e-\tilde{r}_{GNSS}^e=\tilde{r}_{IMU}^e+\tilde{C}_b^el^b-r_{GNSS}^e+n_{GNSS}\\
&\approx r_{IMU}^e+\delta r_{IMU}^e+(I+\phi^e\times)C_b^el^b-r_{GNSS}^e+n_{GNSS}\\
&=r_{IMU}^e+C_b^el^b-r_{GNSS}^e+\delta r_{IMU}^e+\phi^e\times(C_b^el^b)+n_{GNSS}\\
&=\delta r_{IMU}^e-(C_b^el^b)\times \phi^e+n_{GNSS}
=\delta \tilde{r}_{ib}^e-(C_b^el^b)\times \phi^e+n_{GNSS}\\
&\approx J\rho_{r}^e-\tilde{r}_{ib}^e \times \phi^e-(C_b^el^b)\times \phi^e+n_{GNSS}\\
\end{aligned}
\end{equation}

Thus the measurement matrix can be written as
\begin{equation}\label{measurement_matrix_right}
H=\begin{bmatrix}
-(\tilde{r}_{ib}^e+C_b^el^b) \times & 0& I &0&0
\end{bmatrix}
\end{equation}
\section{$SE_2(3)$ based EKF for transformaed INS Mechanization in ECEF Frame}
Similar to the auxiliary velocity defined by equation(\ref{auxiliary_velocity}) in the navigation frame, for the inertial-integrated navigation in ECEF frame, a new auxiliary velocity can be defined as
\begin{equation}\label{auxiliary_velocity_ECEF}
\overline{v}_{eb}^e=v_{eb}^e+\omega_{ie}^e\times r_{eb}^e
\end{equation}

Then, the error state dynamical equation can be manipulated in parallel to the manipulation in section \ref{transformed_INS}, so the similar $SE_2(3)$ based filtering algorithms in ECEF frame are naturally obtained. The details will be given soon.

With the introduced auxiliary velocity vector, the INS mechanization in ECEF frame is given by
\begin{equation}\label{C_b_n_d_e_invariant2}
\dot{C}_b^e=C_b^e(\omega_{ib}^b\times)-(\omega_{ie}^e\times)C_b^e
\end{equation}
\begin{equation}\label{v_eb_n_d_e_invarante2}
\dot{v}_{eb}^e=C_b^ef_{ib}^b-(\omega_{ie}^e)\times \overline{v}_{eb}^e+G_{ib}^e
\end{equation}
\begin{equation}\label{r_eb_n_d_e_invariant2}
\dot{r}_{eb}^e=-\omega_{ie}^e\times r_{eb}^e+\overline{v}_{eb}^e
\end{equation}

Then defining the state composed by the attitude $C_b^e$, the velocity $\overline{v}_{eb}^e$, and the position $r_{eb}^e$ as the elements of the matrix Lie group $SE_2(3)$, that is
\begin{equation}\label{new_state2}
\mathcal{X}=\begin{bmatrix}
C_b^e & \overline{v}_{eb}^e & r_{eb}^e\\
0_{1\times 3} & 1 &0\\
0_{1\times 3} & 0 & 1
\end{bmatrix}
\end{equation}

Therefore, equation(\ref{C_b_n_d_e_invariant2}), equation(\ref{v_eb_n_d_e_invarante2}), equation(\ref{r_eb_n_d_e_invariant2}) can be rewritten in a compact form as
\begin{equation}\label{differential_invariant2}
\begin{aligned}
&\frac{d}{dt}\mathcal{X}=f_{u_t}(\mathcal{X})=\frac{d}{dt}\begin{bmatrix}
C_b^e & \overline{v}_{eb}^e & r_{eb}^e\\
0_{1\times3} & 1 & 0\\
0_{1\times 3} & 0& 1
\end{bmatrix}
=\begin{bmatrix}
\dot{C}_b^e & \dot{\overline{v}}_{eb}^e & \dot{r}_{eb}^e\\
0_{1\times3} & 0 & 0\\
0_{1\times 3} & 0& 0
\end{bmatrix}=\mathcal{X}W_1+W_2\mathcal{X}\\
=&\begin{bmatrix}
C_b^e(\omega_{ib}^b\times)-(\omega_{ie}^e\times)C_b^e & C_b^ef_{ib}^b-(\omega_{ie}^e)\times \overline{v}_{eb}^e+G_{ib}^e & -\omega_{ie}^e\times r_{eb}^e+\overline{v}_{eb}^e\\
0_{1\times3} & 0 & 0\\
0_{1\times 3} & 0& 0
\end{bmatrix}
\end{aligned}
\end{equation}
where $W_1$ and $W_2$ are denoted as
\begin{equation}\label{W_1_W_2_invariant2}
W_1=\begin{bmatrix}
\omega_{ib}^b\times & f_{ib}^b & 0\\
0_{1\times3} & 0 & 0\\
0_{1\times 3} & 0& 0
\end{bmatrix},W_2=\begin{bmatrix}
-\omega_{ie}^e\times & G_{ib}^e & \overline{v}_{eb}^n\\
0_{1\times3} & 0 & 0\\
0_{1\times 3} & 0& 0
\end{bmatrix}
\end{equation}

The right invariant error state dynamical equations of $\eta=\tilde{\mathcal{X}}\mathcal{X}^{-1}$ for attitude, velocity, and position are given as

\begin{equation}\label{attitude_n_d_invariant1}
\dot{\phi}^e=(\phi^e\times\omega_{ie}^e)+\delta\omega_{ib}^e=-\omega_{ie}^e\times\phi^e+{C}_b^e\delta\omega_{ib}^b
\end{equation}
\begin{equation}\label{Velocity_n_d_estimated_body_frame_invariant_right2}
\frac{d}{dt}\eta^v=
C_b^e\delta f_{ib}^b-\omega_{ie}^e\times J\rho_v^e+(\overline{v}_{eb}^e\times)C_b^e\delta\omega_{ib}^b+\tilde{G}_{ib}^e\times \phi^e
\end{equation}
\begin{equation}\label{Position_n_d_estimated_body_frame_invariant_right1}
\frac{d}{dt}\eta^r= 
(r_{eb}^e\times )C_b^e\delta{\omega}_{ib}^b-{\omega}_{ie}^e\times J\rho_r^e+J\rho_v^e
\end{equation}
\section{The equivalence of the $SE_2(3)$ based EKF with estimated body frame attitude and the Invariant EKF} 
\subsection{left-invariant measurement equation}
When the error state is left invariant by the left group action, this is the world-centric estimator formulation and is suitable for sensors such as GNSS, 5G, etc.
The GNSS provides navigation information in a global frame and has the left-invariant measurement equations on matrix Lie group.
GNSS  positioning solution gives the position coordinates of the antenna phase center(or other reference point), while SINS's mechanization gives the navigation results of the IMU measurement center. The two do not coincide physically, so the integrated navigation needs to correct the lever arm effect.
In the case of the arm lever error, we rearrange every measurement from GNSS as:
\begin{equation}\label{left_measurement_equation_e}
y_t=\begin{bmatrix}
r_{GNSS}^e \\ 0\\1
\end{bmatrix}=\begin{bmatrix}
C_b^e & v_{ib}^e & r_{ib}^e\\
0_{1\times3} & 1 & 0\\
0_{1\times 3} & 0& 1
\end{bmatrix}\begin{bmatrix}
l^b \\0\\1
\end{bmatrix}+\begin{bmatrix}
r_t\\0\\0
\end{bmatrix}\triangleq\mathcal{X}_tb+V_t
\end{equation}
where $r_{GNSS}^e$ is the positioning result calculated by GNSS and expressed in the ECEF frame; $l^b$ is the lever arm measurement vector expressed in the body frame; $r_t$ is measurement white noise with covariance $R_t$.

Then, the left-innovation can be defined as 
\begin{equation}\label{left_innovation_e}
\begin{aligned}
&z_t=\tilde{\mathcal{X}}_t^{-1}y_t-b=\tilde{\mathcal{X}}_t^{-1}(\mathcal{X}_tb+V_t)-b=\varepsilon_e b-b+\tilde{\mathcal{X}}_t^{-1}V_t\\
\approx & (I+\Lambda (\rho^{\tilde{b}}) )b-b+\tilde{\mathcal{X}}_t^{-1}V_t=\Lambda (\rho^{\tilde{b}}) b+\tilde{\mathcal{X}}_t^{-1}V_t\\
=&\begin{bmatrix}
\phi^{\tilde{b}}\times & \rho_v^{\tilde{b}} & \rho_r^{\tilde{b}}\\
0_{1\times 3} &0 &0\\
0_{1\times 3} &0&0
\end{bmatrix}\begin{bmatrix}
l^b \\0\\1
\end{bmatrix}+\begin{bmatrix}
\tilde{C}_e^b & -\tilde{v}_{ib}^e & -\tilde{r}_{ib}^e\\
0_{1\times 3} &1 &0\\
0_{1\times 3} &0&1
\end{bmatrix}\begin{bmatrix}
r_t\\0\\0
\end{bmatrix}=\begin{bmatrix}
\phi^{\tilde{b}}\times l^b+\rho_r^{\tilde{b}} \\0\\0
\end{bmatrix}+\begin{bmatrix}
\tilde{C}_e^b r_t \\0\\0
\end{bmatrix}\\
=&H\rho^{\tilde{b}}+\tilde{V}_t
\end{aligned}
\end{equation}
where $H$ can be abbreviated as its reduced form as $H_{rt}=\begin{bmatrix}
-l^b{\times} & 0_{3\times3}& I_{3\times 3}
\end{bmatrix}$ by considering the computational efficiency, and $H$ is independent of the system state, but only related to the known vector $b$. $\tilde{V}_t$ can be abbreviated as $\tilde{r}_t=\tilde{C}_n^e r_t=(\tilde{C}_e^n)^{-1} r_t=M_tr_t$.
It is worth noting that the invariant-innovation can be termed as innovation expressed in the body frame.
\begin{remark}
	\label{remark_left_innnovation_e}
	From the definition of left-innovation, the inverse of the estimated system state is used to multiply the measurement is reasonable as the we get the state-independent measurement matrix H. Meanwhile, the form of the left-innovation can be viewed as analogous to the GNSS positioning results minus the SINS predicted values which is .
\end{remark}

When the biased of the acceleration and gyroscope are considered, the innovation vector can be quantified as 
\begin{equation}\label{innovation_bias_e}
\tilde{z}_t=\begin{bmatrix}
-l^b{\times} & 0_{3\times3}& I_{3\times 3}& 0_{3\times3}& 0_{3\times3}
\end{bmatrix}\begin{bmatrix}
\rho^{\tilde{b}} \\ \xi^b
\end{bmatrix}+\tilde{V}_t\triangleq H_t \delta x+M_tr_t
\end{equation}
where $\xi^b$ represents error state about the bias term.

Therefore, the Kalman filter gain can be partitioned into two parts: 
\begin{equation}\label{Kalman_filter_gain_invariant_e}
K_t=\begin{bmatrix}
K_t^{\zeta} \\K_t^{\xi}
\end{bmatrix}=P_tH_t^T(H_tP_tH_t^T+M_tR_tM_t^T)^{-1}
\end{equation}

The covariance update can be calculated as  
\begin{equation}\label{covariance_update_e}
P_t^+=(I-K_tH_t)P_t(I-K_tH_t)^T+K_tM_tR_tM_t^TK_t^T
\end{equation}
\subsection{The proof of the equivalence}
Comparing equation(\ref{measurement_matrix_left_estimated}) equation(\ref{innovation_bias_e}) and we can find
\begin{equation}\label{H_new_H_invariant_e}
H_{new}=\tilde{C}_b^eH_t \Rightarrow H_t=\tilde{C}_e^bH_{new}
\end{equation}

Then, by considering the Kalman filter gain in the $SE_2(3)$-based EKF, the Kalman gain in the $SE_2(3)$ based EKF can be written as
\begin{equation}\label{Kalman_gain_new_e}
\begin{aligned}
&K_{new}=P_tH_{new}^T\left(H_{new}P_tH_{new}^T+R_t  \right)^{-1}\\
=&P_t H_{new}^T \tilde{C}_b^e \left(\tilde{C}_e^b H_{new} P_t H_{new}^T\tilde{C}_b^e +\tilde{C}_e^bR_t\tilde{C}_b^e\right)^{-1}\tilde{C}_e^b\\
=&P_t\left( \tilde{C}_e^bH_{new}\right)^T \left(\left(\tilde{C}_e^bH_{new}\right) P_t \left( \tilde{C}_e^bH_{new}\right)^T+M_tR_tM_t^T  \right)^{-1}\tilde{C}_e^b\\
=&P_tH_t^T(H_tP_tH_t^T+M_tR_tM_t^T)^{-1}\tilde{C}_e^b=K_t\tilde{C}_e^b
\end{aligned}
\end{equation}

As all the KF algorithms execute the reset state in closed loop after each measurement update step, the error state will be set as "zero" to indicate the nominal value is the same as the estimation~\cite{shin2001accuracy}. Consequently, there is no need to implement the error state prediction step after feedback is made, and the correction of the error state can be described as
\begin{equation}\label{error_state_correction_e}
\hat{x}\approx K_z\tilde{z}_t+x=K_t\tilde{z}_t
\end{equation}
Substituting equation(\ref{innovation_bias_e}) and equation(\ref{Kalman_filter_gain_invariant_e}) into the above equation, we can get
\begin{equation}\label{error_state_correction_equivalence_e}
\begin{aligned}
&\hat{x}=K_t\tilde{z}_t=P_tH_t^T(H_tP_tH_t^T+M_tR_tM_t^T)^{-1}(H_t \delta x+M_tr_t)\\
=& K_{new}\tilde{C}_b^e(\tilde{C}_e^bH_{new} \delta x+M_tr_t)=K_{new}\tilde{C}_b^e(\tilde{C}_e^bH_{new} \delta x+\tilde{C}_e^br_t)\\
=&K_{new}(H_{new} \delta x+r_t)=K_{new}\delta z_l
\end{aligned}
\end{equation}

When confronting with the covariance update, the posterior covariance update continues as normal. Equation(\ref{H_new_H_invariant_e}) and equation (\ref{Kalman_gain_new_e}) are substituted into the covariance update equation of $SE_2(3)$ based EKF, which can be obtained as
\begin{equation}\label{posterior_covariance_update_e}
\begin{aligned}
&P_{new,t}^+=(I-K_{new}H_{new})P_t(I-K_{new}H_{new})^T+K_{new}R_tK_{new}^T\\
=&(I-K_t\tilde{C}_e^b\tilde{C}_b^eH_t)P_t(I-K_t\tilde{C}_e^b\tilde{C}_b^eH_t)^T+K_t\tilde{C}_e^bR_t(K_t\tilde{C}_e^b)^T\\
=&(I-K_tH_t)P_t(I-K_tH_t)^T+K_tR_tK_t^T=P_t^+
\end{aligned}
\end{equation}

It can be seen that the error state update and covariance update have the same form. 
Therefore, the innovation in the invariant EKF can be regarded as the innovation under the body frame, while the innovation obtained by the $SE_2(3)$ based EKF with estimated body frame attitude can be regarded as the innovation under the ECEF frame. 
Although the expressions of the two are different, the final result obtained by the error state is the same, which is the error state under the body frame. 
This also proves the rationality and accuracy of the definition of left invariant innovation in invariant EKF.
\begin{remark}
	The equivalence of the left-invariant EKF and the $SE_2(3)$ based EKF can be verified in the same way and all the properties and conclusions shown in this manuscript can be obtained and proved. The detailed derivation will be given soon. Especially, the right-invariant form of the $SE_2(3)$ based EKF is suitable for the local navigation such as vision, lidar, whose measurements have the right-invariant observation form. Moreover, it is more reasonable to define the error as the estimated state multiplies the inverse of state, i.e., $\eta=\hat{\mathcal{X}}\mathcal{X}^{-1}$.
\end{remark}
\section{$SE_2(3)$ based smoothing algorithm}
Since the $SE_2(3)$ based EKF is easier to understand and its formulation is more intuitive than the Invariant EKF, we propose the $SE_2(3)$ based smoothing algorithm which is essentially an application of the $SE_2(3)$ based EKF.
Our formulation is simple and easy to understand and different to the invariant RTS smoother~\cite{van2020invariant}.

As the $SE_2(3)$ based EKF is more simple than the invariant-EKF, the $SE_2(3)$ based EKF implement the smoothing procedure as the RTS smoothing. The only difference lies in the full state update procedure, which need the matrix exponential map and multiplication operation on matrix Lie group.
\section{Conclusions}
In this paper, $SE_2(3)$ based EKF and smoothing framework is derived from the perspective of matrix Lie group. This is nature and reasonable, consequently leads to common error representation for the inertial-integrated navigation system.
The major contribution of this paper is the complete theory development of $SE_2(3)$ based EKF framework for inertial-integrated navigation system which can be applied to four different state representations and four different error definitions. 
The group-affine property of the dynamics is verified.
The attitude error, the velocity error, and the position error are defined on the common frame.
The experiments show that the proposed $SE_2(3)$ is robust for initial-integrated navigation with large misaligned angle.
In the future, the biases of the accelerometer and gyroscope can be considered to incorporate into matrix Lie group $SE_4(3)$ so that the orientation error can be considered for the biases in the inertial-integrated navigation.
Furthermore, navigation application requirements are generally much different, the theory proposed in this manuscript is supposed to be applied to more inertial-integrated navigation applications such as initial alignment, tightly couple integration, filter-based SLAM, etc.
\vspace{2ex}

\noindent
{\bf\normalsize Acknowledgement}\newline
{This research was supported by a grant from the National Key Research and Development Program of China (2018YFB1305001). 
We express thanks to professor Xiaoji Niu from the GNSS Research Center, Wuhan University.} \vspace{2ex}

\bibliographystyle{IEEEtran}
\bibliography{ref.bib}

\end{document}